\documentclass[final,12pt]{msml2022} 

\usepackage{amsmath,amssymb,amsfonts}

\usepackage{graphicx}
\usepackage{xspace}

\usepackage{nicefrac}
\usepackage{cite}
\usepackage{framed}
\usepackage{algorithm}
\usepackage{algpseudocode}
\usepackage[algo2e]{algorithm2e} 
\usepackage{enumerate}
\usepackage{cleveref}

\newtheorem{assumption}{Assumption}

\newcommand{\dotprod}[1]{\left< #1 \right>}
\newcommand{\lp}{\left(}
\newcommand{\rp}{\right)}

\newcommand{\n}[1]{\left\|#1 \right\|}
\newcommand{\ns}[1]{\n{ #1}^2}
\newcommand{\E}[1]{\mathbb{E}\left[#1\right]}
\newcommand{\Ek}[1]{\mathbb{E}_k\left[#1\right]}
\newcommand{\Ekp}[1]{\mathbb{E}_{k+1}\left[#1\right]}
\newcommand{\Exi}[1]{\mathbb{E}_\xi\left[#1\right]}
\newcommand{\avein}{\frac{1}{n}\sum_{i=1}^n}

\newcommand{\eqdef}{\stackrel{\mathrm{def}}{=}}

\newcommand{\x}{x}
\newcommand{\xk}{\x^{k}}
\newcommand{\xkp}{\x^{k+1}}
\newcommand{\xs}{\x^{\star}}
\newcommand{\xis}{y_i^k}
\newcommand{\g}{g}
\newcommand{\gik}{\g_i^k}
\newcommand{\gis}{s_i^k}
\newcommand{\fis}{f_i^k}
\newcommand{\xik}{\xi_i^k}

\newcommand{\G}{G}

\newcommand{\stp}{\texttt{StoP}\xspace}
\newcommand{\stps}{\texttt{StoPS}\xspace}

\newcommand{\grad}{\texttt{GraD}\xspace}
\newcommand{\grads}{\texttt{GraDS}\xspace}
\newcommand{\sps}{\texttt{SPS}\xspace}

\newcommand{\step}{\eta}
\newcommand{\stepk}{\step^k}

\newcommand{\gd}{\text{pgd}}
\newcommand{\agd}{\text{agd}}

\newcommand{\R}{\mathbb R}

\newcommand{\cL}{\mathcal L}
\newcommand{\cO}{\mathcal O}

\DeclareMathOperator{\argmin}{argmin}

\def\<#1,#2>{\langle #1,#2\rangle}

\makeatletter
\newcommand{\printfnsymbol}[1]{%
  \textsuperscript{\@fnsymbol{#1}}%
}

\title[Adaptive Learning Rates for Faster Stochastic Gradient Methods]{\stp and \grad: Adaptive Learning Rates for Faster Stochastic Gradient Methods}
\usepackage{times}

\msmlauthor{%
 \Name{Samuel Horv\'{a}th}\thanks{Equal contribution.} \Email{samuel.horvath@mbzuai.ac.ae}\\
 \addr Mohamed bin Zayed University of Artificial Intelligence, United Arab Emirates%
  \AND
 \Name{Konstantin Mishchenko}\printfnsymbol{1} \Email{konsta.mish@gmail.com}\\
 \addr DI ENS, Ecole normale supérieure, Université PSL, CNRS, INRIA, France%
 \AND
 \Name{Peter Richt\'{a}rik} \Email{richtarik@gmail.com}\\
 \addr King Abdullah University of Science and Technology, Saudi Arabia%
}

\makeatletter
 \let\Ginclude@graphics\@org@Ginclude@graphics
\makeatother

\begin{document}

\maketitle

\begin{abstract}%
In this work, we propose new adaptive step size strategies that improve several stochastic gradient methods. Our first method (\stps) is based on the classical Polyak step size~\citep{polyak1987introduction} and is an extension of the recent development of this method for the stochastic optimization--\sps~\citep{sps}, and our second method, denoted \grads,  rescales step size by ``diversity of stochastic gradients''. We provide a theoretical analysis of these methods for strongly convex smooth functions and show they enjoy deterministic-like rates despite stochastic gradients. Furthermore, we demonstrate the theoretical superiority of our adaptive methods on quadratic objectives. Unfortunately, both \stps and \grads are dependent on unknown quantities, which are only practical for the overparametrized models. To remedy this, we drop this undesired dependence and redefine \stps and \grads to \stp and \grad, respectively. We show that these new methods converge linearly to the neighbourhood of the optimal solution under the same assumptions. Finally, we corroborate our theoretical claims by experimental validation, which reveals that \grad is particularly useful for deep learning optimization. 
\end{abstract}


\section{Introduction}
We consider minimizing the stochastic objective of the form
\begin{align}
\label{eq:problem}
	\min_{x \in \R^d} \left\{f(x)\eqdef \Exi{f(x, \xi)} \right\}.
\end{align}
We assume to have access to an oracle that provides us with independent and unbiased gradient samples of $f$, $\nabla f(\cdot, \xi)$, such that $\Exi{\nabla f(\cdot, \xi)} = \nabla f(\cdot)$. We denote by $x^\star$ the optimal solution of \eqref{eq:problem}.

We are interested in stochastic gradient methods, for which one step can be written in the form
\begin{align*}
	\xkp = \xk - \stepk T^k\lp \G^k \rp,
\end{align*}
where $\G^k$ is a set of all the past gradients, $T^k$ is a gradient transformation and $\stepk > 0$ is a step size. The canonical example is Stochastic Gradient Descent (\texttt{SGD})~\citep{robbins1951stochastic, nemirovsky1983problem, nemirovski2009robust, SGD-AS}, in which updates are based on single data point or small batches of points, i.e.\;  $T^k\lp \G^k  \rp = \avein \gik$, where $\g_i^k \eqdef \nabla f(\xk, \xik)$. This is a workhorse used for training supervised machine learning problems of the generic form \eqref{eq:problem}. The terrain around the basic \texttt{SGD} method has been thoroughly explored in recent years, resulting in theoretical and practical enhancements such as  Nesterov acceleration \citep{allen2017katyusha}, Polyak momentum \citep{polyak1964some, sutskever2013importance}, adaptive step sizes \citep{duchi2011adaptive, kingma2014adam, reddi2019convergence, malitsky2019adaptive}, variance reduction~\citep{shalev2013stochastic, johnson2013accelerating}, distributed optimization \citep{ma2017distributed, alistarh2017qsgd, stich2018local}, importance sampling \citep{zhao2015stochastic, qu2015quartz}, higher-order optimization \citep{tripuraneni2018stochastic, kovalev2019stochastic}, and several other useful techniques. 

A particularly productive approach to enhancing \texttt{SGD} is \textbf{a careful step size selection}. Step size is one of the most crucial parameters for guaranteeing the convergence of \texttt{SGD}. Many works have discussed different ways of selecting step size over the last few years. Non-asymptotic analysis of SGD with constant step size for convex and (quasi) strongly convex functions was provided in~\citep{moulines2011non, needell2016stochastic, nguyen2018sgd, SGD-AS}. 
For non-convex functions, such an analysis can be found in~\citep{ghadimi2013stochastic, bottou2018optimization}. Using a constant step size for \texttt{SGD} guarantees convergence to a neighbourhood of the solution in case of non-zero stochastic gradient variance at the solution. A common technique to overcome this issue and to guarantee convergence to the exact optimum is to use a decreasing step size~\citep{robbins1951stochastic, ghadimi2013stochastic,nemirovski2009robust, karimi2016linear} or to employ variance reduction~\citep{shalev2013stochastic, johnson2013accelerating} in case \eqref{eq:problem} is of the finite-sum form. 
More recently, adaptive methods~\citep{duchi2011adaptive, liu2019variance, kingma2014adam, li2019convergence, ward2019adagrad, malitsky2019adaptive} that adjust the step size on the fly have become widespread and are particularly useful when training deep neural networks.

A promising recent direction is in extending the Polyak stepsize beyond deterministic gradient descent. For instance, \citep{gower2021stochastic} and \citep{gower2022cutting} consider stochastic reformulations with extra slack variables that allow for a better theory when interpolation does not hold or optimal values of the stochastic losses are not known. A similar approach was used by \citep{li2022sp2} to design a second-order variant of Polyak's method.

 \begin{algorithm}[t]
 \KwIn{  $x^0 \in \R^d$, $\stepk$'s, and $T^k$'s}
 \For{$k = 1,2,\dots$}{
 	  Sample minibatch of gradients $g_1^k,\dotsc, g_n^k$\; \\
 	 $\xkp = \xk - \stepk T^k\lp \G^k \rp$\;
}
\caption{General Stochastic Gradient Methods (\texttt{GSGM}s)}
 \label{alg:scagrad}
 \end{algorithm}

\section{Contributions}
The construction of adaptive optimization methods is especially challenging in the stochastic setting. The main problem stems from the stochastic noise due to non-exact gradient estimation. We overcome this problem by employing correction for gradient estimates and adaptive step size scalings. We show that this adjustment can bridge the gap between the convergence rates of deterministic and stochastic methods. Unfortunately, these corrections are only theoretical as they depend on the information at the minimizer of $f$. On the other hand, these quantities are known for over-parametrized models, i.e., models that fit the training dataset entirely.

Our contributions can be summarized as follows.
\begin{itemize}
\item \textbf{\stps}: Inspired by the classical Polyak step size~\citep{polyak1987introduction} commonly used with the deterministic subgradient method~\citep{hazan2019revisiting, boyd2003subgradient} and recent interest in a stochastic version of this step size~\citep{berrada2020training, oberman2019stochastic, sps, gower2021stochastic}, we propose a novel adaptive learning rate for \texttt{SGD}. In particular, we propose an extension of the recently proposed stochastic Polyak step size--\sps of \citet{sps}. We refer to it as \textit{stochastic Polyak step size} (\stps). 
\item \textbf{\grads}: Our second method is based on the notion of gradient diversity, which is a measure of dissimilarity between concurrent gradient updates proposed in \citep{yin2018gradient}. The authors showed that this quantity plays a crucial role in the performance of mini-batch \texttt{SGD}. In our work, we propose to scale the stochastic gradient updates by their adjusted diversity, defined as 
\[
	\frac{\avein \ns{\nabla f(x, \xi_i) - \nabla f(\xs, \xi_i)}}{\ns{\avein \nabla f(x, \xi_i) - \nabla f(\xs, \xi_i)}}.
\]
We refer to this method as \textit{gradient diversity step size} (\grads). We show that \grads and \stps can be linked together, where \grads step size is always lower than the \stps step size.
\item \textbf{Linear convergence}:  We provide a theoretical convergence analysis for both methods, assuming that the global objective is strongly convex and stochastic functions are almost surely smooth and convex. We show that both \stps and \grads converge linearly to the optimal solution. Furthermore, we take a closer look at consistent linear models where we show that \stps and \grads are identical in this regime, and they have superior performance both theoretically and empirically when compared to \texttt{GD}).
\item \textbf{Practical variants --\stp and \grad}: We show that our adaptive step sizes are also applicable beyond overparameterization. We introduce \stp and \grad algorithms---implementable versions of \stps and \grads, respectively, which are still adaptive but do not depend on any unknown quantities. We show that these methods achieve linear convergence to a neighbourhood of the optimal solution. The price that we pay for not using unknown quantities is that the neighbourhood size can't be controlled with decreasing step size. Even though such a solution might not be precise enough, in this case our adaptive methods still provide a good starting point. Furthermore, since in modern machine learning tasks the solution accuracy is needed only up to the statistical accuracy~\citep{bottou2007tradeoffs}, the convergence to a neighbourhood is sufficient for good generalization performance. 
\item \textbf{Combining different adaptive strategies}: We show that our adaptive methods can be combined with other adaptive strategies such as Polyak momentum. To be more concrete, we propose combining the \grad algorithm and Polyak momentum. We show that such combination leads to superior theoretical convergence compared to the plain Polyak momentum; see Section~\ref{sec:grad_momentum} for details.
\item \textbf{Experimental validation}: Lastly, we conclude several experiments to support further and validate our theoretical results. In addition, we show that \grad is particularly useful for optimizing deep neural networks. 
\end{itemize}

\section{Notations and Assumptions}

In this section, we define several quantities and assumptions that are used throughout the paper.

\begin{assumption}[Smoothness]
\label{ass:smoothness}
	$f(\cdot, \xi)$ is $L$-smooth almost surely, so for any $x,y\in\R^d$ it holds
	\begin{align}
	\label{eq:loc_smoothnes}
		f(x, \xi) - f(y, \xi) - \dotprod{\nabla f(y, \xi), x - y} \le \frac{L}{2}\ns{x-y} \text{ with probability } 1.
	\end{align}
\end{assumption}

\begin{assumption}[Strong-convexity]
\label{ass:str_convexity}
$f(\cdot)$ is $\mu$ strongly-convex if for every $x, y\in\R^d$ it holds
	\begin{align}
	\label{eq:mu_convexity}
		\frac{\mu}{2}\ns{x-y} \le f(x) - f(y) - \dotprod{\nabla f(y), x - y}.
	\end{align} 
\end{assumption}

\begin{assumption}[Convexity]
\label{ass:convexity}
$f(\cdot, \xi)$ is almost surely convex if for every $x, y\in\R^d$ it holds
	\begin{align}
	\label{eq:convexity}
		0\le f(x, \xi) - f(y, \xi) - \dotprod{\nabla f(y, \xi), x - y} \text{ with probability } 1.
	\end{align} 
\end{assumption}

\begin{assumption}[Overparameterization]
\label{ass:overparametrization}
$f$ is overparameterized, i.e.,
 \begin{align}
 \label{eq:overparametrization}
 \nabla f(\xs, \xi)=0 \text{ with probability } 1,
 \end{align}
  where $\xs$ is a minimizer of $f$.
\end{assumption}

\begin{definition}[Gradient Diversity]
\label{def:grad_div}
Let $f$ have the unique minimizer $\xs$. We define adjusted gradient diversity to be 
\begin{align}
\label{eq:adjusted_grad_div}
\agd(x, \xi) \eqdef \frac{\avein \ns{\nabla f(x, \xi_i) - \nabla f(\xs, \xi_i)}}{\ns{\avein [\nabla f(x, \xi_i) - \nabla f(\xs, \xi_i)]}}.
\end{align}
We also define plain gradient diversity
\begin{align}
\label{eq:grad_div}
\gd(x, \xi) \eqdef \frac{\avein \ns{\nabla f(x, \xi_i)}}{\ns{\avein \nabla f(x, \xi_i)}}.
\end{align}
\end{definition}

As previously discussed, our notion of plain gradient diversity is identical to the notion of gradient diversity introduced in \citep{yin2018gradient}, and adjusted gradient diversity is a novel definition introduced in this work. Both $\agd(x, \xi) $ and $\gd(x, \xi)$ are always greater than $1$, which follows from Cauchy-Schwartz inequality. Finally, note that $\agd(x, \xi) $ and $\gd(x, \xi)$ are random variables as they depend on stochastic gradients.

\section{Adaptive Learning Rates for Faster Stochastic Gradient Methods}

This section presents our methods and their theoretical convergence guarantees. 

\subsection{Stochastic Polyak Method: \stps}
\label{sec:stps}
For \stps, our proposed step size has the following form 
\begin{align}
\label{eq:stps_stepsize}
\gamma_{\stps}(x, \xi) \eqdef 2 \frac{\avein \left[ f(x, \xi_i) - f(\xs, \xi_i) - \dotprod{\nabla f(\xs, \xi_i), x - \xs}\right]}{\ns{\avein [\nabla f(x, \xi_i) -  \nabla f(\xs, \xi_i)]}}.
\end{align}
In comparison to the \sps step size of~\citep{sps},
\[
	\gamma_{\sps}(x, \xi) = 2 \frac{\avein \left[ f(x, \xi_i) - f(\xs, \xi_i)\right]}{\ns{\avein \nabla f(x, \xi_i)}},
\] 
we have an extra quantity $\dotprod{\nabla f(\xs, \xi_i), x - \xs}$ in the numerator and $\nabla f(x, \xi_i)$ in denominator, which are crucial components to obtain deterministic-like convergence guarantees as we show later in this section.
Equation~\eqref{eq:stps_stepsize} reduces to standard Polyak step size~\citep{polyak1987introduction} in the deterministic case as $ f(\xs, \xi)  =  f(\xs) = \min_{x \in \R^d} f(x)$ and $\nabla f(\xs, \xi)  = \nabla  f(\xs) = \mathbf{0}$  for unconstrained optimization. 
In addition, for the overparameterization defined in Assumption~\ref{ass:overparametrization}, $\nabla f(\xs, \xi_i)$ is all-zeros vector by definition and $f(\xs, \xi_i)$ is usually easy to compute, e.g., it is $0$ for majority of commonly used losses in the modern machine learning, e.g., mean squared error or cross-entropy loss.  We note that in this case $\gamma_{\stps}(x, \xi) = \gamma_{\sps}(x, \xi)$.

Below, we introduce our version of the Polyak step size \texttt{SGD} algorithm (\stps), i.e., \texttt{SGD}  with $\gamma_{\stps}$ step size
\begin{align}
\label{eq:SGD_stps}
\xkp = \xk  - \gamma_{\stps}^k \avein [\gik - \gis],
\end{align}
 where $\gis \eqdef \nabla f(\xs, \xi_i^k)$ and $\gamma_{\stps}^k \eqdef \gamma_{\stps}(\xk)$. Note that it is not only important to remove stochastic noise from $\gamma_{\stps}^k$ but also from the gradient estimator, i.e., $\gik \rightarrow \gik - \gis$.
It is clear that this algorithm is only practical in the overparameterized regime since it requires knowledge of $f(\xs, \xi)$, $\gis$ and $\dotprod{\gis, x - \xs}$  for every $x, \xi$. Although not practical in general, we believe this algorithm still has theoretical value. It sheds light on obtaining the linear rate to the exact solution for the stochastic version of Polyak step size. Furthermore, it recovers the same rate in the deterministic and overparameterized settings, as shown in the following theorem. 

\begin{theorem}
\label{thm:stps_convergence}
Let Assumptions~\ref{ass:str_convexity} and \ref{ass:convexity} hold for $f(\cdot)$~\eqref{eq:problem}, then the iterates defined by \eqref{eq:SGD_stps} satisfy
\begin{align}
\label{eq:stps_convergence}
\E{\ns{x^K - \xs}} \leq \lp 1 - \mu \gamma_{\stps}^{\min} \rp^K \ns{x^0 - \xs},
\end{align}
where 
$
\gamma_{\stps}^{\min}= \inf_{x \in \R^d} \gamma_{\stps}(x).
$
\end{theorem}

We see that $\gamma_{\stps}^{\min}$ appears in the convergence rate. It is useful to study its lower-bound to provide the worst-case guarantee. If Assumptions~\ref{ass:smoothness} and \ref{ass:convexity} hold, we can  apply smoothness and convexity to the numerator in~\eqref{eq:stps_stepsize} followed by Cauchy-Schwartz inequality and we obtain the following
\begin{align}
\label{eq:stps_geq_grads}
\gamma_{\stps}(x) \geq  \frac{2}{2L} \frac{\avein\ns{\nabla f(x, \xi_i) - \nabla f(\xs, \xi_i)}}{\ns{\avein [\nabla f(x, \xi_i) - \nabla f(\xs, \xi_i)]}} \geq \frac{1}{L}.
\end{align}
Therefore, the speed of convergence to the $\epsilon$-approximate solution, i.e., $\E{\ns{x^K - \xs}} \leq \epsilon$ is, in the worst-case, $K\ge \nicefrac{L}{\mu} \log\left(  \nicefrac{\ns{x^0 - \xs}}{\epsilon} \right) = \cO\left(\nicefrac{L}{\mu} \log \nicefrac{1}{\epsilon} \right)$.

\subsection{Scaling Gradient by its Diversity: \grads}
\label{sec:grads}
For \grads, our proposed step size scaling has the following form 
\begin{align}
\label{eq:grads_stepsize}
\gamma_{\grads}(x, \xi) \eqdef  \agd(x, \xi).
\end{align}
Comparing to $\gamma_{\stps}(x)$, $\gamma_{\grads}(x)$ does not requires knowledge of $f(\xs, \xi)$. In the deterministic case, $\gamma_{\grads}(x, \xi) = \nicefrac{\ns{\nabla f(x)}}{\ns{\nabla f(x)}} = 1$. If we have overparameterization, $\gamma_{\grads}(x)$ does not require knowledge of $\nabla f(\xs, \xi)$'s as these are zero vectors by definition. 
 
Below, we introduce the update formula for the \grads algorithm, i.e., \texttt{SGD}  with $\gamma_{\grads}$ step size scaling
\begin{align}
\label{eq:SGD_grads}
\xkp = \xk  - \eta^k \gamma_{\grads}^k \avein (\gik - \gis),
\end{align}
where $\gis = f(\xs, \xi_i^k)$, $\gamma_{\grads}^k = \gamma_{\grads}(\xk)$ and $\eta^k \geq 0$ is a step size without scaling. As in the previous subsection, it is clear that this algorithm is only practical in the overparameterized regime since it requires the knowledge of $\gis$ for each $\xi_i$. Below, we provide a convergence guarantee for this algorithm and show that it converges linearly to the exact solution.

\begin{theorem}
\label{thm:grads_convergence}
Let Assumptions~\ref{ass:smoothness}, \ref{ass:str_convexity} and \ref{ass:convexity} hold for $f(\cdot)$~\eqref{eq:problem}, then for $\eta^k \eqdef \eta = \nicefrac{1}{L}$ for all $k \in [K]$ the iterates defined by \eqref{eq:SGD_grads} satisfy
\begin{align}
\label{eq:grads_convergence}
\E{\ns{x^K - \xs}} \leq \lp 1 - \gamma_{\grads}^{\min}\frac{\mu}{L}  \rp^K \ns{x^0 - \xs},
\end{align}
where 
$
\gamma_{\grads}^{\min}= \inf_{x \in \R^d} \gamma_{\grads}(x) \geq 1.
$
\end{theorem}

One of the direct consequences of the above theorem is that the speed of convergence to  $\epsilon$-approximate solution is, in the worst-case, $\nicefrac{L}{\mu} \nicefrac{\log \ns{x^0 - \xs}}{\epsilon} = \cO(\nicefrac{L}{\mu} \log \nicefrac{1}{\epsilon})$.

\begin{remark}
The smoothness constant $L$ that appears in the convergence analysis is clearly larger than the smoothness constant for the full function $f$ (denote $L_f$). For non-adaptive \texttt{SGD} with overparameterized models, one expects to obtain the convergence rate  $\tilde{\cO}(\nicefrac{L_f}{\mu})$. This rate can be obtained for both \stps and \grads algorithms with step sizes 
\[\max\{\nicefrac{1}{L_f},  \gamma_{\stps}^k\} \text{ and } \max\{\nicefrac{1}{L_f}, \nicefrac{\gamma_{\grads}^k}{L}\},\] respectively. In the next section, we show that this extra step might be not needed  as \[\max\{\nicefrac{1}{L_f},  \gamma_{\stps}^k\} = \gamma_{\stps}^k \text{ and } \max\{\nicefrac{1}{L_f}, \nicefrac{\gamma_{\grads}^k}{L}\} = \nicefrac{\gamma_{\grads}^k}{L} \] for quadratics.
\end{remark}

Finally, we would like to compare $\gamma_{\grads}$ to $\gamma_{\stps}$ to see which method is better. \stps is more demanding to extra knowledge of parameters at the optimum, but it is independent of the smoothness constant $L$. On the other hand, \eqref{eq:stps_geq_grads} implies that $\gamma_{\grads} \leq\gamma_{\stps}$ under the Assumptions~\ref{ass:smoothness}  and \ref{ass:convexity}, therefore \stps is not theoretically worse then \grads if all parameters are known, e.g., overparameterization holds.

\subsection{\stps and \grads for Solving Consistent Linear Systems}
In this section, we assume the following specific version of the objective \eqref{eq:problem}
\begin{align}
\label{eq:linear}
	\min_{x \in \R^d} \left\{\frac{1}{2n}\ns{Ax - b} = \avein \frac{1}{2} \lp a_i^\top x - b_i \rp^2 \right\},
\end{align}
where $A$ is a $n \times m (n \leq m)$ matrix and $n$ is the same as batch size. Moreover, we assume that $A$ has normalized rows $a_i^\top$, i.e., the norm of each row $\ns{a_i} = 1$, and there exists a unique solution $A\xs = b$. This setting corresponds to an overparametrized model as we can minimize the objective to $0$, thus \stps and \grads are applicable.  For linear systems, local smoothness $L$ and global smoothness $L_f$ can be computed exactly. $L = \ns{a_i^\top} = 1$ and $L_f = \frac1n\ns{A}$. We assume that in each step we compute full gradient $\nabla f(x) = \avein (a_i^\top x - b_i) a_i$. It is easy to show that both \stps and \grads lead to the same method with the step size equal to $\nicefrac{\avein (a_i^\top x - b_i)^2 }{\ns{\avein (a_i^\top x - b_i)a_i^\top}}$. Let us define $y = A(x - \xs)$, then
\begin{align}
\label{eq:stps_geq_gd}
\frac{\avein (a_i^\top x - b_i)^2 }{\ns{\avein (a_i^\top x - b_i)a_i^\top}} = \frac{\ns{y}}{\frac1n \ns{A^\top y}} = \frac{\ns{y} \ns{A^\top}}{\ns{A^\top y}} \frac{1}{\frac1n \ns{A^\top}} \geq \frac{1}{\frac1n \ns{A}}  = \frac{1}{L_f}.
\end{align}
$\nicefrac{1}{L_f}$ is up to a constant $2$ the optimal constant step size for non-adaptive \texttt{GD} (the optimal value is $\nicefrac{2}{(\mu + L_f)}$), therefore our adaptive methods leads to a provable improvement as shown in \eqref{eq:stps_geq_gd}.  In addition, our adaptive methods converge in one step as long as $x - \xs$ is an eigenvector of $A^\top A$ while non-adaptive \texttt{GD} only converges in one step if  $x - \xs$ corresponds to the eigenvector with the largest eigenvalue; see Figure~\ref{fig:linear_2d}. Finally, we would like to remark that the existence of a unique solution, full gradient computation and the normalized rows are not necessary and can be relaxed using stochastic reformulation from~\citep{richtarik2020stochastic}. 

\subsection{Beyond Deterministic and Overparameterized Models: \stp and \grad}
\label{sec:stp_and_grad}
As we presented in Sections~\ref{sec:stps} and \ref{sec:grads}, \stps and \grads might be not practical methods as they require knowledge of quantities that are not usually available. Therefore, we design \stp and \grad\ -- practical adaptive methods. The main difference is that we implicitly assume overparameterization, i.e.,  $\nabla f(x, \xi) = \mathbf{0}$. It is no surprise that this change comes with a price. We need to add a control mechanism for step size, and we can only guarantee the convergence to a fixed size neighbourhood due to potentially inaccurate step size estimation.  

For \stp our method has the following form
\begin{align}
\label{eq:SGD_stp}
\xkp = \xk  - \gamma_{\stp}^k \avein \gik, \text{ where } \gamma_{\stp}^k =  \min\left\{\gamma_{\max}^k,  2\frac{\avein (f(\xk, \xik) - \fis)}{\ns{\avein \gik}}\right\}
\end{align}
where $\gamma_{\max}^k$'s are given and $\fis$'s are chosen such that $\fis \leq \inf_{x \in \R^d} f(x, \xik)$. A convergence guarantee follows.

\begin{theorem}
\label{thm:stp_convergence}
Let Assumptions~\ref{ass:str_convexity} and \ref{ass:convexity} hold for $f(\cdot)$ from \eqref{eq:problem}, then the iterates defined by \eqref{eq:SGD_stp} satisfy
\begin{align}
\label{eq:stp_convergence}
\E{\ns{x^K - \xs}} \leq \lp 1 - \mu \gamma_{\stp}^{\min} \rp^K \ns{x^0 - \xs} + 4\frac{\sigma^2_{\stp}}{\mu},
\end{align}
where $\gamma_{\max}^k = \lp 1 + \nicefrac{\mu \gamma_{\stp}^{\min}}{2}\rp \gamma_{\stp}^{k-1}$, $\gamma_{\max}^0 = \gamma_{\stp}^{\min}$, $\sigma^2_{\stp} = \Exi{f(\xs, \xik) - \fis}$ and 
\[
\gamma_{\stp}^{\min} = \inf_{x \in \R^d}  2\frac{\avein (f(x, \xik) - \fis)}{\ns{\avein \nabla f(x, \xik)}}.
\]
\end{theorem}

If Assumption~\ref{ass:smoothness} holds and $\fis = \inf_{x \in \R^d} f(x, \xik)$ we can lower-bound $\gamma_{\stp}^{\min} \geq \nicefrac{1}{L} $ by applying smoothness and convexity of $f(\cdot, \xi)$ to \eqref{eq:SGD_stp}. Therefore,  Theorem~\ref{thm:stp_convergence} guarantees linear convergence to the neighborhood of size $\nicefrac{4\sigma^2_{\stp}}{\mu}$ with the worst-case convergence rate of $\cO(\nicefrac{L}{\mu} \log \frac{1}{\epsilon})$. 

In the literature, we are aware of two similar methods --- $\texttt{SPS}_{\max}$~\citep{sps} and \texttt{ALI-G}~\citep{berrada2020training}, with the following step sizes
\begin{align*}
\gamma^k_{\texttt{SPS}_{\max}} &= \min\left\{\gamma_{\max},  \frac{\avein (f(\xk, \xik) - \fis)}{c\ns{\avein \gik}}\right\} \\
  \gamma^k_{\texttt{ALI-G}} &= \min\left\{\gamma_{\max},  \frac{\avein (f(\xk, \xik) - \fis)}{\ns{\avein \gik} + \delta}\right\}.
\end{align*}

The main difference is that the upper bound on the step size is fixed, while we allow increasing unbounded upper bound, which is a popular practical algorithmic choice~\citep{sps}.  In addition,  \texttt{ALI-G} adds a small constant $\delta$ to the norm of the gradient in the denominator to avoid step size explosion.

For \grad, we scale step size by gradient diversity.
\begin{align}
\label{eq:SGD_grad}
\xkp = \xk  - \eta^k\gamma_{\grad}^k \avein \gik, \text{ where } \gamma_{\grad}^k =  \min\left\{\gamma_{\max}^k,  \frac{\avein \ns{\gik}}{\ns{\avein \gik}}\right\}
\end{align}
where $\gamma_{\max}^k$'s are given. A convergence guarantee follows.

\begin{theorem}
\label{thm:grad_convergence}
Let Assumptions~\ref{ass:smoothness}, \ref{ass:str_convexity} and \ref{ass:convexity} hold for $f(\cdot)$ from \eqref{eq:problem} and $\eta^k \eqdef \eta = \nicefrac{1}{2L}$ for all $k \in [K]$, then the iterates defined by \eqref{eq:SGD_grad} satisfy
\begin{align}
\label{eq:grad_convergence}
\E{\ns{\xk - \xs}} \leq \lp 1 - \nicefrac{\mu \gamma_{\grad}^{\min}}{2L} \rp^k \ns{x^0 - \xs} + 16\nicefrac{\sigma^2_{\grad}}{\mu},
\end{align}
where $\gamma_{\max}^k = \lp 1 + \nicefrac{\mu \gamma_{\grad}^{\min}}{4}\rp \gamma_{\grad}^{k - 1}$, $\gamma_{\max}^0 = \gamma_{\grad}^{\min}$, $\sigma^2_{\grad} = \Exi{L\ns{\xs - \xis}}$,  \\ $\xis = \argmin_{x \in \mathcal{X}^\star} \ns{x - \xs}$, $\mathcal{X}^\star = \argmin_{x \in \R^d} f(x, \xik)$, and 
\begin{align}
\label{eq:gamma_grad_min}
\gamma_{\grad}^{\min} = \min_{x \in \R^d} \frac{\avein \ns{\nabla f(x, \xik)}}{\ns{\avein \nabla f(x, \xik)}} \geq 1.
\end{align}
\end{theorem}
\begin{algorithm}[t]
\KwIn{$x^0 \in \R^d$, $m^0=0$, step sizes $\gamma^k>0$ and momentum values $\beta^k$ for $k=1,2, \dots$}
 \For{$k = 1,2,\dots$}{
 	Sample minibatch of gradients $g_1^k,\dotsc, g_n^k$\; \\
 	$m^k=(1-\beta^k)\gamma^k_{\grad}\avein g_i^k + \beta^k m^{k-1}$\; \\
 	$x^{k+1}=x^k  - \eta^k m^k$\;
}
 \caption{\grad with momentum}
 \label{alg:moscagrad}
 \end{algorithm}
The above theorem guarantees linear convergence to the $16\nicefrac{\sigma^2_{\grad}}{\mu}$ neighborhood of the optimal solution with the worst-case convergence rate of $\cO(\nicefrac{L}{\mu} \log \nicefrac{1}{\epsilon})$, which is the same as for \stp with $\fis = \inf_{x \in \R^d} f(x, \xik)$. 

In the literature, there is one similar algorithm \texttt{AdaScale SGD}~\citep{adascale} that uses the same step size scaling without the upper bound $\gamma_{\max}^k$. Contrary to our \grad,  this algorithm comes with no theoretical guarantees. 

As a possible future work, we believe that for the finite-sum type of problems there is a potential to employ variance reduction on top of the \grad algorithm, e.g., using the loopless estimator from~\citep{kovalev2020don}, to remove the neighbourhood. 

Finally, we want to provide a meaningful comparison between \stp and \grad. Using smoothness and convexity of $f(\cdot, \xi)$, it is easy to see that it always holds that $\gamma_{\stp}^k \geq \gamma_{\grad}^k$. To compare neighborhoods,  we assume that we pick the best possible $\fis = \argmin_{x \in \R^d} f(x, \xi_i)$ for \stp, otherwise the comparison is unclear and depends on how close are $\{\fis = \argmin_{x \in \R^d} f(x, \xi_i)\}$. In this best-case scenario, $\sigma^2_{\stp} \leq \sigma^2_{\grad}$ which follows directly from the smoothness assumption. Therefore, if we can select $\fis = \argmin_{x \in \R^d} f(x, \xi_i)$ and all assumptions hold then \stp should be the preferred method.
  
\subsection{Combination with Other Adaptive Methods: \grad + Momentum}
\label{sec:grad_momentum}

In this section, we show that our adaptive methods can be combined with other adaptive strategies, more concretely, in Algorithm~\ref{alg:moscagrad} we present a momentum version of the \grad algorithm. We show that one can scale stochastic gradients with $\gamma^k_{\grad}$ defined in \eqref{eq:SGD_grad}  and use this scaled gradient estimator with standard momentum. In the theorem below, we provide a convergence guarantee under the convexity assumption to showcase that our adaptive methods are also applicable beyond strongly-convex objectives. For ease of exposition, we assume a simplistic setting with constant step size and overparameterization. A more general setting can be also analyzed using techniques presented in the previous section.

\begin{theorem}
\label{thm:momentum_convergence}
	Let Assumptions~\ref{ass:smoothness}, \ref{ass:convexity} and \ref{ass:overparametrization} hold for $f(\cdot)$ from \eqref{eq:problem} and $\eta^k \eqdef \eta = \nicefrac{1}{2L}$ for all $k \in [K]$. Furthermore, let $\gamma^k_{\max} \eqdef \gamma^{\min}_{\grad}$ defined in \eqref{eq:gamma_grad_min} and $\beta^k = \nicefrac{(k-1)}{(k+1)}$ for all $k \in [K]$, then
	\begin{align*}
		\E{f(x^K)-f^*}\le \frac{L\|x^1-\xs\|^2}{2\gamma^{\min}_{\grad} K}.
	\end{align*}
\end{theorem}

Comparing our rate to the rate of Polyak momentum under the same assumptions, the difference is an extra $\gamma^{\min}_{\grad}$ term in the denominator. We know that $\gamma^{\min}_{\grad} \geq 1$, therefore, \grad with momentum is always better or the same as plain momentum. 

\section{Experiments}

\begin{figure}[t]
\centering
\includegraphics[width=0.245\textwidth]{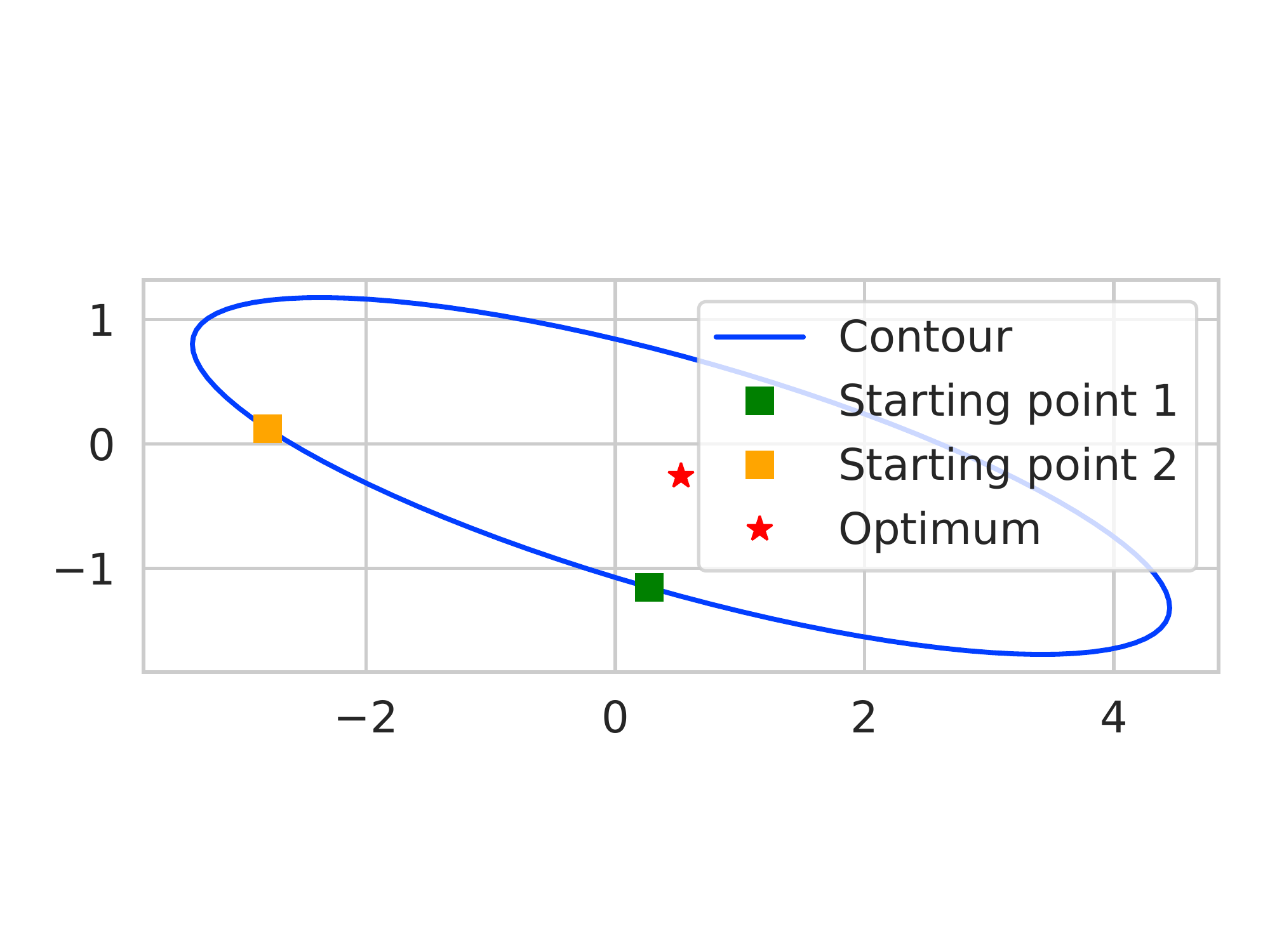}
\includegraphics[width=0.245\textwidth]{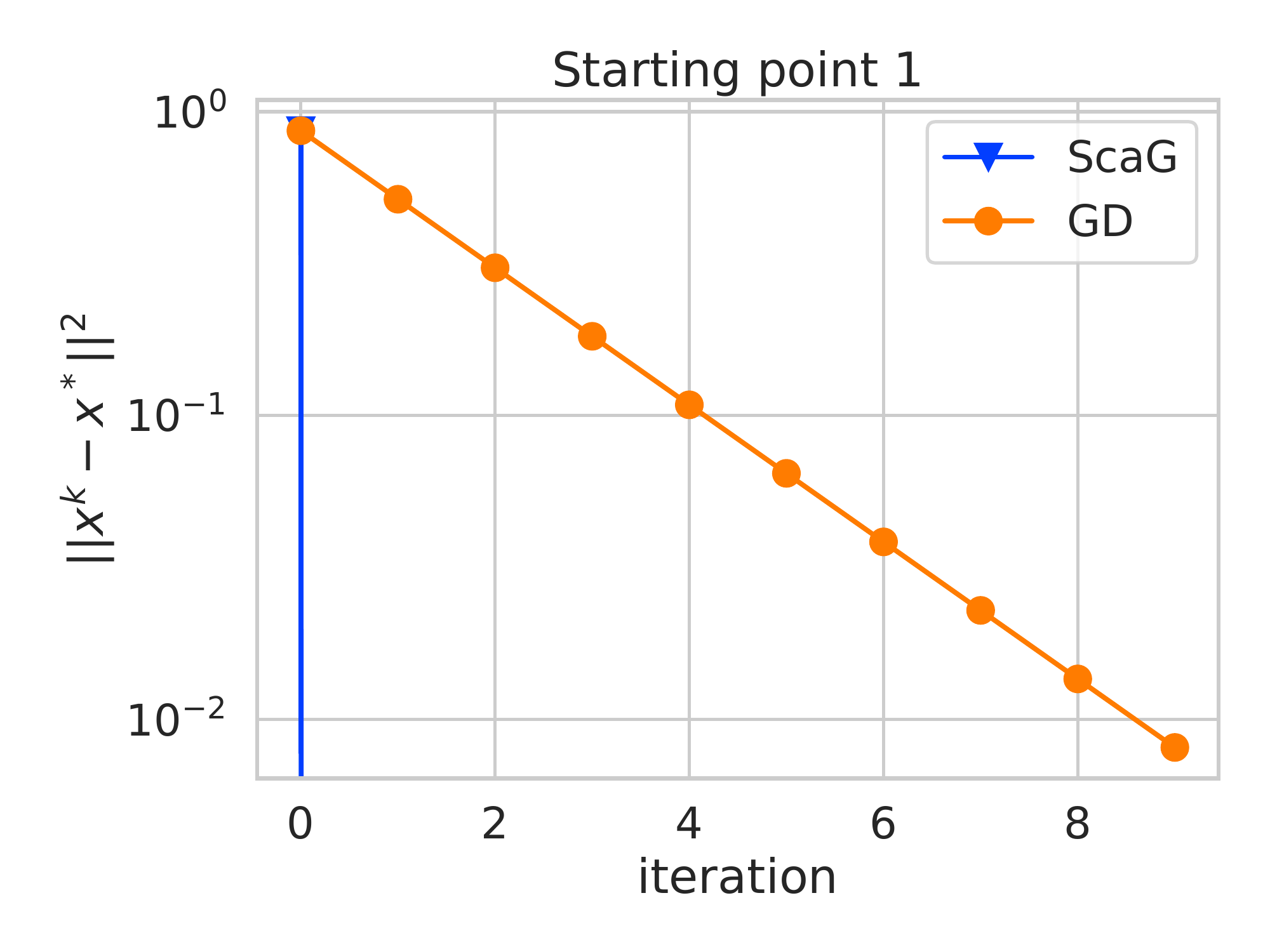}
\includegraphics[width=0.245\textwidth]{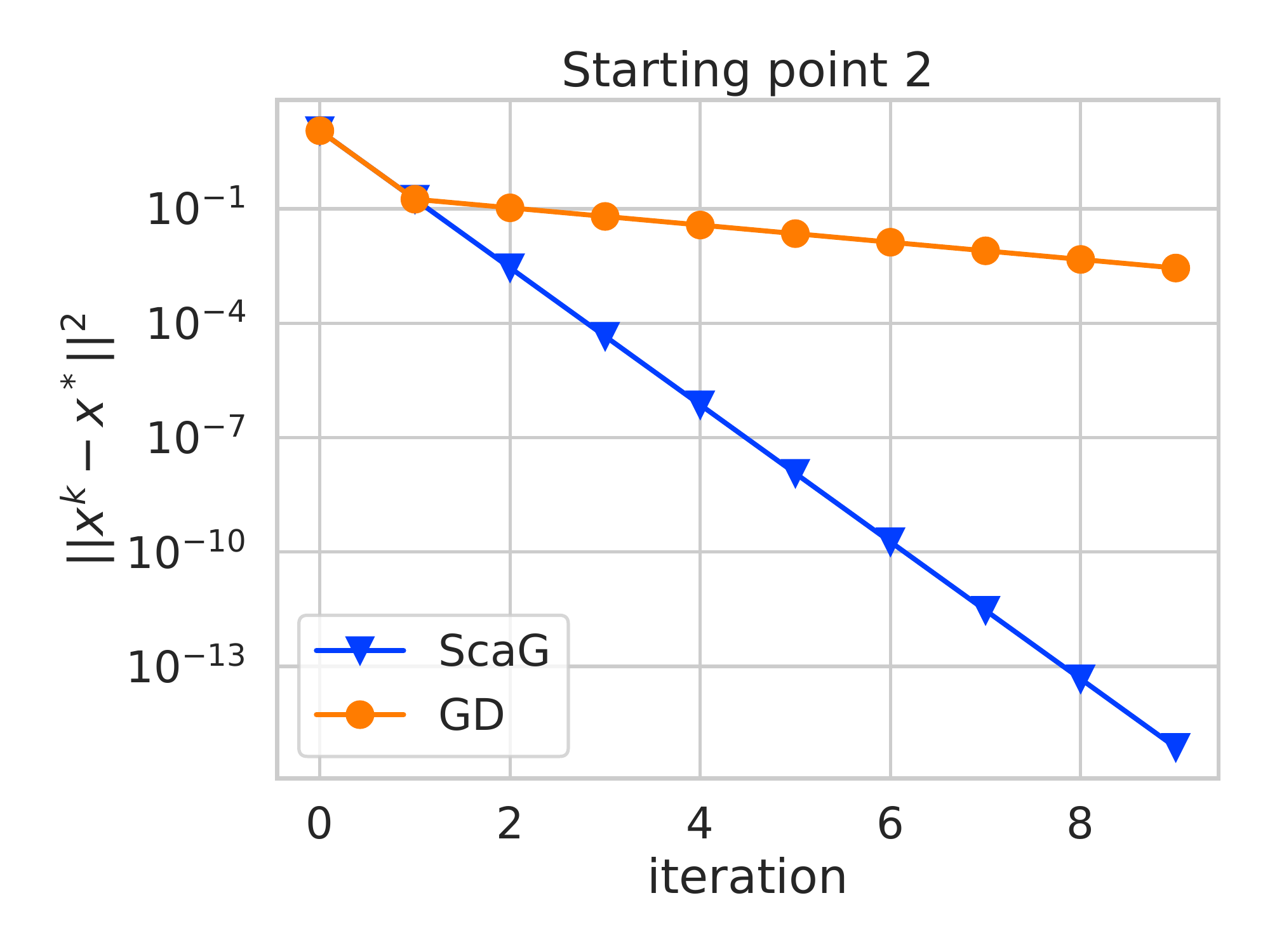}
\includegraphics[width=0.245\textwidth]{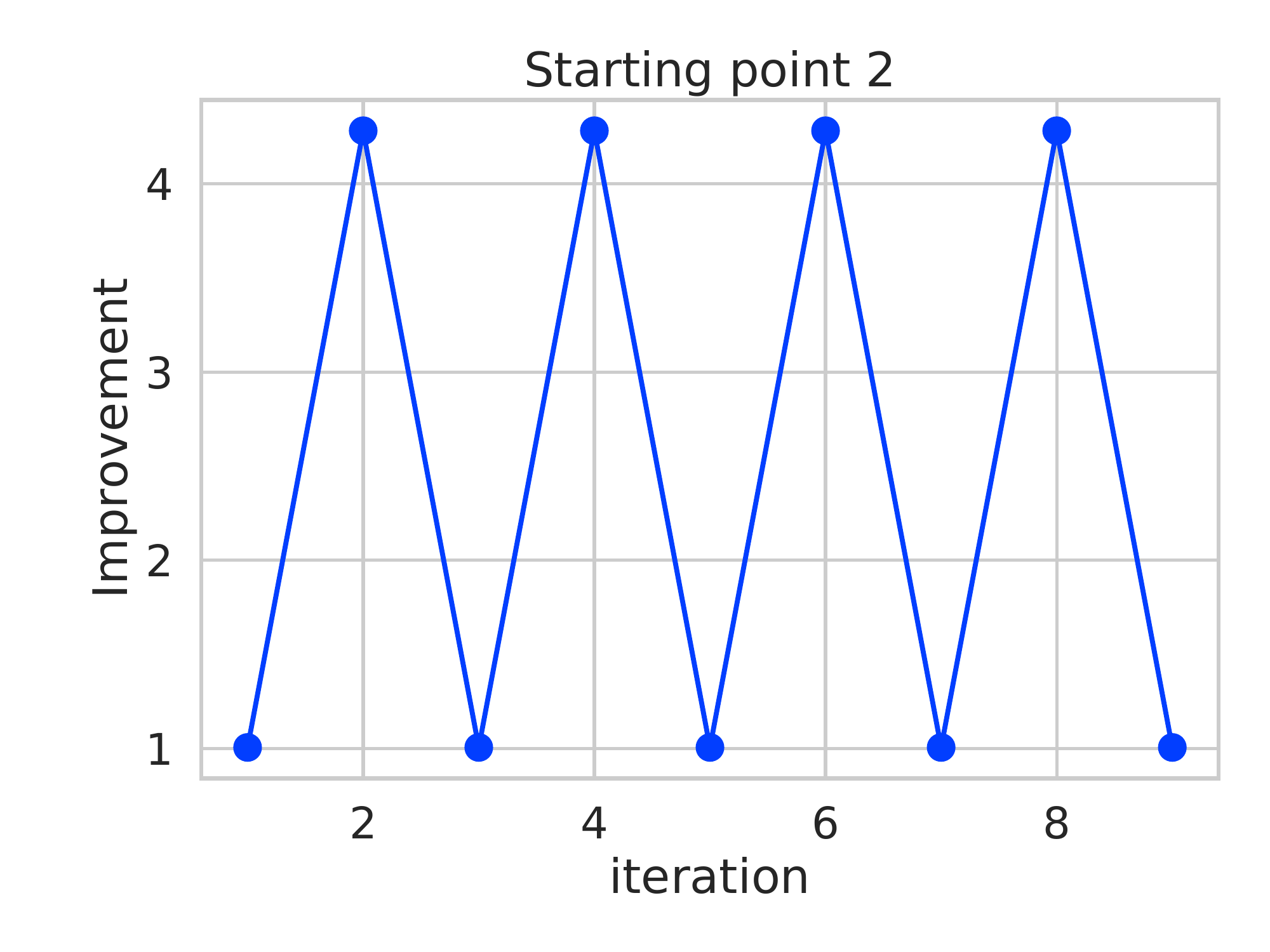}
\caption{Convergence for 2D linear regression. The left plot display a countour line with two starting points. The middle plots showcase the convergence for both of these starting point for the \texttt{GD} and \texttt{ScaG} (identical to \stps and \grads)  methods. The right plot displays the improvement factor which is defined as the multiplicative factor of \texttt{ScaG}'s with respect to \texttt{GD}'s step size.}
\label{fig:linear_2d}
\end{figure}

\begin{figure}[t]
\centering
\includegraphics[width=0.245\textwidth]{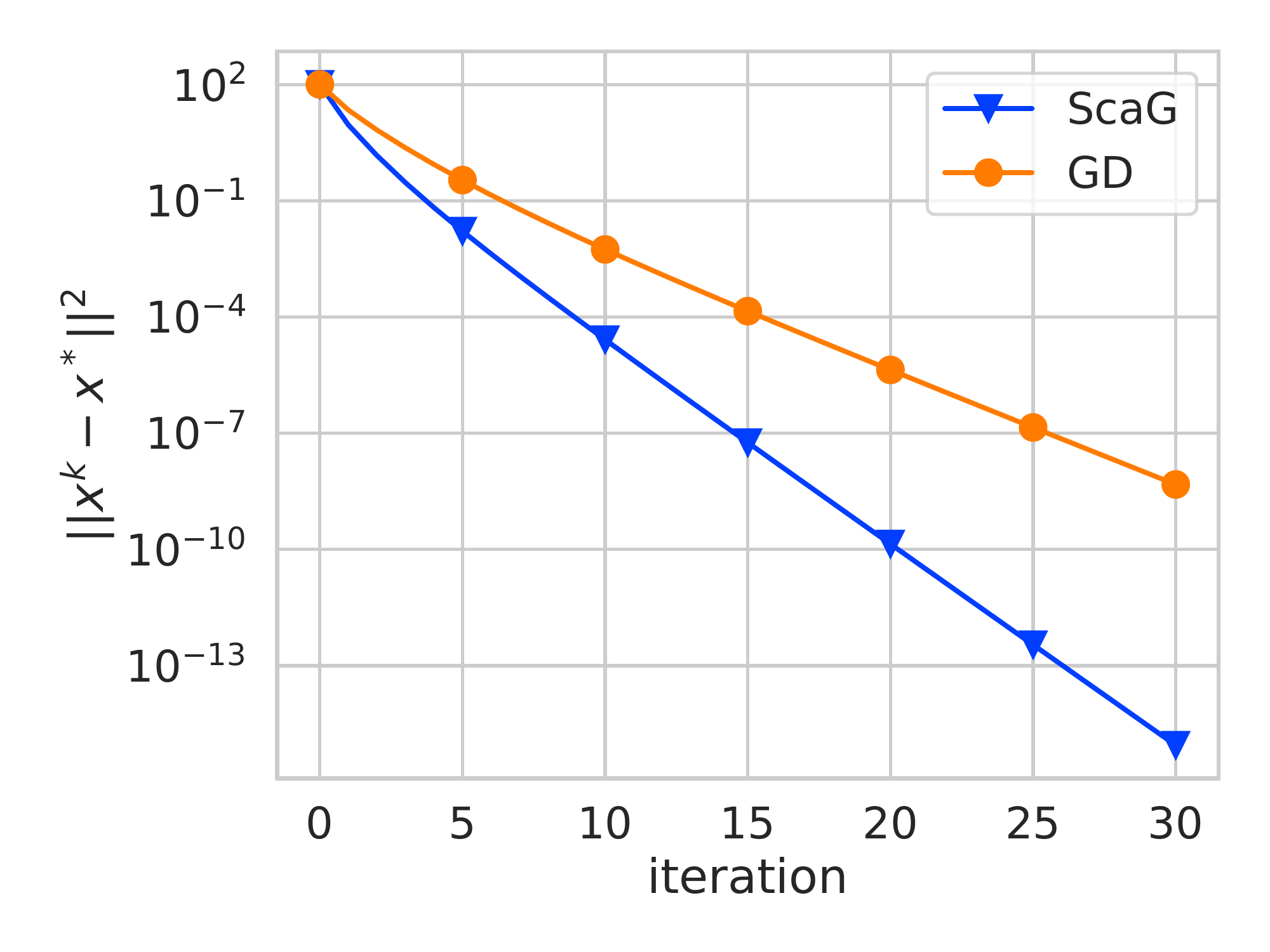}
\includegraphics[width=0.245\textwidth]{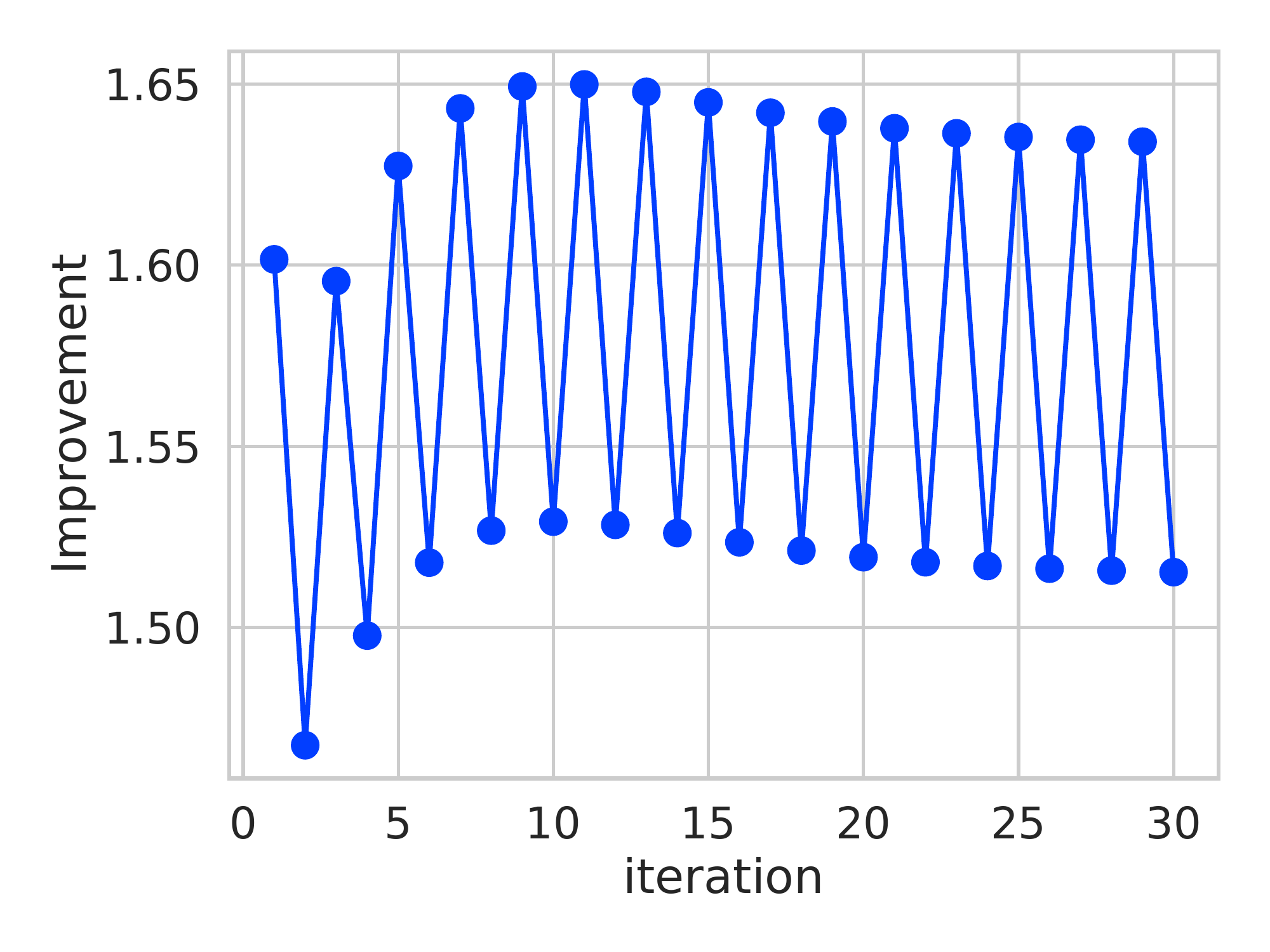}
\includegraphics[width=0.245\textwidth]{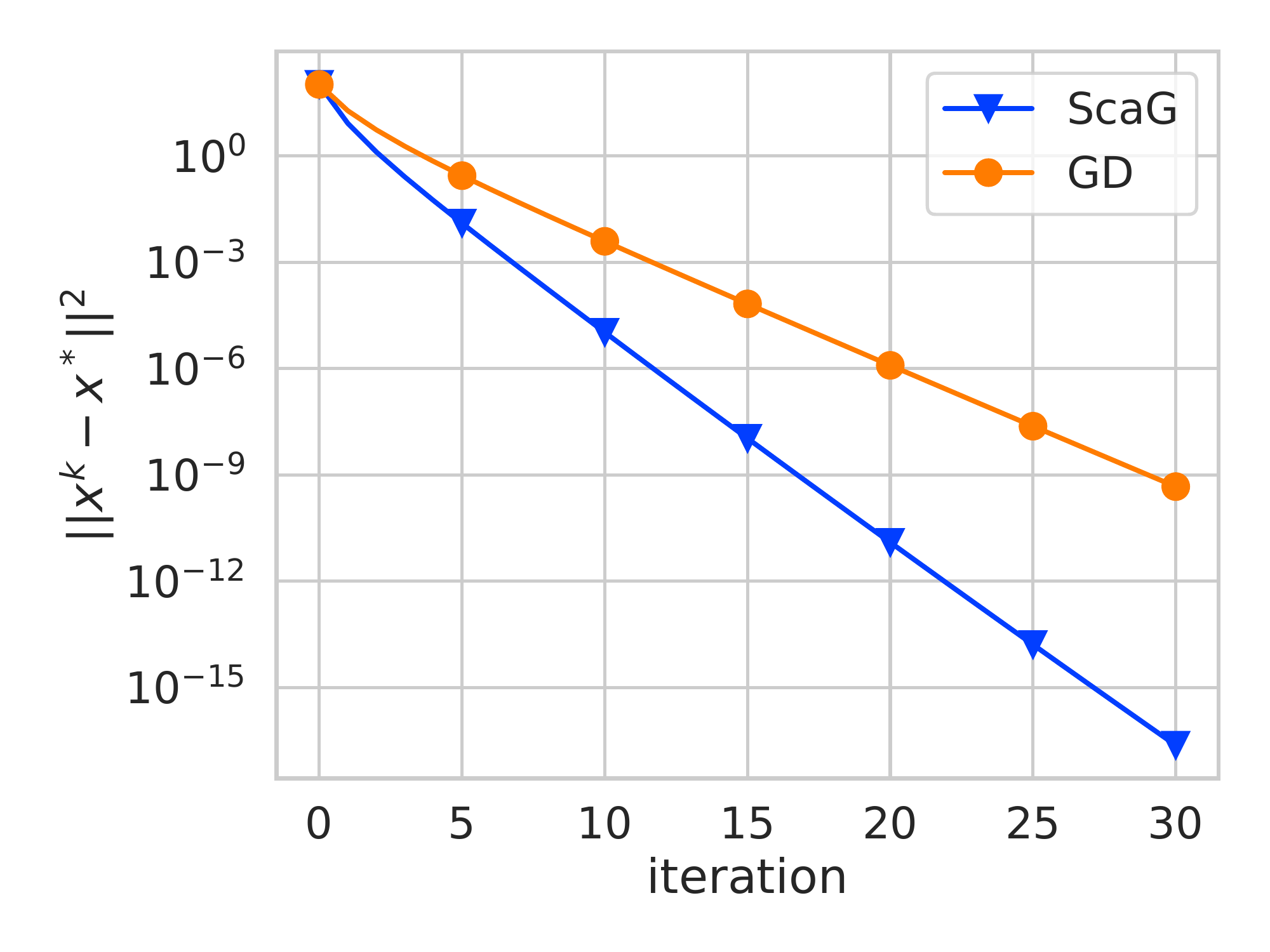}
\includegraphics[width=0.245\textwidth]{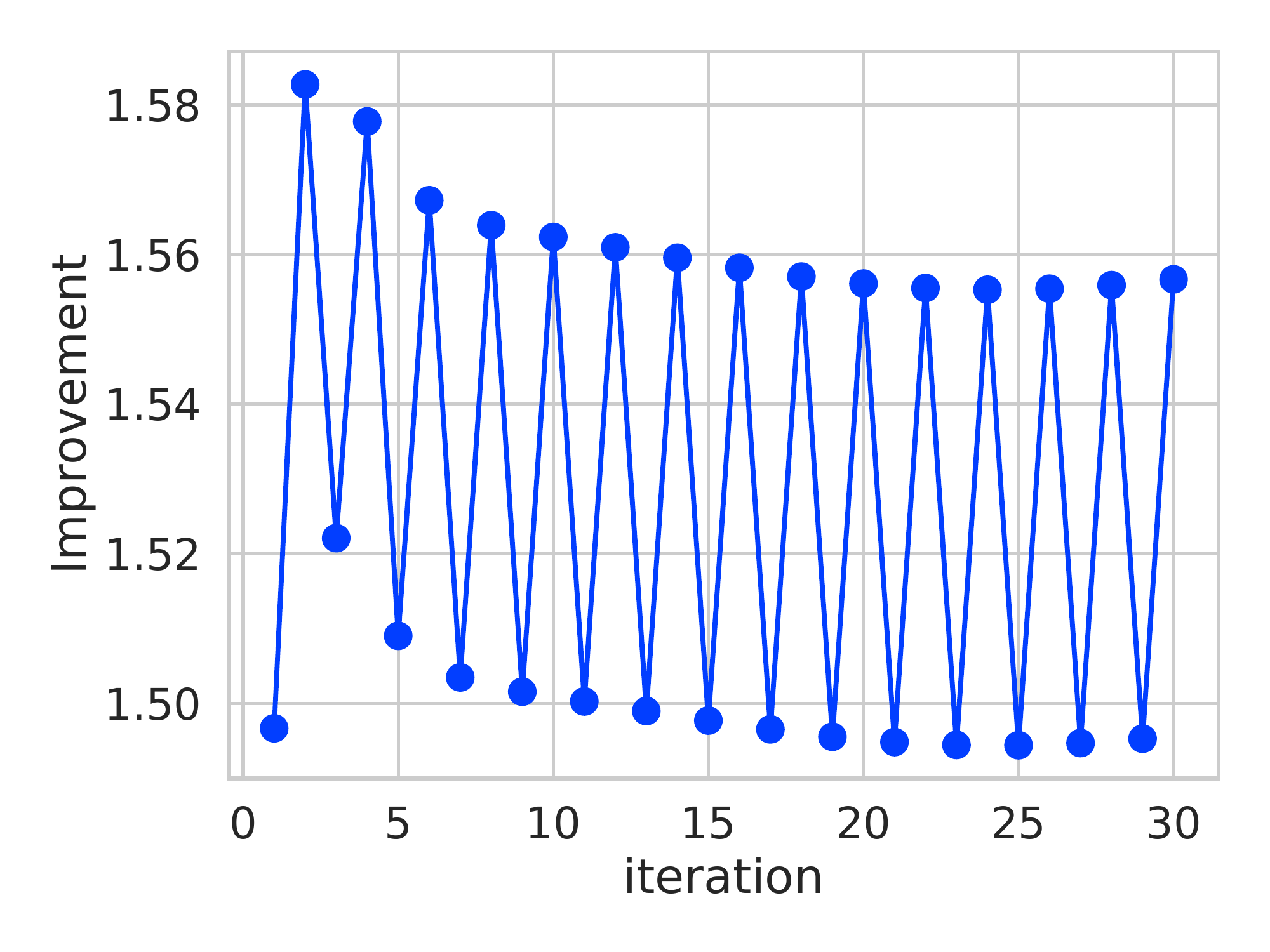}
\caption{Convergence with an improvement factor for linear regression for the \texttt{GD} and \texttt{ScaG} (identical to \stps and \grads) methods with synthetic data of the size $100 \times 1000$. Each pair of plots corresponds to randomly generated data matrix. The improvement factor is defined in the above figure.}
\label{fig:linear_100d}
\end{figure}

\subsection{Consistent Linear Systems: Synthetic Experiment}
For the first experiment, we demonstrate the theoretical superiority of \stps and \grads on consistent linear systems. As previously mentioned, \stps and \grads are the same methods in this scenario and we refer to them as scaled gradient method (\texttt{ScaG}). The main advantage is that for these systems all the theoretical parameters can be exactly computed and therefore there is no need for tuning. 

Firstly, we randomly generated $5 \times 2$ data matrix $A$, where each entry is sampled from normal distribution with zero mean and unit variance. We normalize each row of matrix $A$ and run \texttt{ScaG} with full gradient with step size defined in \eqref{eq:stps_geq_gd} and compare it to Gradient \texttt{GD} with the step size $\nicefrac{1}{L_f} = \nicefrac{n}{\ns{A^\top A}}$. In Figure~\ref{fig:linear_2d}, we display contour plot in the left where we show starting points that we consider for the optimization, convergence plots in the middle and the improvement factor that we compute as the ratio of the current \texttt{ScaG} and \texttt{GD} step sizes on the right. As predicted by theory, if $x^0 - \xs$ corresponds to the smallest eigenvalue \texttt{ScaG} converges in one step while \texttt{GD} does not. In addition, \texttt{ScaG} allows for larger step sizes and, therefore, faster convergence.

Secondly, we repeat the same construction for larger matrices of the size $100 \times 1000$ and we display both convergence and the improvement factor. As a starting point, we select all zeros vector. As for the $2$-D case, we can conclude that our adaptive method leads to a significant boost in the convergence speed; see Figure~\ref{fig:linear_100d}.

\begin{table}[t]
\centering
\caption{Final validation accuracy of LeNet on MNIST.}
\label{tab:lenet_mnist}
\begin{tabular}{|c|c|c|}
\hline
 & \texttt{SGD} & \texttt{SGDM}              \\ \hline
\texttt{None}      & $98.44 \pm 0.12$ & $98.57 \pm 0.04$ \\ \hline
\stp      & $93.67 \pm 2.93$ & $92.74 \pm 1.24$  \\ \hline
\grad      & ${\color{red}\mathbf{98.88}} \pm 0.01$ & ${\color{red}\mathbf{98.88}} \pm 0.09$  \\ \hline
\end{tabular}
\end{table}

\begin{figure}[t]
\centering
\includegraphics[width=0.32\textwidth]{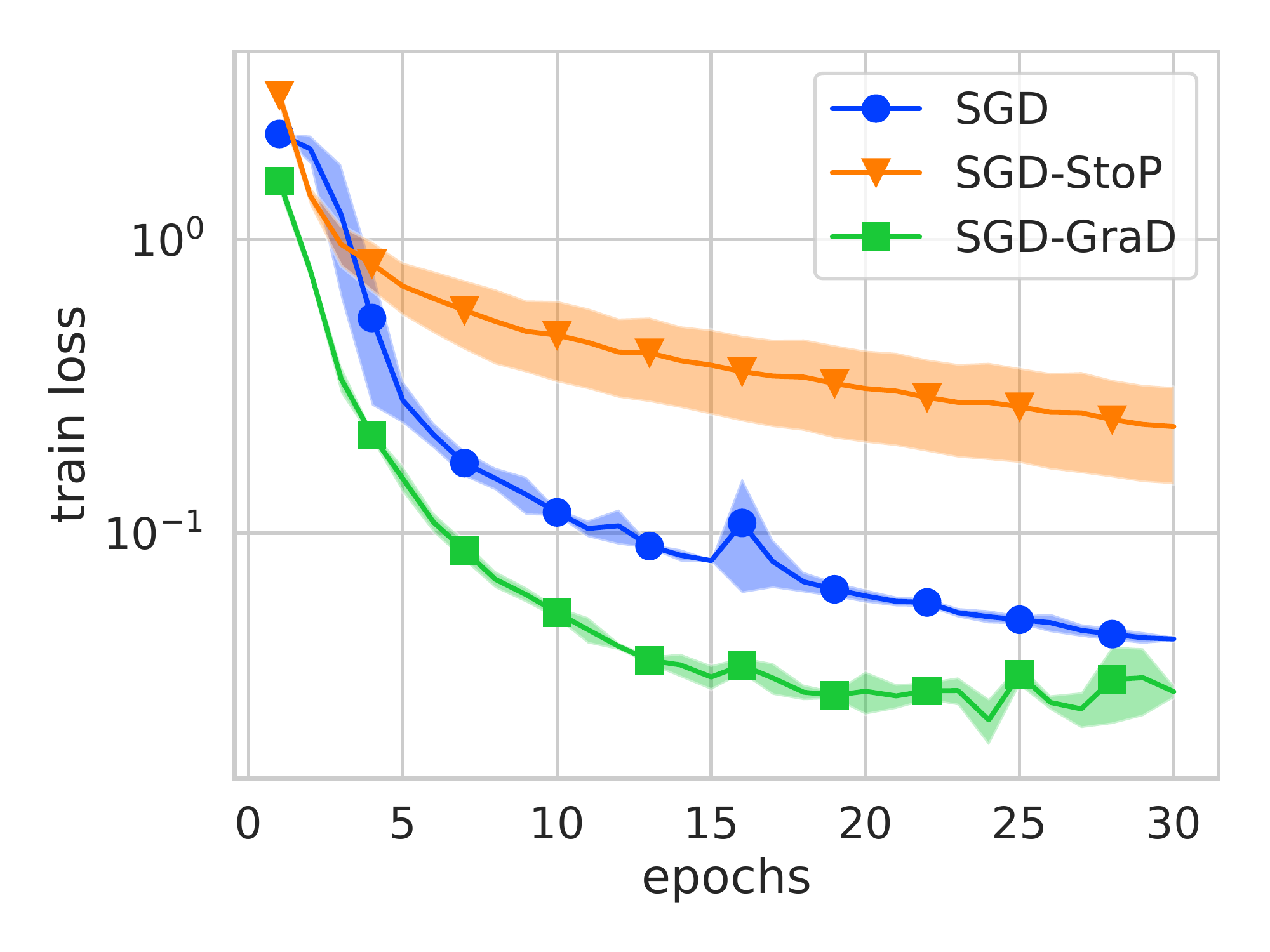}
\includegraphics[width=0.32\textwidth]{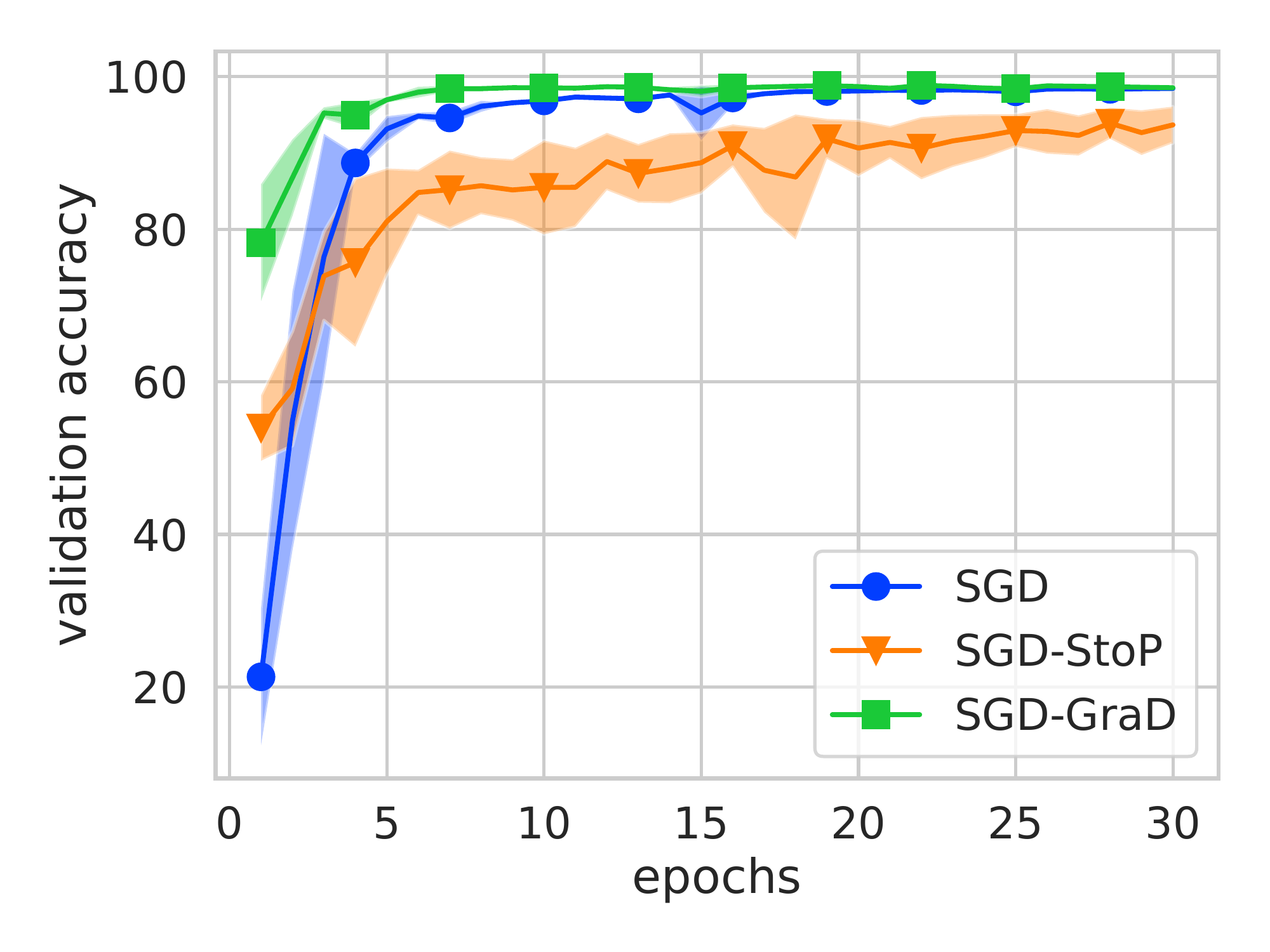}
\includegraphics[width=0.32\textwidth]{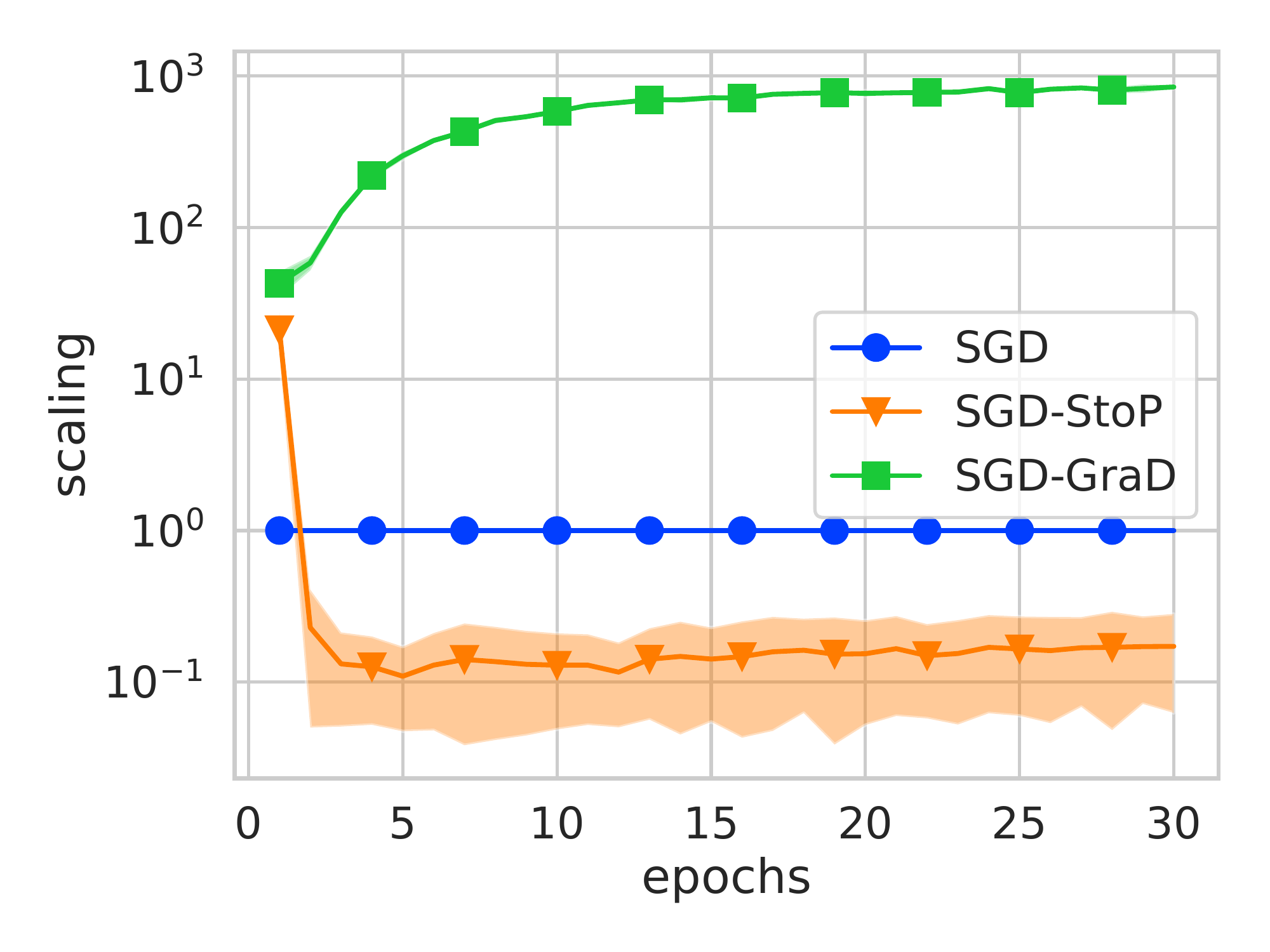}
\includegraphics[width=0.32\textwidth]{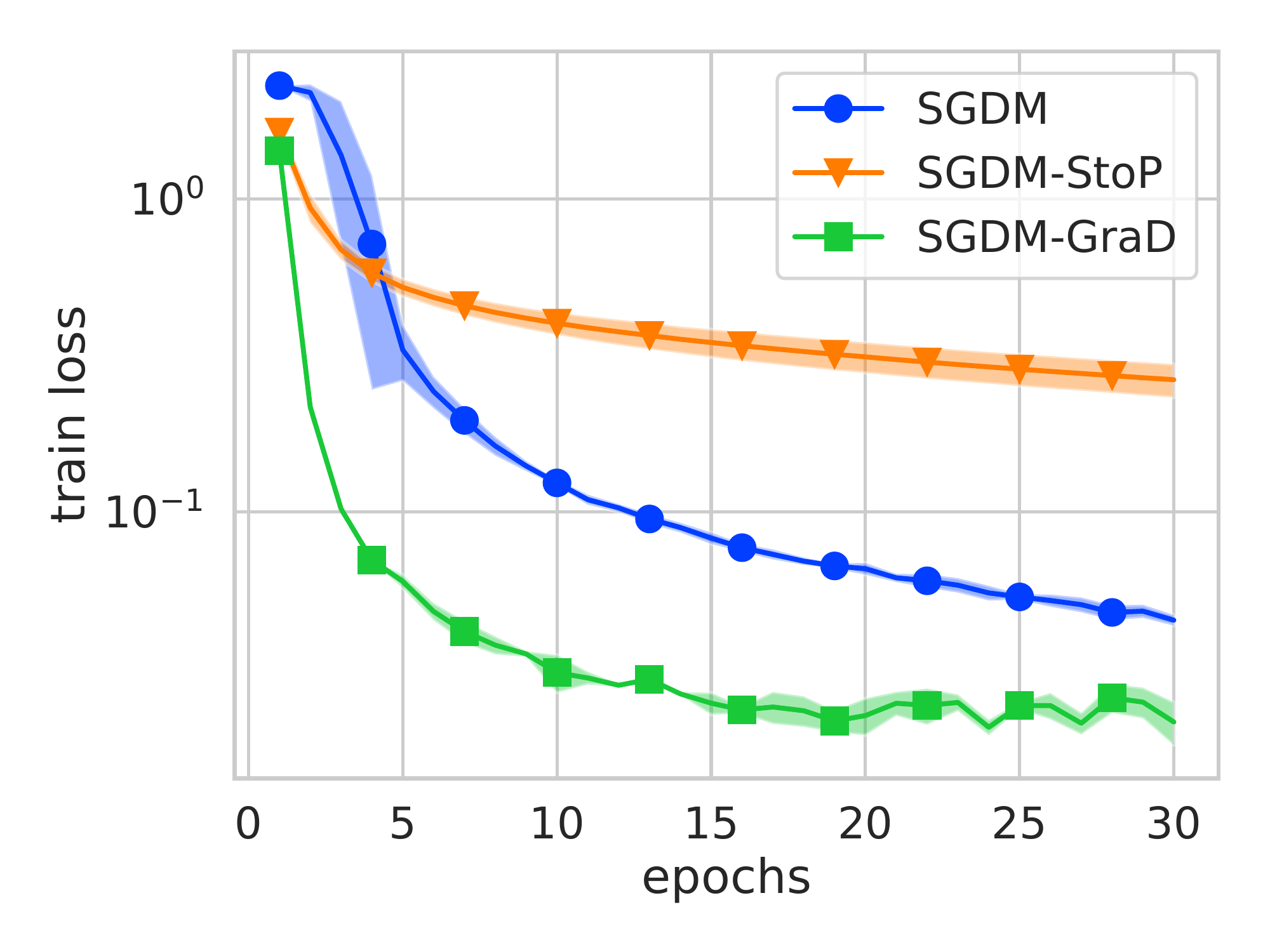}
\includegraphics[width=0.32\textwidth]{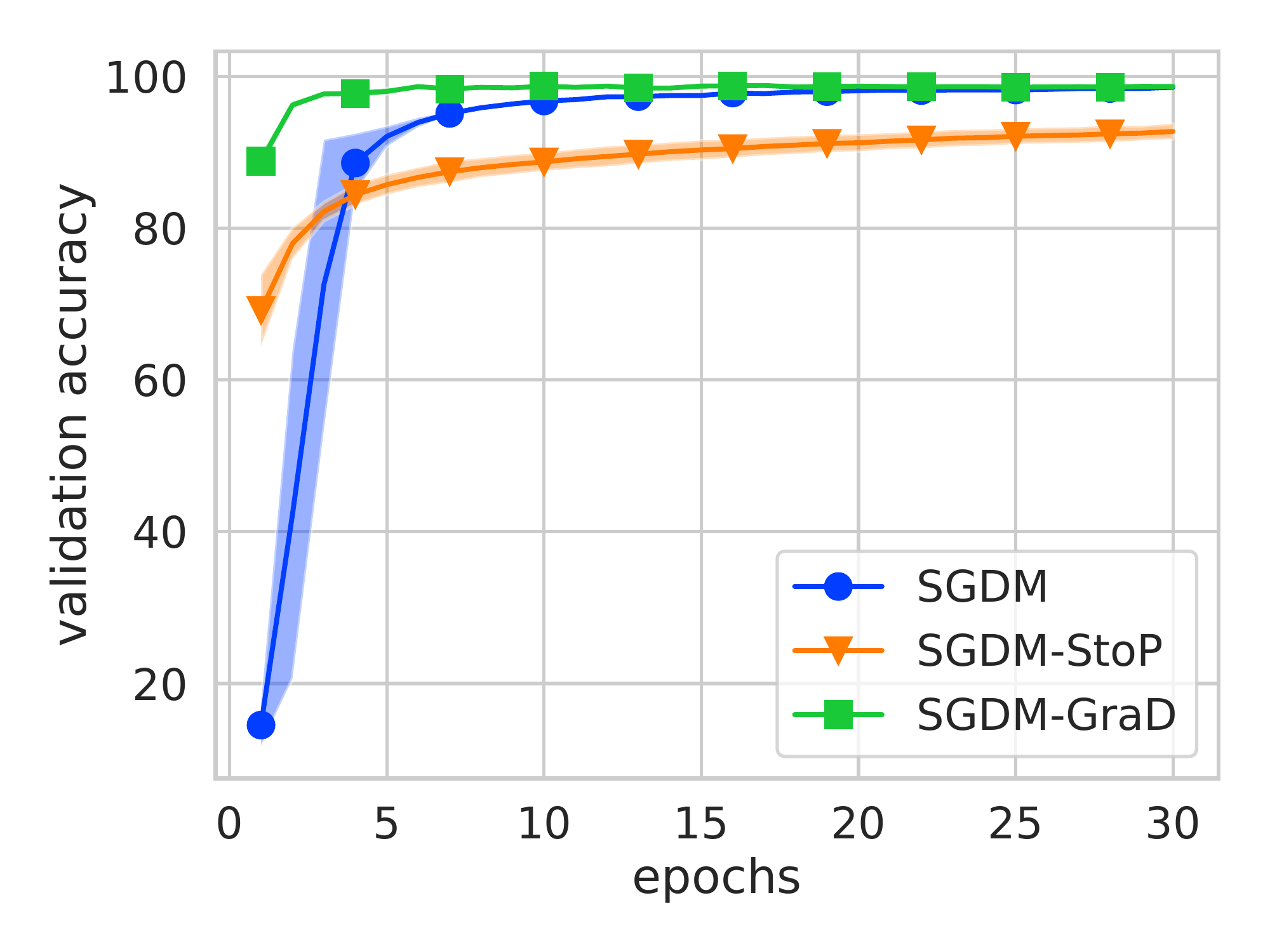}
\includegraphics[width=0.32\textwidth]{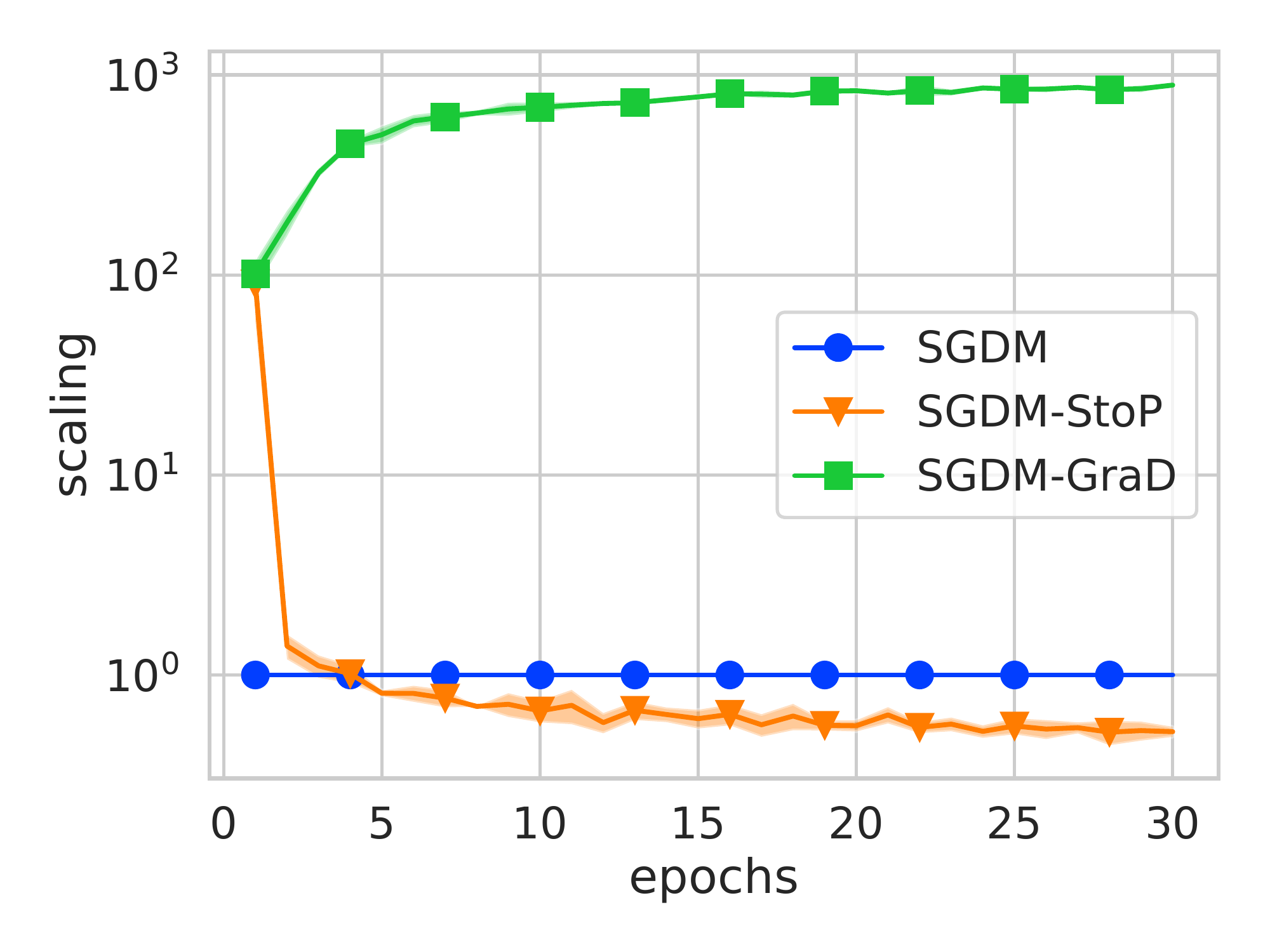}
\caption{LeNet on MNIST. Left: training loss corresponding to the step size with the minimum final average loss. Middle: validation accuracy corresponding to the step size with the maximum final average validation accuracy. Right: train scale ( 1 for plain baseline, step size scaling for \grad and actual step size for \stp).}
\label{fig:lenet_mnist}
\end{figure}

\begin{table}[t]
\centering
\caption{Final validation accuracy of ResNet20 on CIFAR10.}
\label{tab:resnet20_cifar10}
\begin{tabular}{|c|c|c|c|}
\hline
 & \texttt{SGD} & \texttt{SGDM}         & \texttt{Adam}         \\ \hline
\texttt{None}      & $88.18 \pm 0.63$ & $89.34 \pm 0.07$ & $\mathbf{90.39} \pm 0.33$ \\ \hline
\stp      & $89.38 \pm 0.20$ & $90.13 \pm 0.23$ & $89.81 \pm 0.38$ \\ \hline
\grad      & $\mathbf{89.72} \pm 0.29$ & ${\color{red}\mathbf{91.11}} \pm 0.14$ & $89.98 \pm 0.13$ \\ \hline
\end{tabular}
\end{table}

\begin{figure}[t]
\centering
\includegraphics[width=0.32\textwidth]{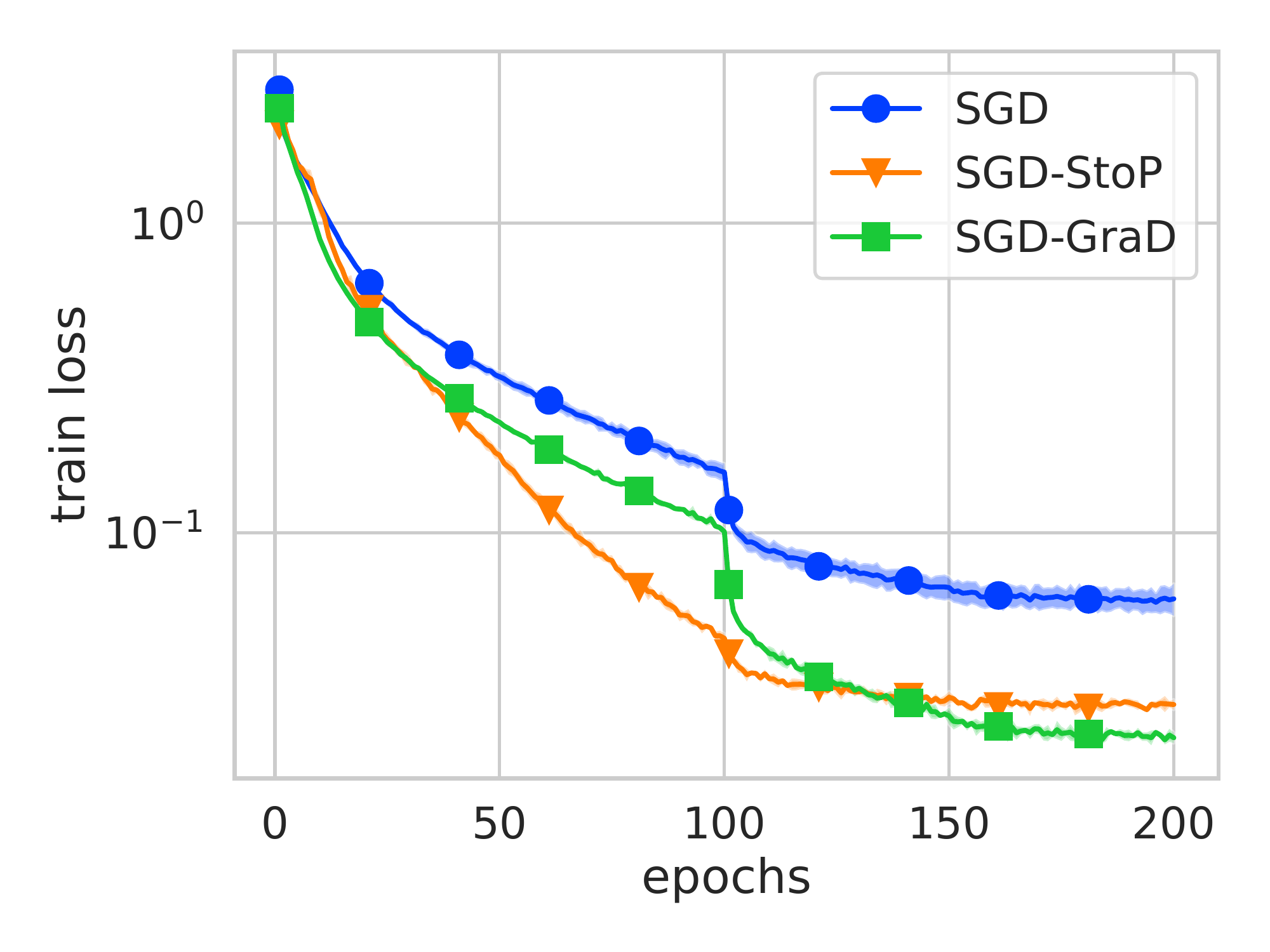}
\includegraphics[width=0.32\textwidth]{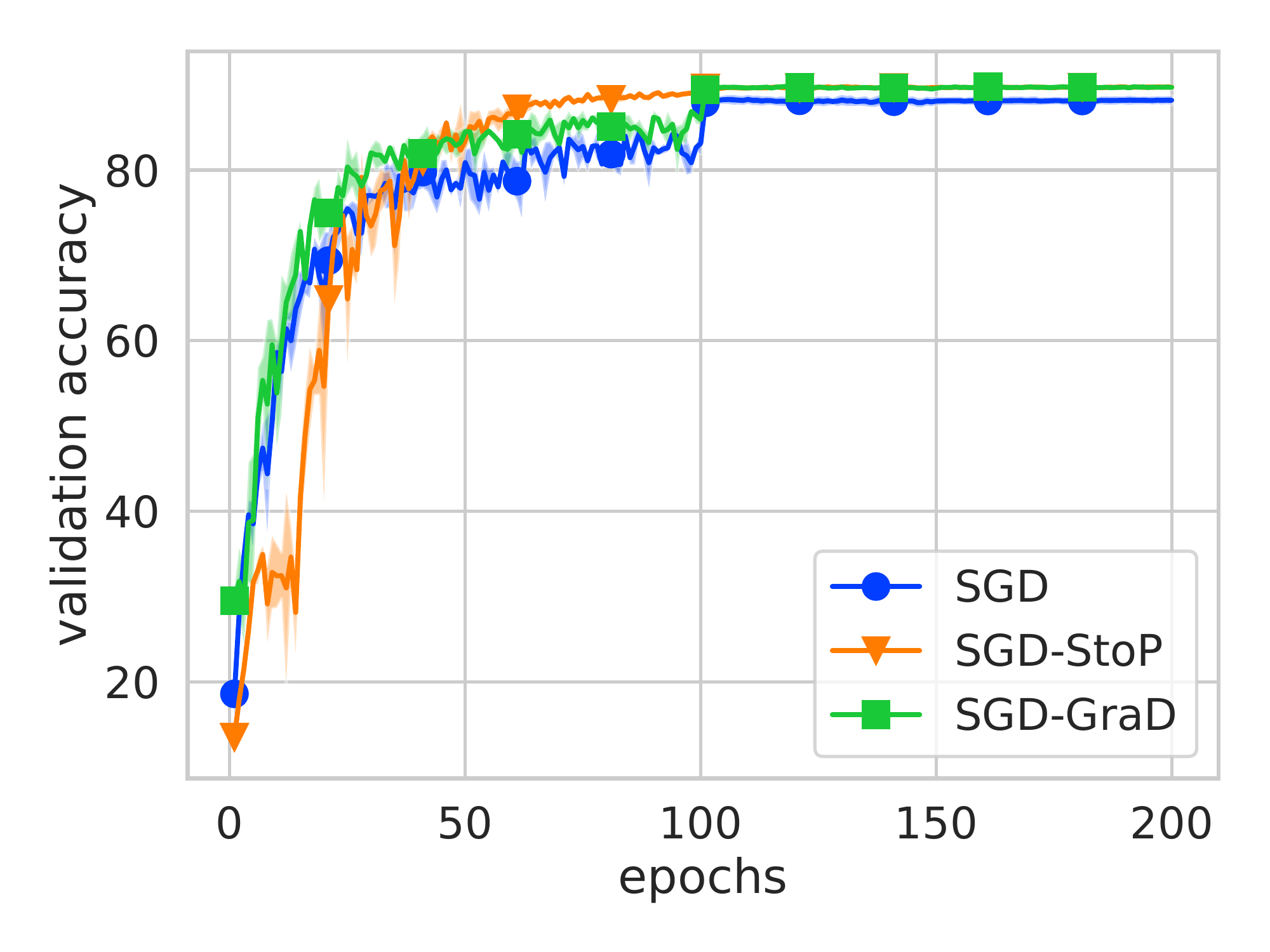}
\includegraphics[width=0.32\textwidth]{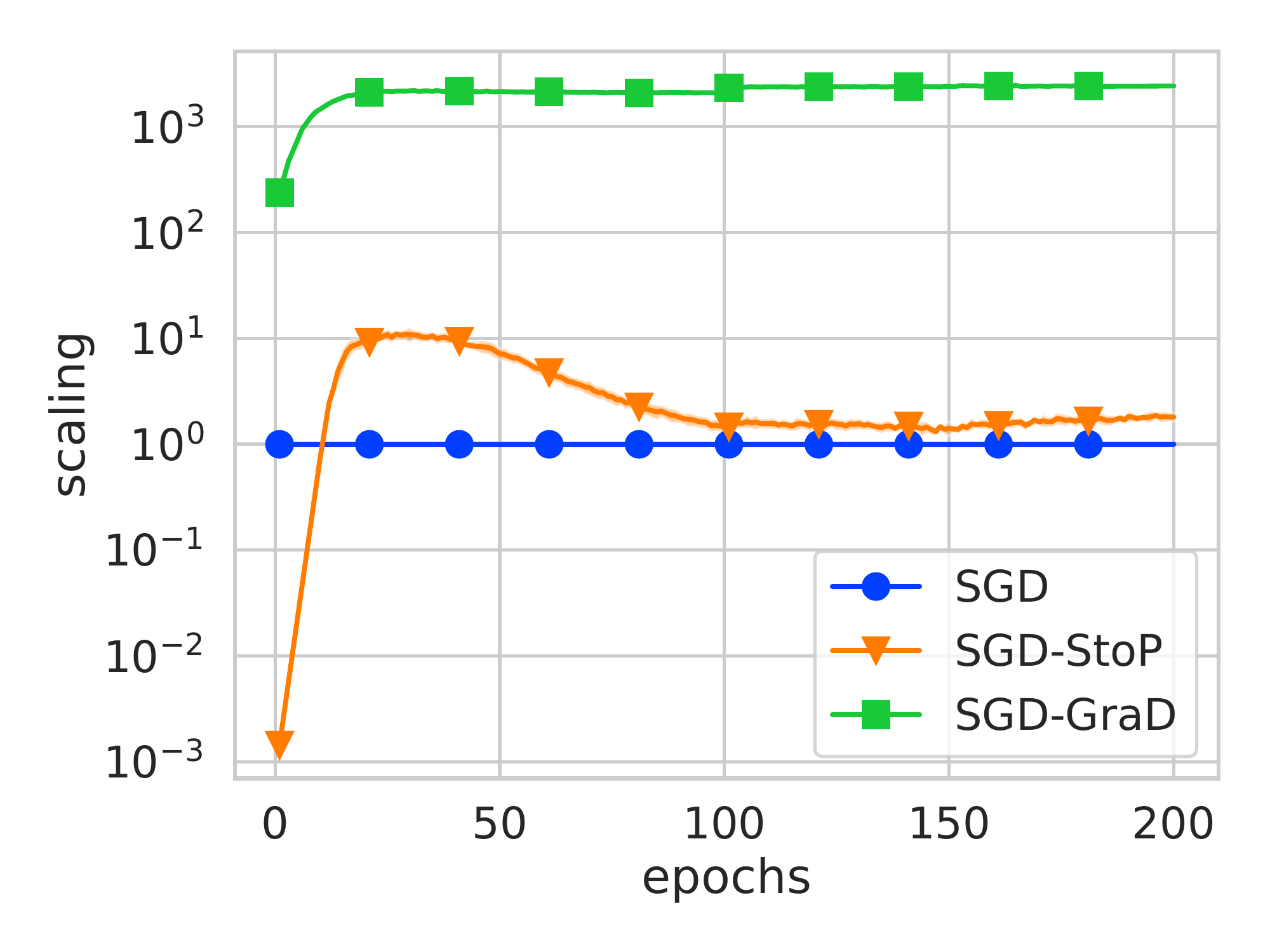}
\includegraphics[width=0.32\textwidth]{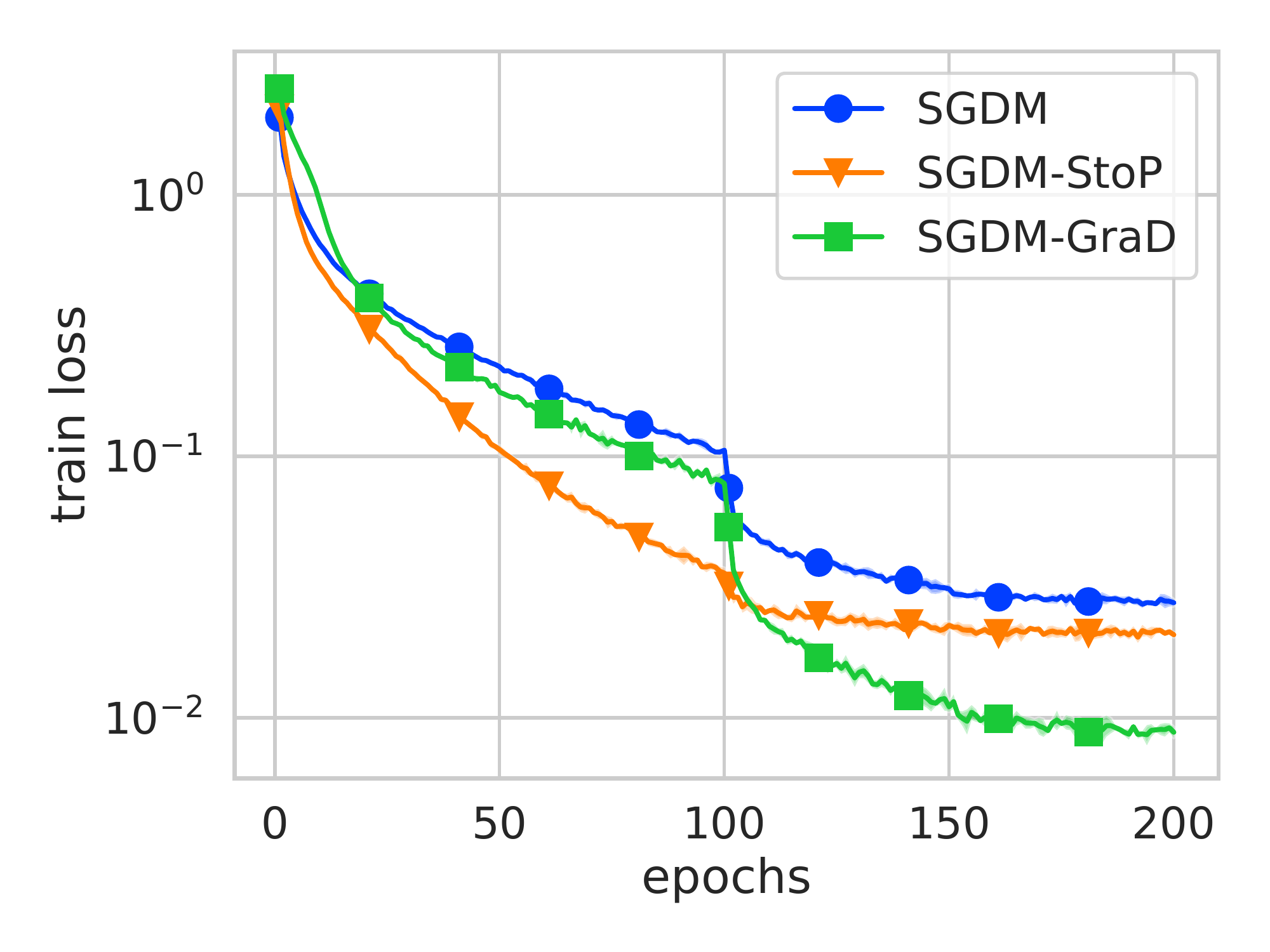}
\includegraphics[width=0.32\textwidth]{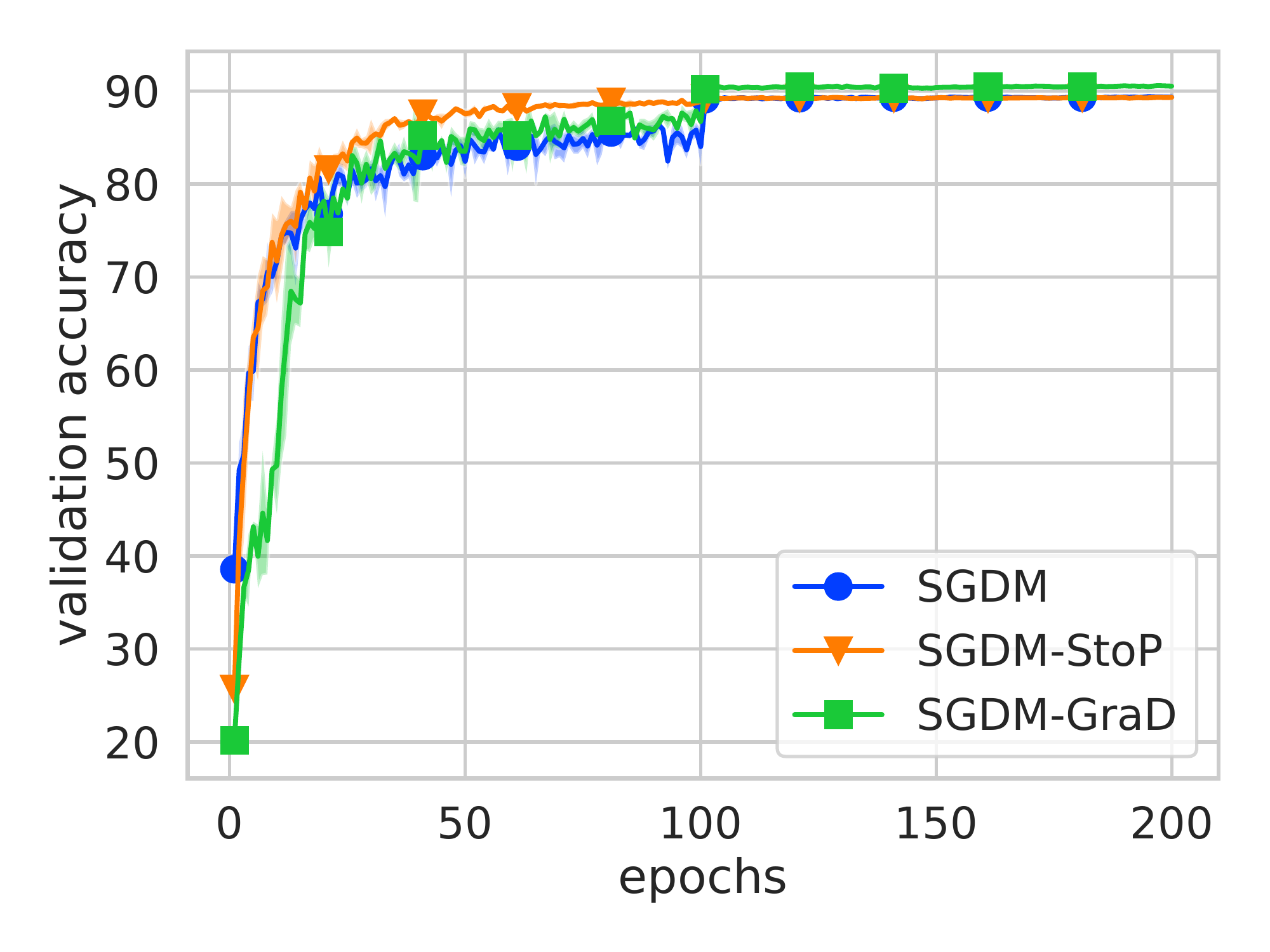}
\includegraphics[width=0.32\textwidth]{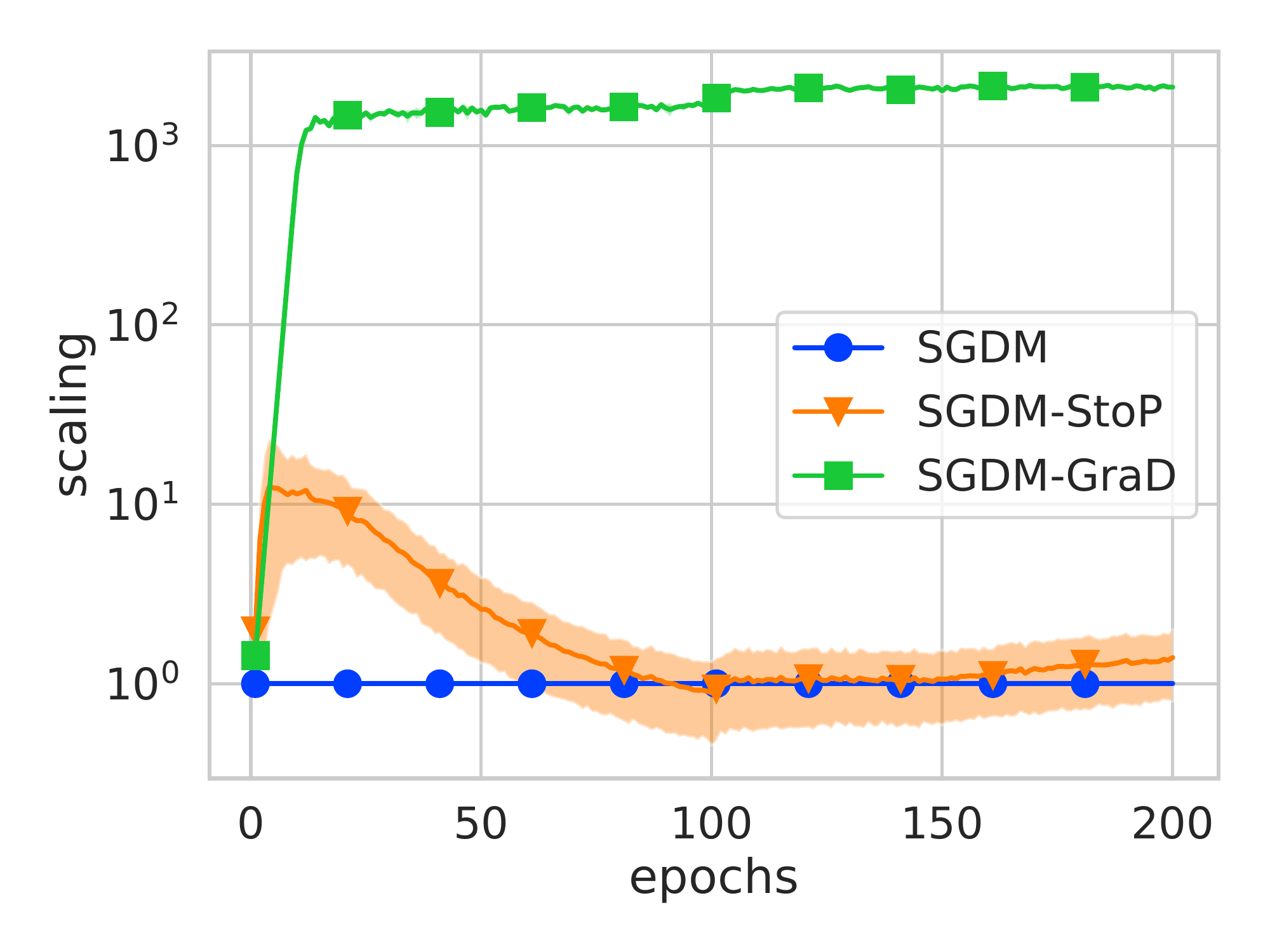}
\includegraphics[width=0.32\textwidth]{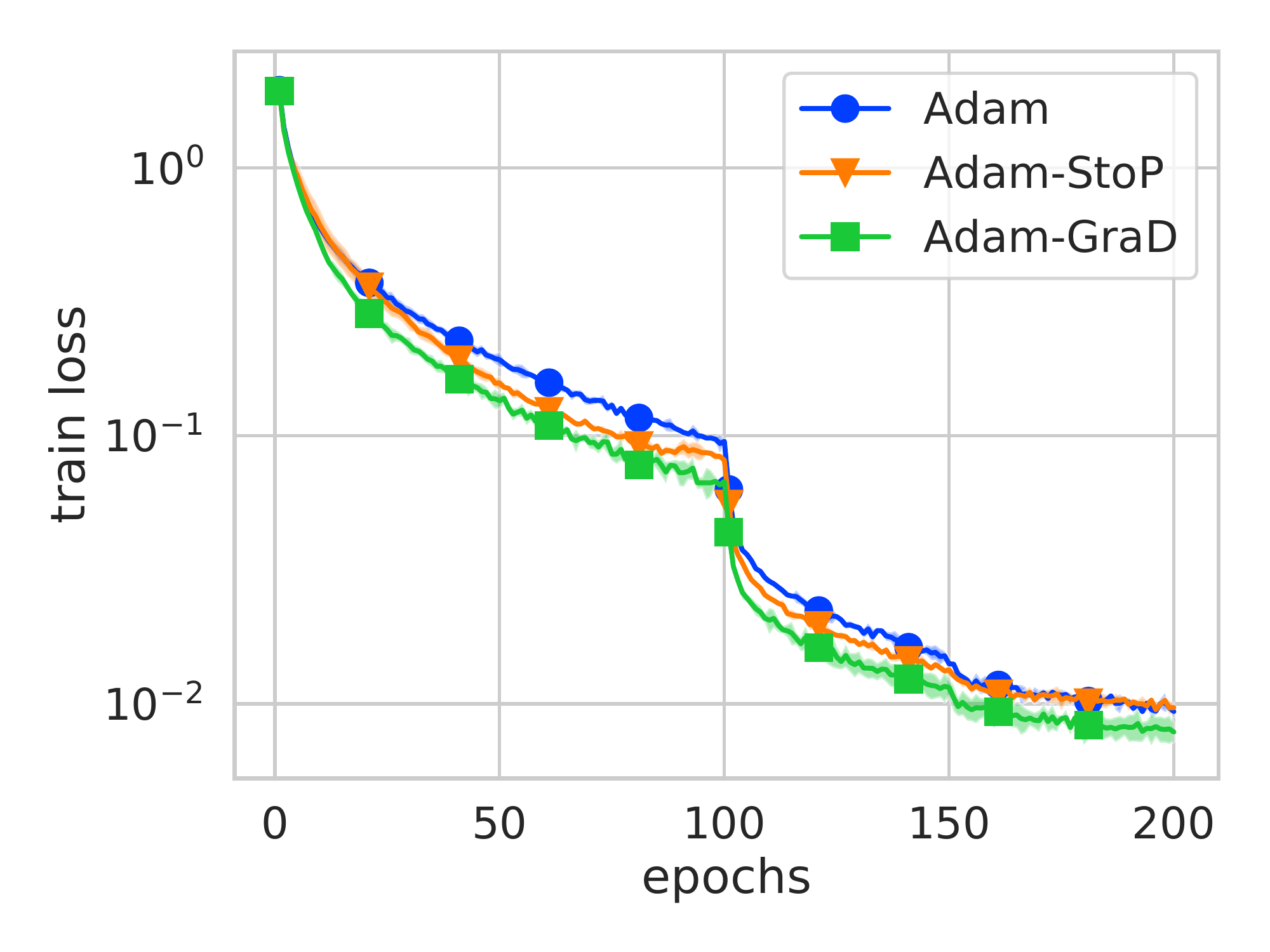}
\includegraphics[width=0.32\textwidth]{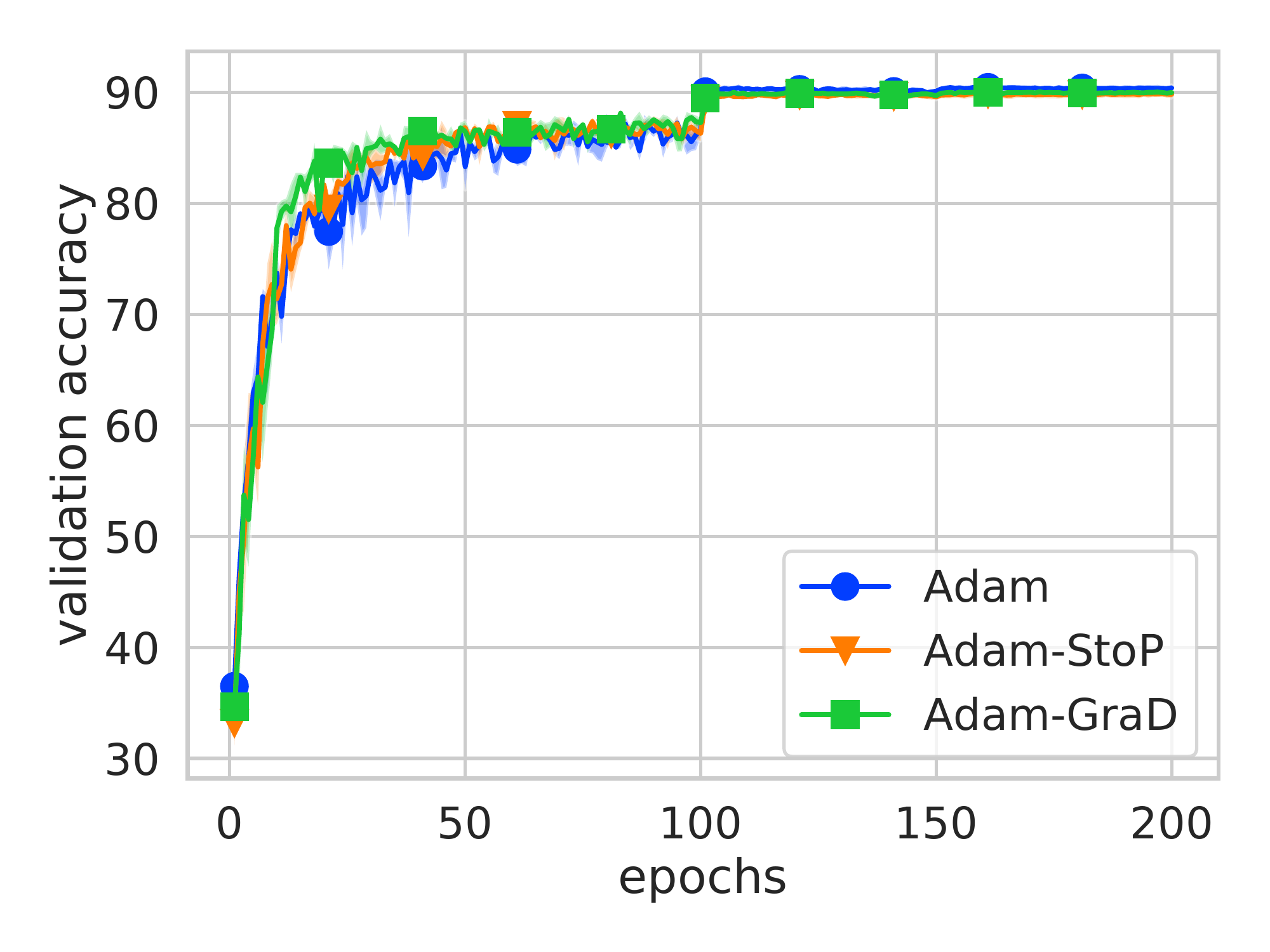}
\includegraphics[width=0.32\textwidth]{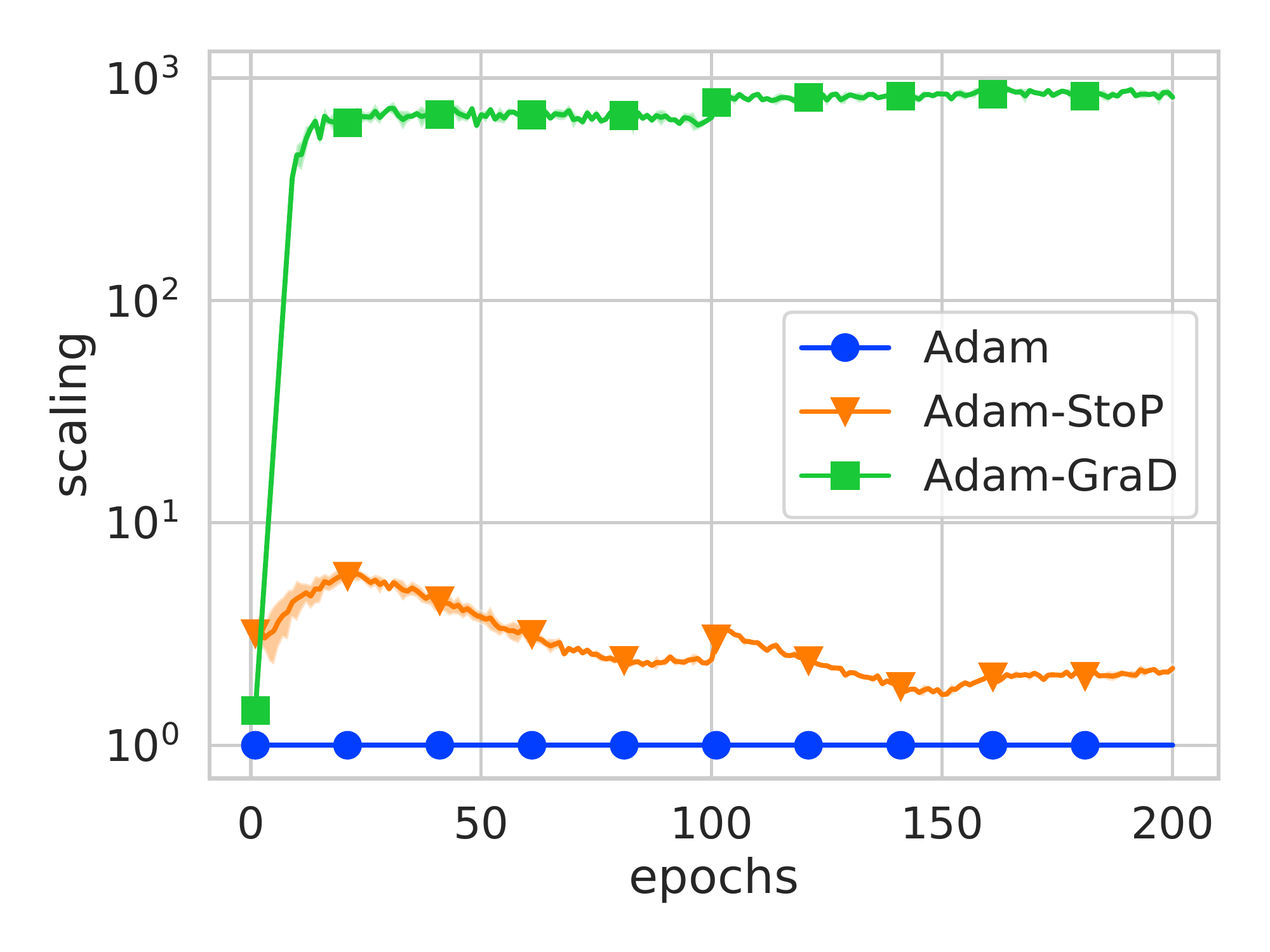}
\caption{ResNet20 on CIFAR10. For details see caption in Figure~\ref{fig:lenet_mnist}.}
\label{fig:resnet20_cifar10}
\end{figure}

\begin{table}[t]
\centering
\caption{Final validation accuracy of ResNet32 on CIFAR100.}
\label{tab:resnet32_cifar100}
\begin{tabular}{|c|c|c|c|}
\hline
 & \texttt{SGD} & \texttt{SGDM}         & \texttt{Adam}         \\ \hline
\texttt{None}      & $61.86 \pm 0.53$ & $64.07 \pm 0.72$ & $63.13 \pm 0.73$ \\ \hline
\stp      & $61.90 \pm 0.47$ & $54.18 \pm 0.29$ & $62.50 \pm 1.32$ \\ \hline
\grad      & $\mathbf{65.14} \pm 0.55$ & ${\color{red}\mathbf{66.14}} \pm 0.11$ & $\mathbf{64.02} \pm 0.30$ \\ \hline
\end{tabular}
\end{table}

\begin{figure}[t]
\centering
\includegraphics[width=0.32\textwidth]{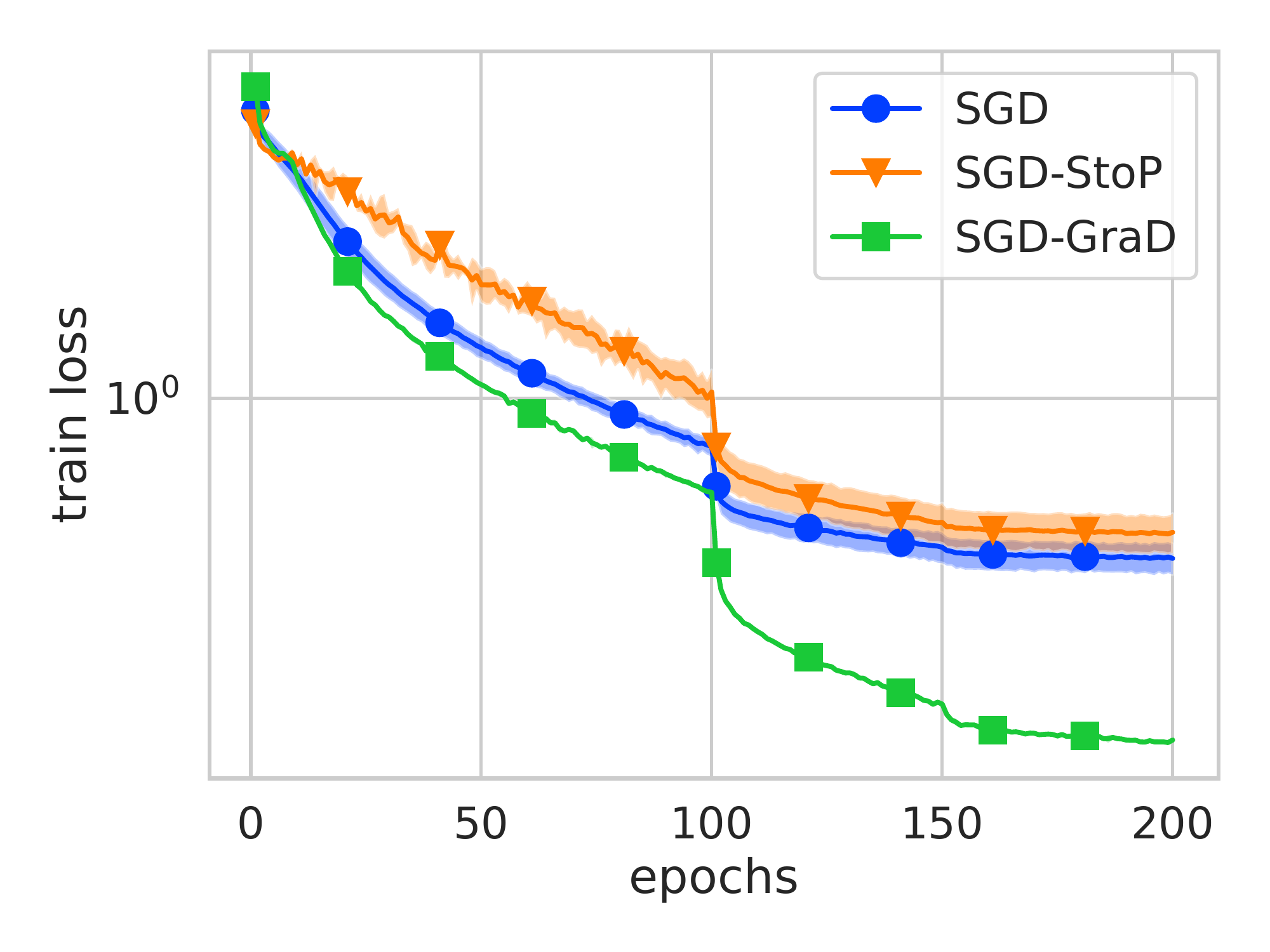}
\includegraphics[width=0.32\textwidth]{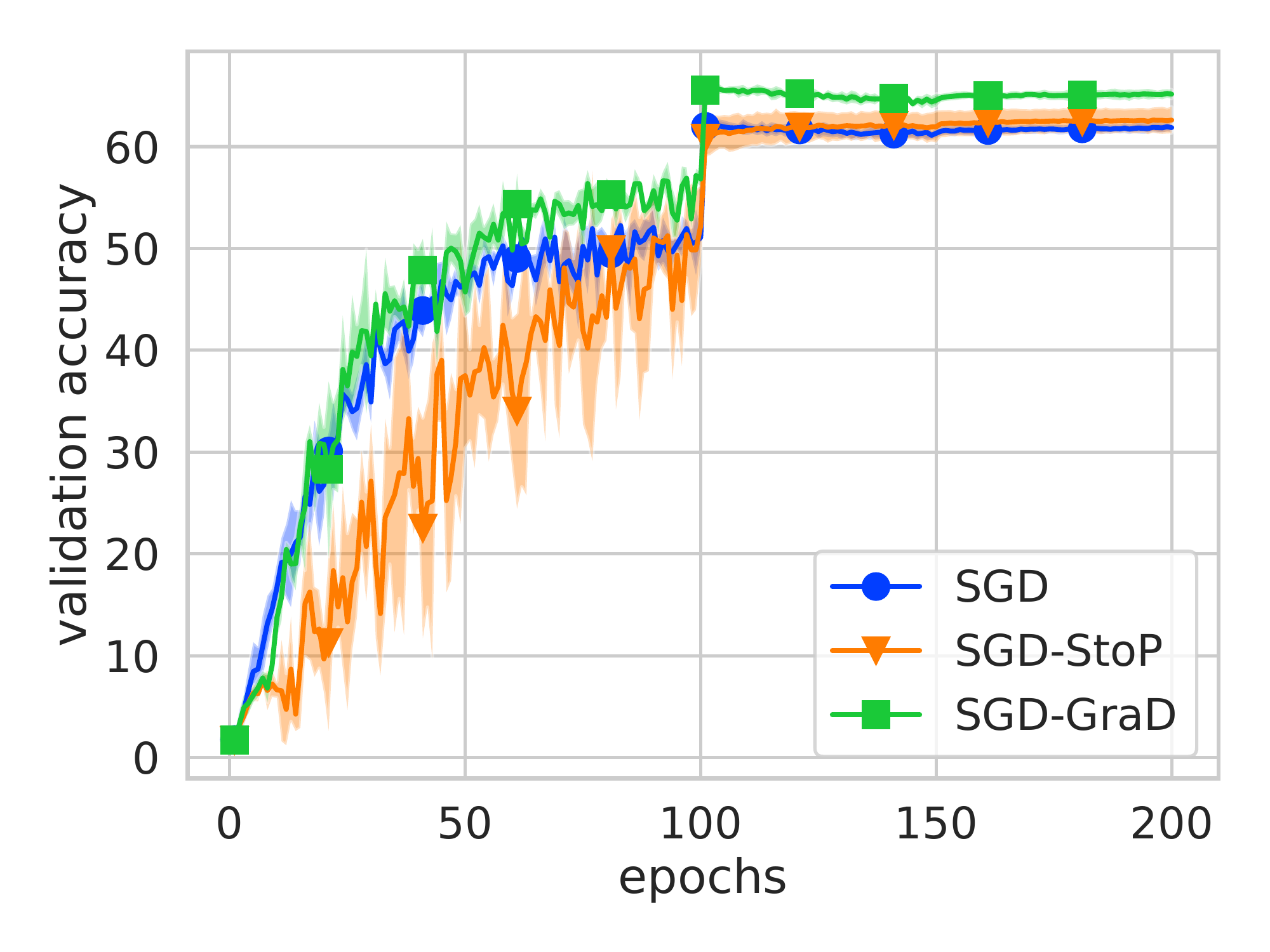}
\includegraphics[width=0.32\textwidth]{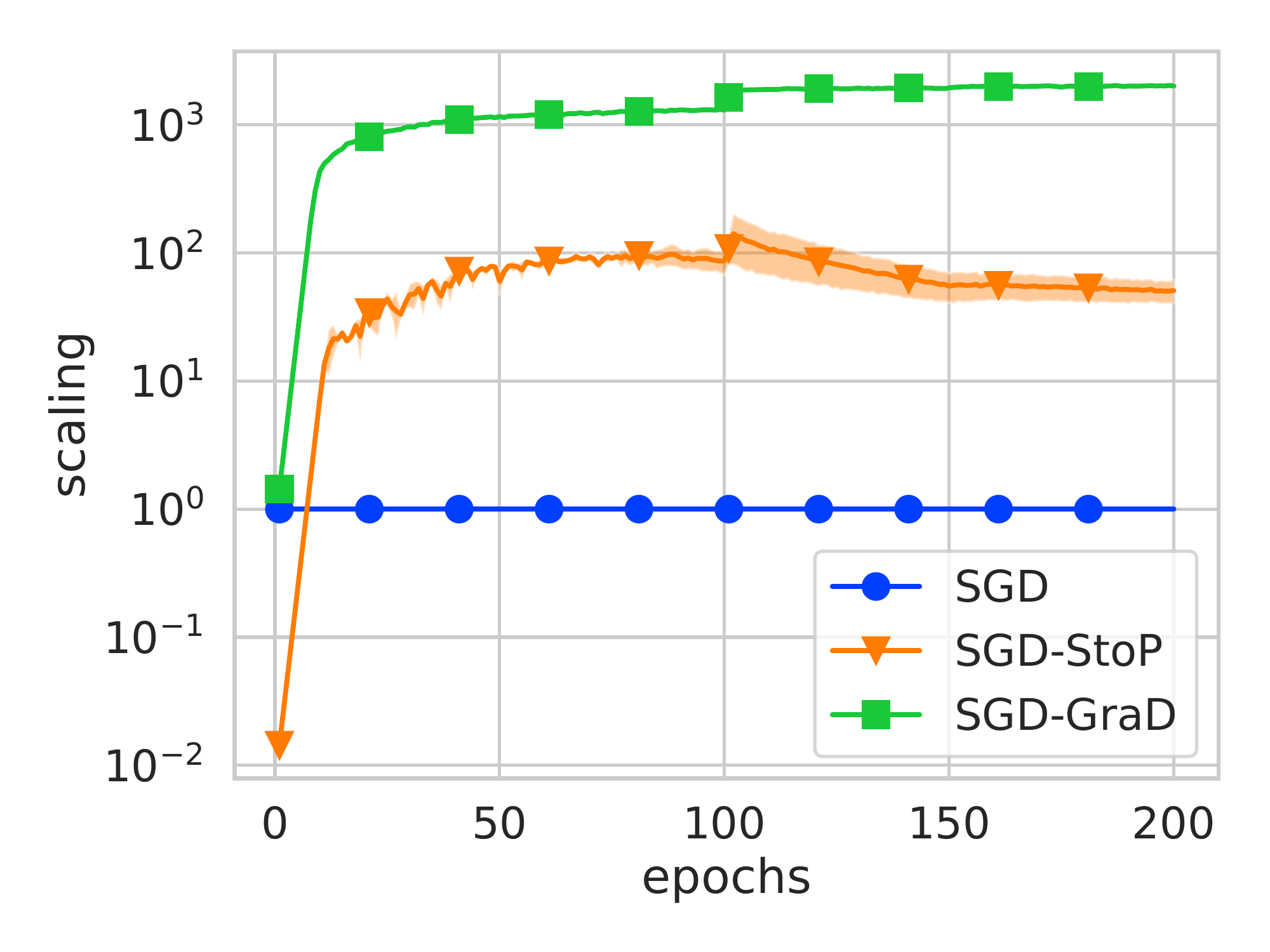}
\includegraphics[width=0.32\textwidth]{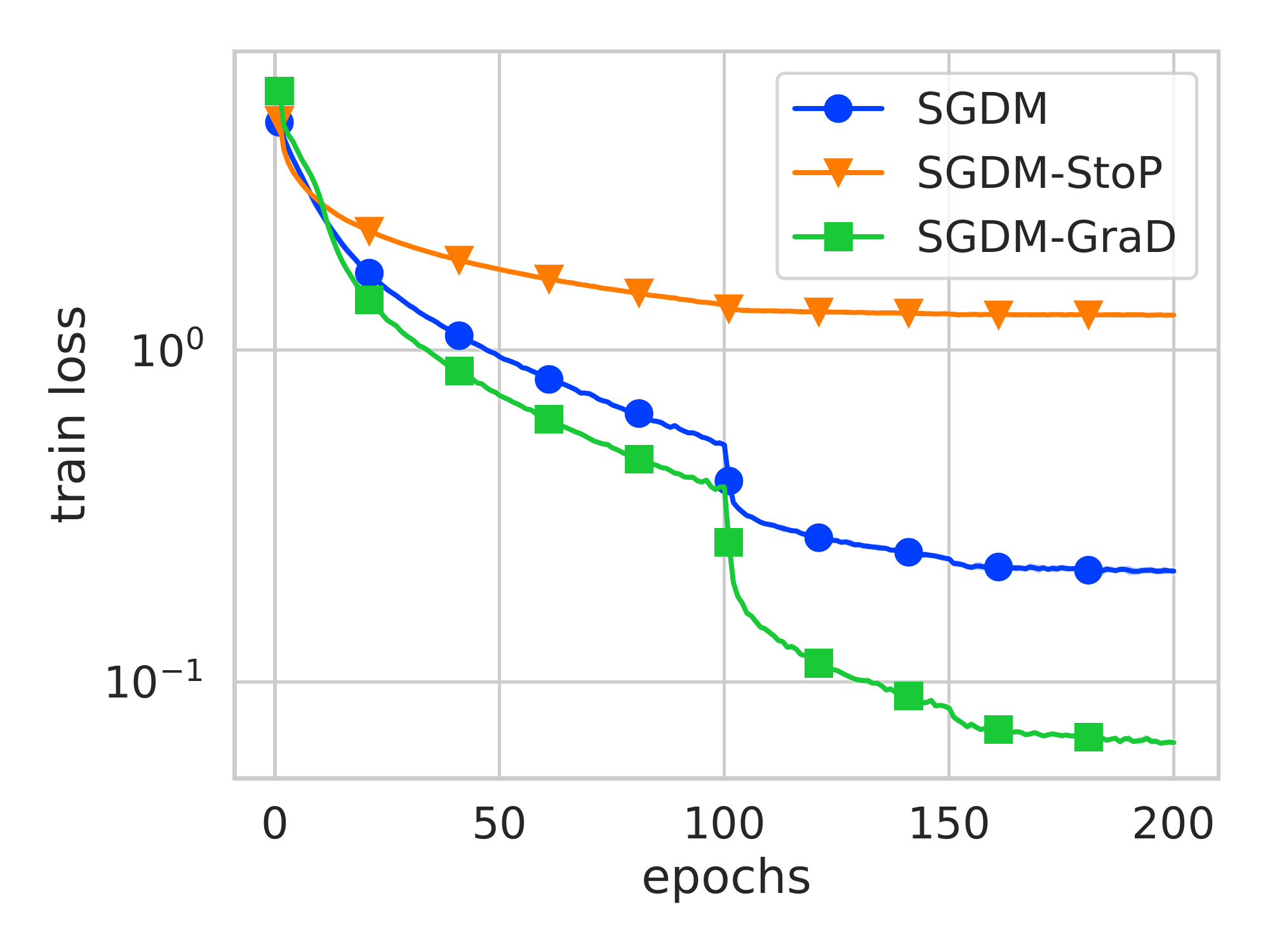}
\includegraphics[width=0.32\textwidth]{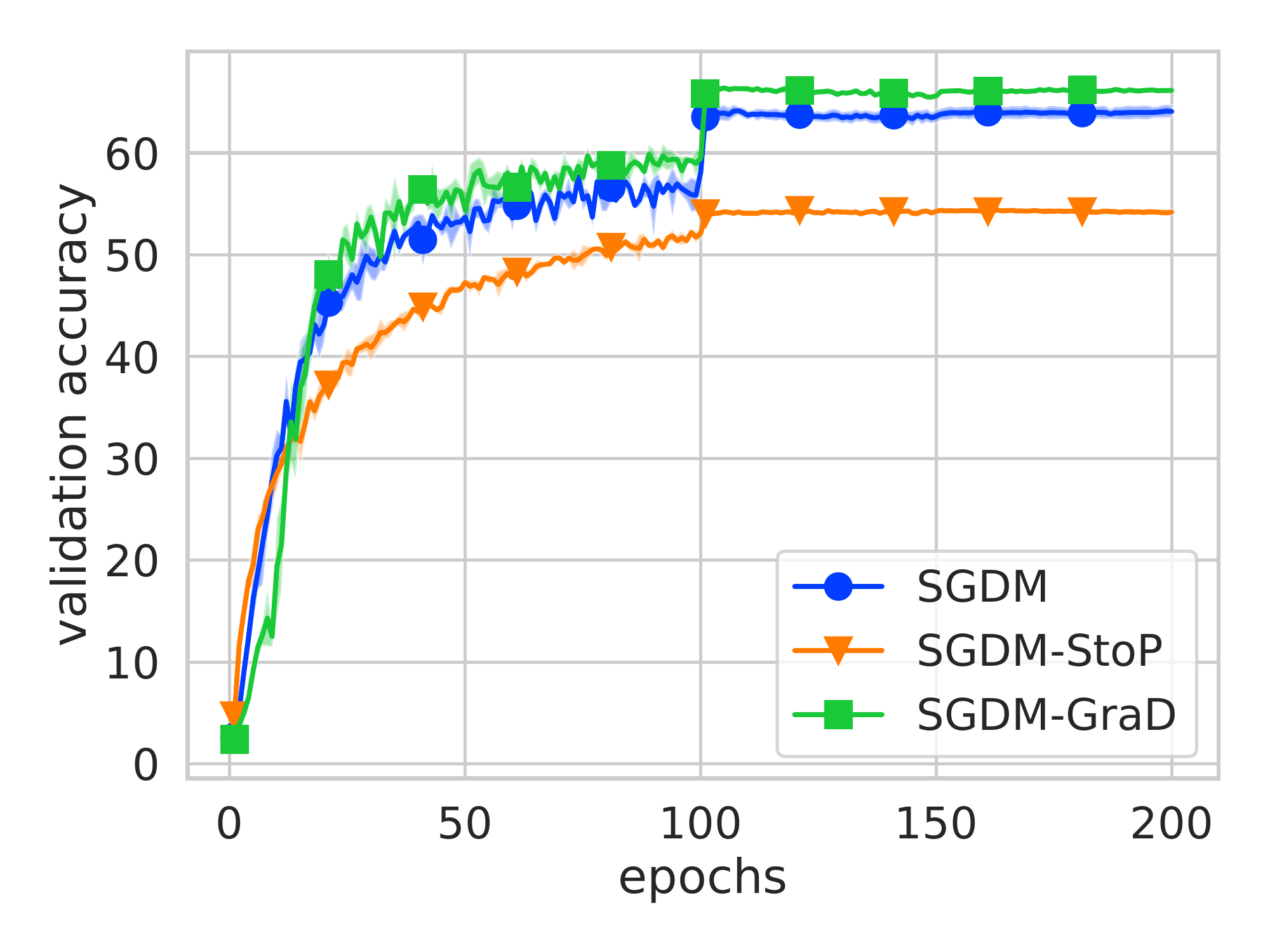}
\includegraphics[width=0.32\textwidth]{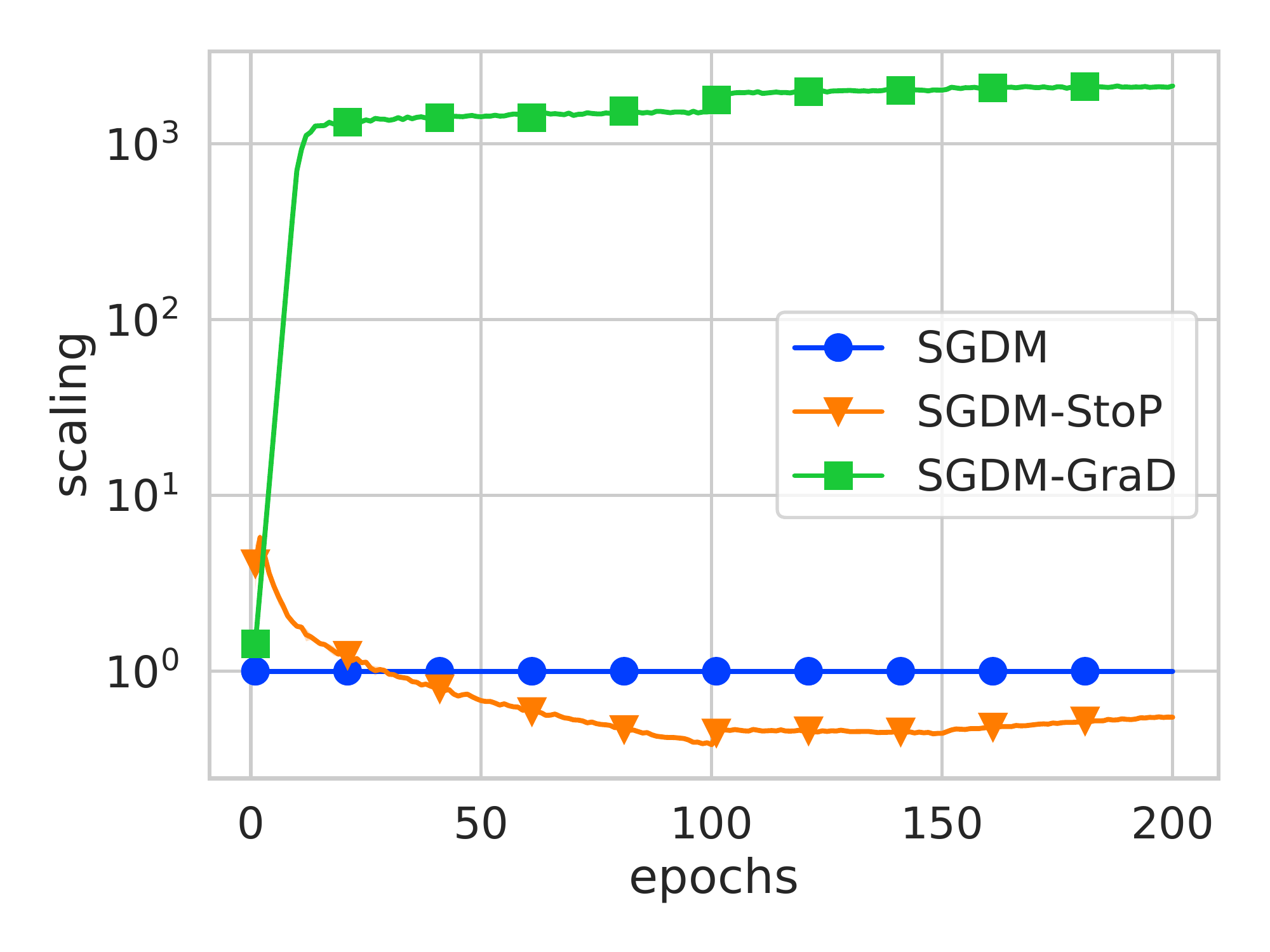}
\includegraphics[width=0.32\textwidth]{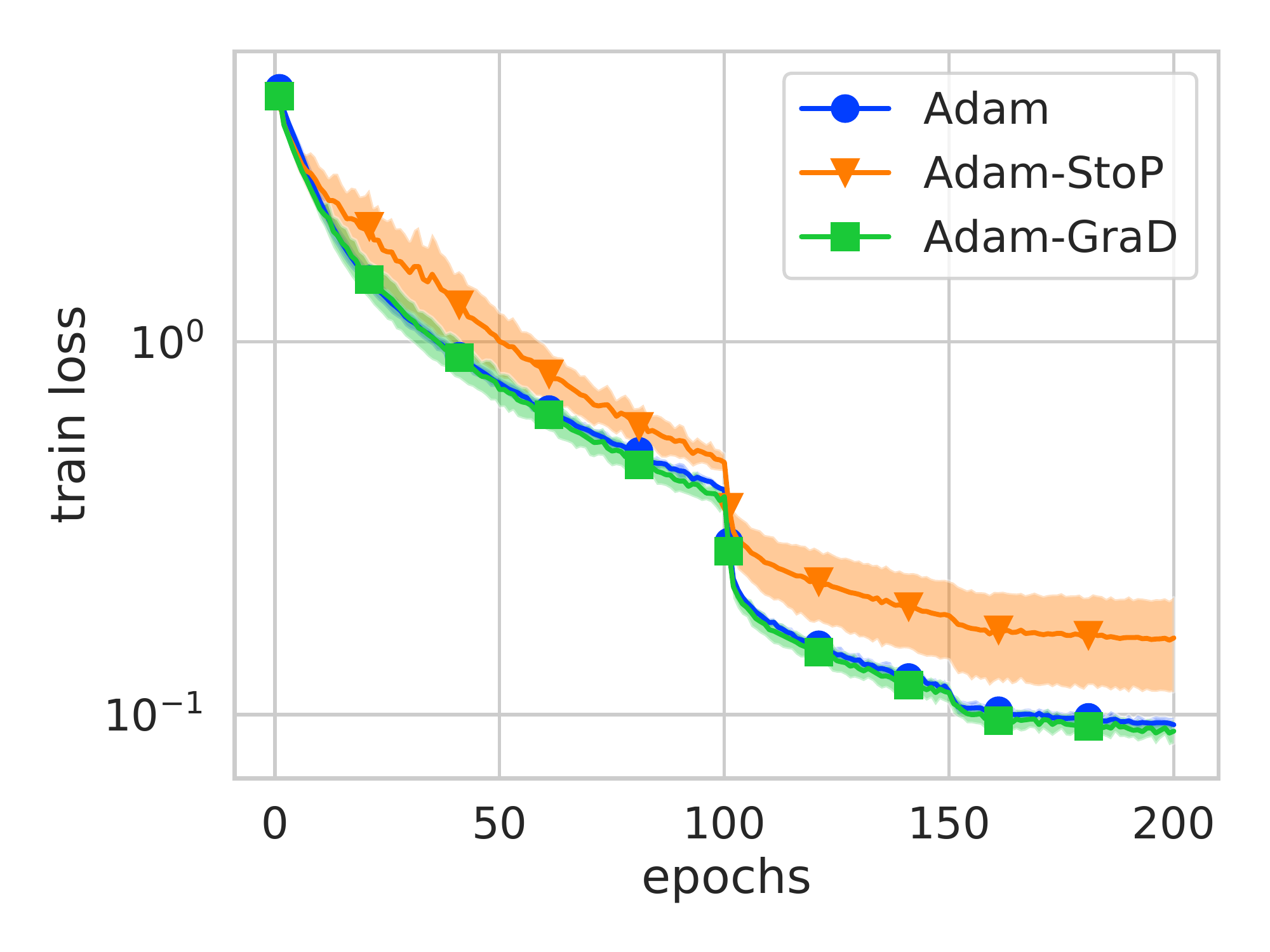}
\includegraphics[width=0.32\textwidth]{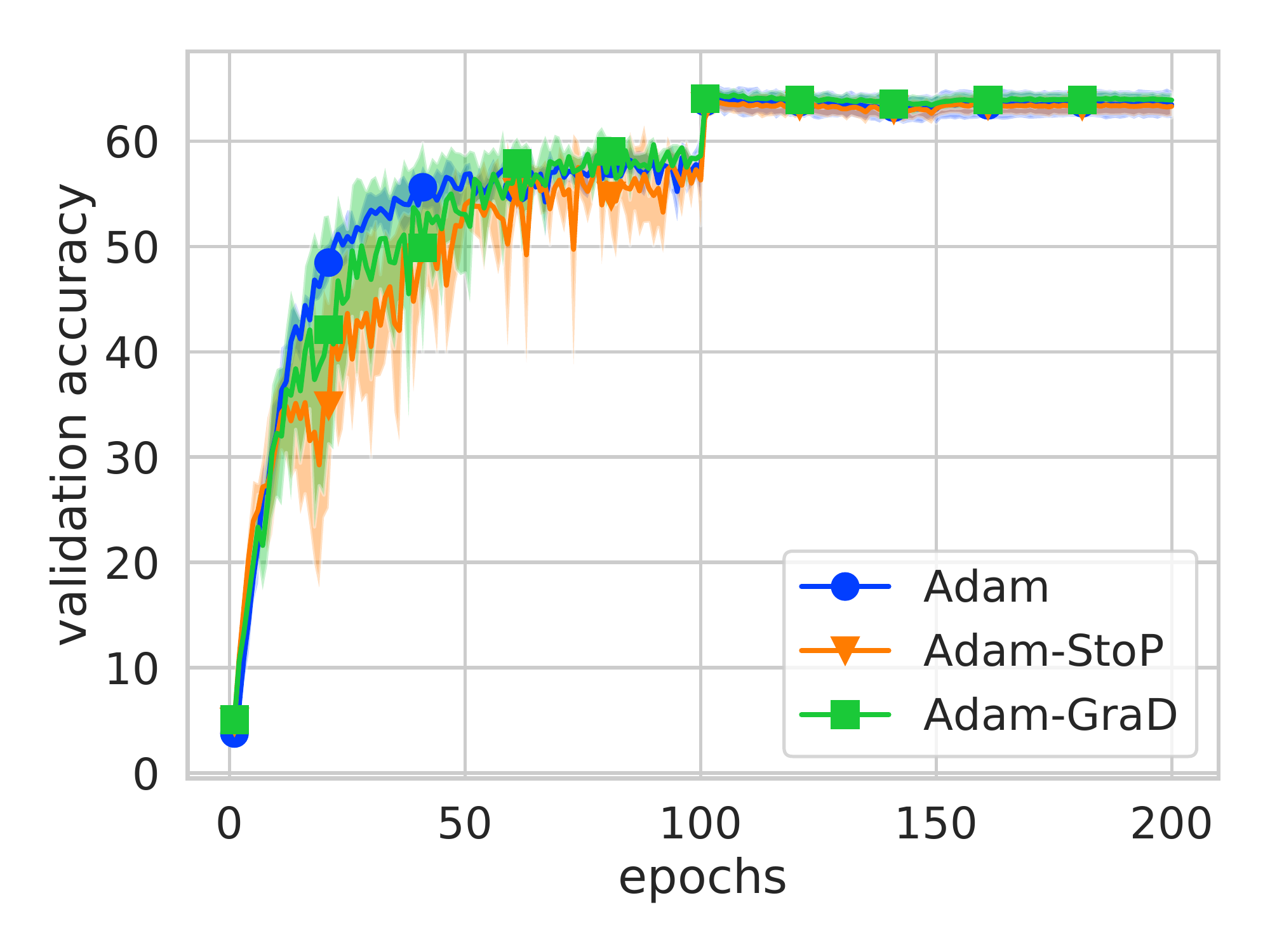}
\includegraphics[width=0.32\textwidth]{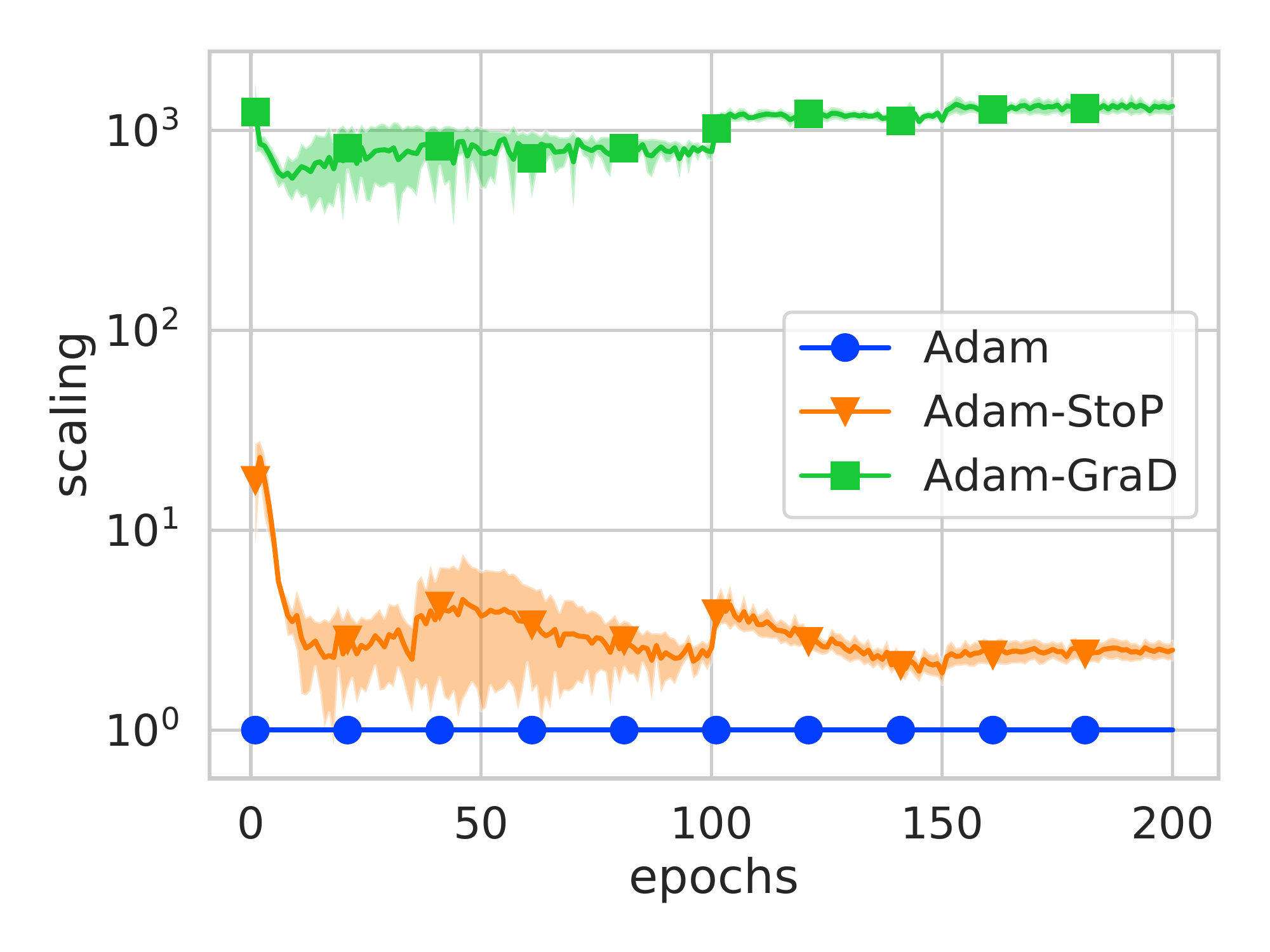}
\caption{ResNet32 on CIFAR100.  For details see caption in Figure~\ref{fig:lenet_mnist}.}
\label{fig:resnet32_cifar100}
\end{figure}

\subsection{Neural Networks: Practical Performance}

In this section, we focus on experiments with neural nets. We select \texttt{SGD}, \texttt{SGDM} (\texttt{SGD} with momentum $0.9$) and \texttt{Adam}~\citep{kingma2014adam} as our baselines. The goal of these experiments is to show that extending these baselines with \stp and \grad leads to superior practical methods. As we show, this is the case for \grad in the vast majority of considered applications. Extending   \texttt{SGD} by the proposed step sizes is equivalent to \stp and \grad as presented in Section~\ref{sec:stp_and_grad}. Combination with \texttt{SGDM} and \texttt{Adam} is motivated by our theoretical analysis as described in Section~\ref{sec:grad_momentum}. Therefore, we first scale the estimated gradient with \stp and \grad step size (scaling) and then use this quantity as a gradient estimator instead of non-scaled version. 

For practical purposes, we add a small constant $\delta = 10^{-6}$ to the denominator of both \stp and \grad scaling for numerical stability, similarly to \texttt{SPS}~\citep{sps} or \texttt{ALI-G}~\citep{berrada2020training}. We initialize $\gamma_{\max}^0 = 1$ for \grad. For \stp, we set step size equal to $1$ and tune $\gamma_{\max}^0$ as \stp step size has no extra dependence. For the upper bound $\gamma_{\max}^k$, we use a smoothing rule $\gamma_{\max}^k = \tau^{\nicefrac{n}{N}} \gamma^{k-1}$ with $\tau = 2$, where $n$ is batch size and $N$ is the size of the dataset. The same upper bound was used in \citep{sps} with inspiration from smoothing procedure for Barzilai-Borwein step size selection procedure for \texttt{SGD}~\citep{NIPS2016_c86a7ee3}. We display average and one standard error for each method over 3 independent runs with fixed seeds. For reproducibility, we provide our PyTorch~\citep{pytorch} implementation as a part of the submission. To access individual gradient norms, which is a feature not directly supported by PyTorch,  we use BackPACK~\citep{dangel2020backpack}. All our experiments are run on the private cluster, each with single V100 GPU. 

Firstly, we evaluate \texttt{SGD}, \texttt{SGDM} as baselines on LeNet~\citep{lenet} on MNIST dataset~\citep{mnist}. We don't employ any upper bound, i.e., $\forall i \in \mathcal{N}^0: \gamma_{\max}^i = \infty$. We run $30$ epochs with $1024$ batch size. For each method, we tune its step size using the set  \\ $\{1, 0.1, 0.01, 0.0001, 0.00001\}$. For \stp tuning, this means scaling $\gamma^k_{\stp}$ by the step size. In Figure~\ref{fig:lenet_mnist}, we display training loss corresponding to the step size with the minimum final average loss, validation accuracy corresponding to the step size with the maximum final average validation accuracy and train scale. i.e., 1 for plain baseline, step size scaling for \grad and actual step size for \stp. One can see that while \stp seems to be a practically inferior method to tuned \texttt{SGD} with or without momentum, \grad can enhance both, and it provides improved accuracy and faster convergence. In Table~\ref{tab:lenet_mnist}, we display final validation accuracy that also suggests the supreme performance of the \grad gradient scaling.  

Secondly, we conclude experiments using two other image classification tasks CIFAR10/100 \citep{cifar} on relatively small ResNet20/32~\citep{resnet, resnet_impl} architectures and large batch size $1024$, respectively. In the appendix, we also provide experiments with larger models ResNet18/34~\citep{resnet} and smaller batch size $128$ on CIFAR10/100, respectively. The obtained results are consistent with the one provided in the main paper. We employ standard data augmentation techniques for the training set, such as normalization, random cropping, and random horizontal flipping. We run \texttt{SGD}, \texttt{SGDM} and \texttt{Adam} as baselines. As for the previous experiment, we tune step size. In addition, we extend tuning by hand-crafted learning schedule designed for \texttt{SGD}-type methods, where we decrease step size by $10$ after running $50\%$ and $75\%$ of epochs, by decoupled weight-decay option~\citep{loshchilov2017decoupled} $\{0., 0.00005\}$ and by smoothing rule for \stp and \grad. We observed that all methods benefit from this learning schedule, and they provide better generalization with non-zero weight decay. In addition, \grad provides better performance when extended with the smoothing rule. This is not always the case for \stp. This is because if we extend \stp by the smoothing rule, we don't scale its step size, but we use the default value $1$, which sometimes makes \stp divergent. This might be fixed by tuning both step size scaling and the upper bound, but we decided to leave this out as it requires tuning two parameters instead of only one. We report the same quantities as for the previous experiment. Looking into Figure~\ref{fig:resnet20_cifar10} and Table~\ref{tab:resnet20_cifar10}, one can see that both \stp and \grad improve upon \texttt{SGD}, \texttt{SGDM} while providing slightly worse performance for \texttt{Adam}. This is the only case where we observe \grad does not surpass all competing methods. For Figure~\ref{fig:resnet32_cifar100} and Table~\ref{tab:resnet32_cifar100}, \grad gradient scaling gives the best performance while \stp provides comparable performance to tuned plain optimizers except for \texttt{SGDM}.

\section{Conclusion}

We proposed and analyzed two new theoretical stochastic adaptive methods \stps and \grads that enjoy deterministic-like convergence to the exact minimizer. We also offer practical variants of these methods--\stp and \grad, for which we show convergence to the exact minimizer under overparameterization and a fixed neighbourhood in general. Furthermore, via experiments on a variety of image classification tasks, we showed solid practical performance of \texttt{SGD}, \texttt{SGDM} and \texttt{Adam} optimizers in combination with \grad gradient scaling as compared to state-of-the-art optimization methods. 

For future work, we are interested in devising variance reduced adaptive methods that provide convergence to the exact solution in general. Furthermore, we would like to remove dependence on smoothness $L$ from the \grad estimator to avoid step size tuning, for instance, by approximation of the inverse local Lipschitz constant as in~\citep{malitsky2019adaptive}.


\bibliography{scaled_grad}

\clearpage
\appendix
\part*{Supplementary Materials}

\section{Extra Experiments}

\begin{table}[h]
\centering
\caption{Final validation accuracy of ResNet18 on CIFAR10.}
\label{tab:resnet18_cifar10}
\begin{tabular}{|c|c|c|c|}
\hline
 & \texttt{SGD} & \texttt{SGDM}         & \texttt{Adam}         \\ \hline
\texttt{None}      & $93.08 \pm 0.12$ & $93.84 \pm 0.13$ & $94.10 \pm 0.11$ \\ \hline
\stp      & $90.99 \pm 0.21$ & $93.45 \pm 0.39$ & $93.85 \pm 0.41$ \\ \hline
\grad      & $\mathbf{94.18} \pm 0.20$ & ${\color{red}\mathbf{94.23}} \pm 0.16$ & $\mathbf{94.11} \pm 0.18$ \\ \hline
\end{tabular}
\end{table}

\begin{figure}[h]
\centering
\includegraphics[width=0.32\textwidth]{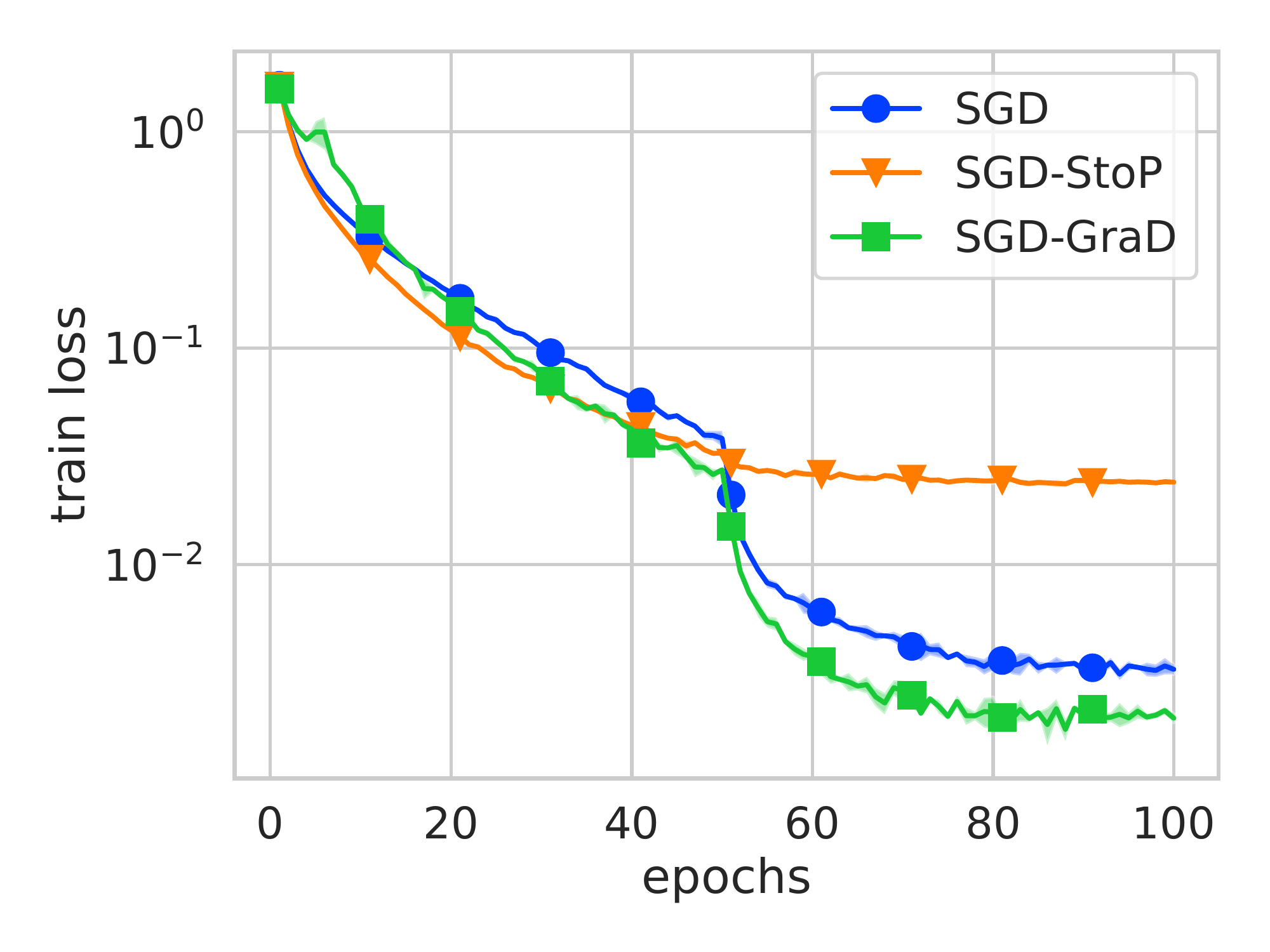}
\includegraphics[width=0.32\textwidth]{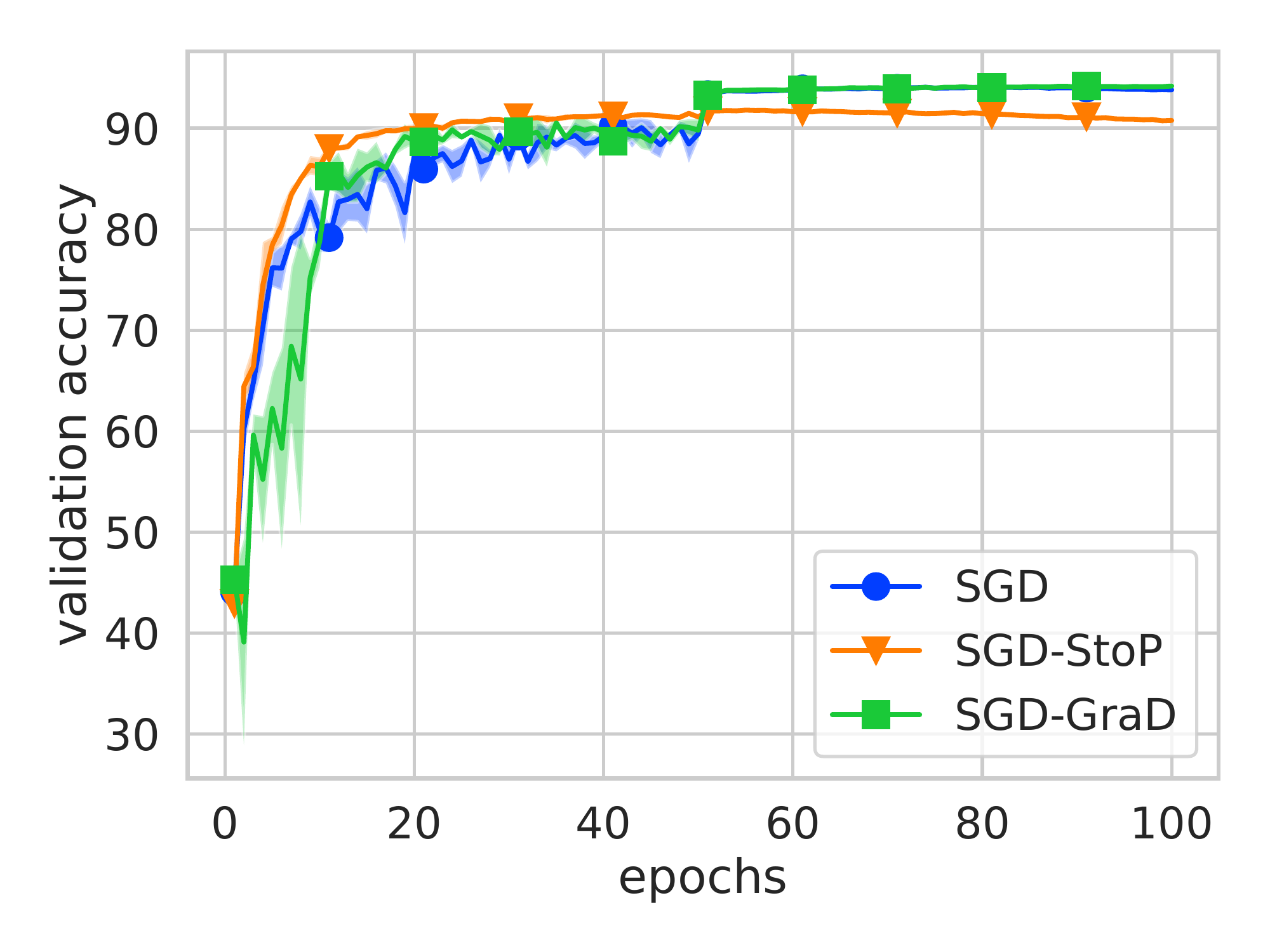}
\includegraphics[width=0.32\textwidth]{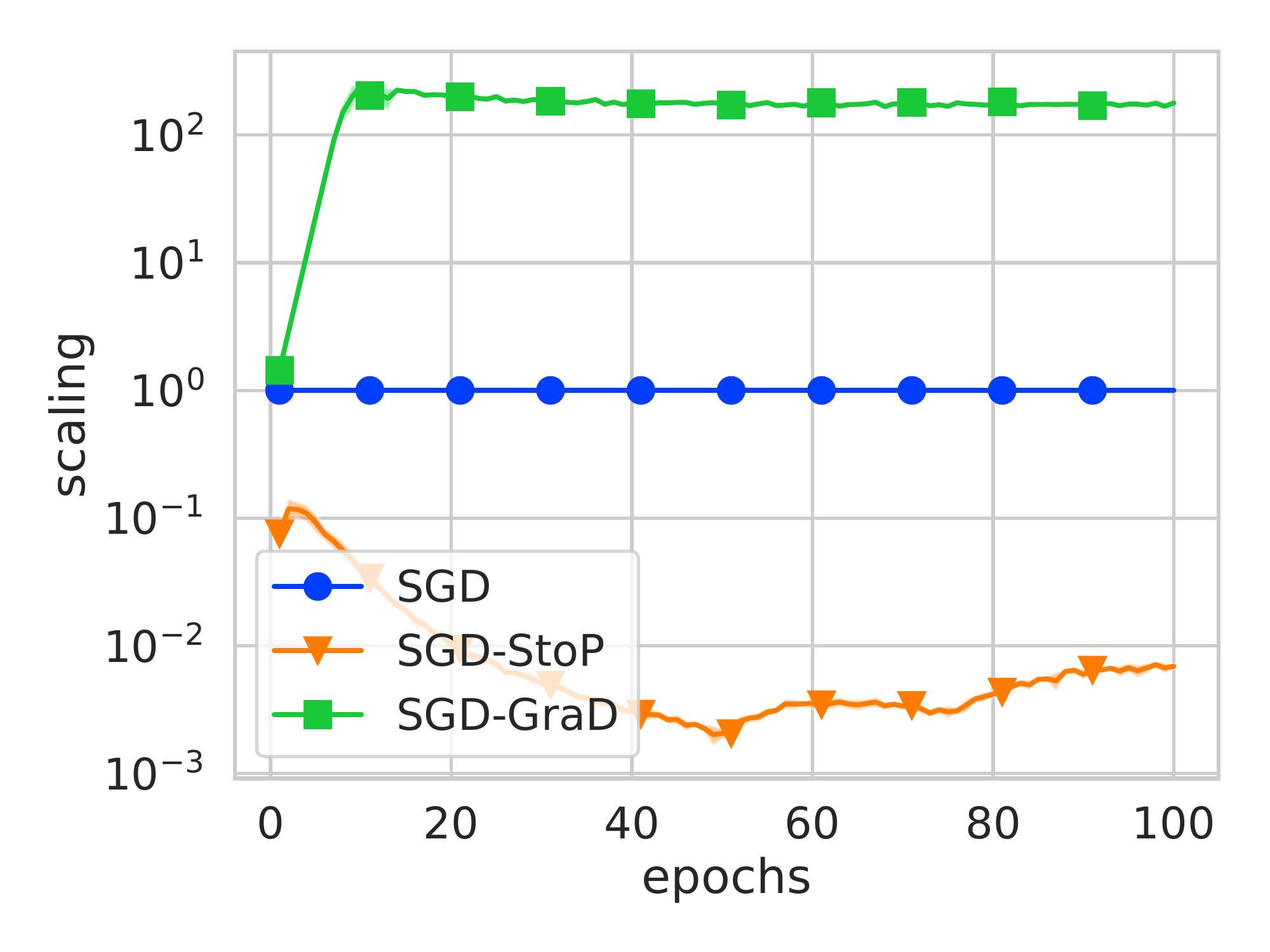}
\includegraphics[width=0.32\textwidth]{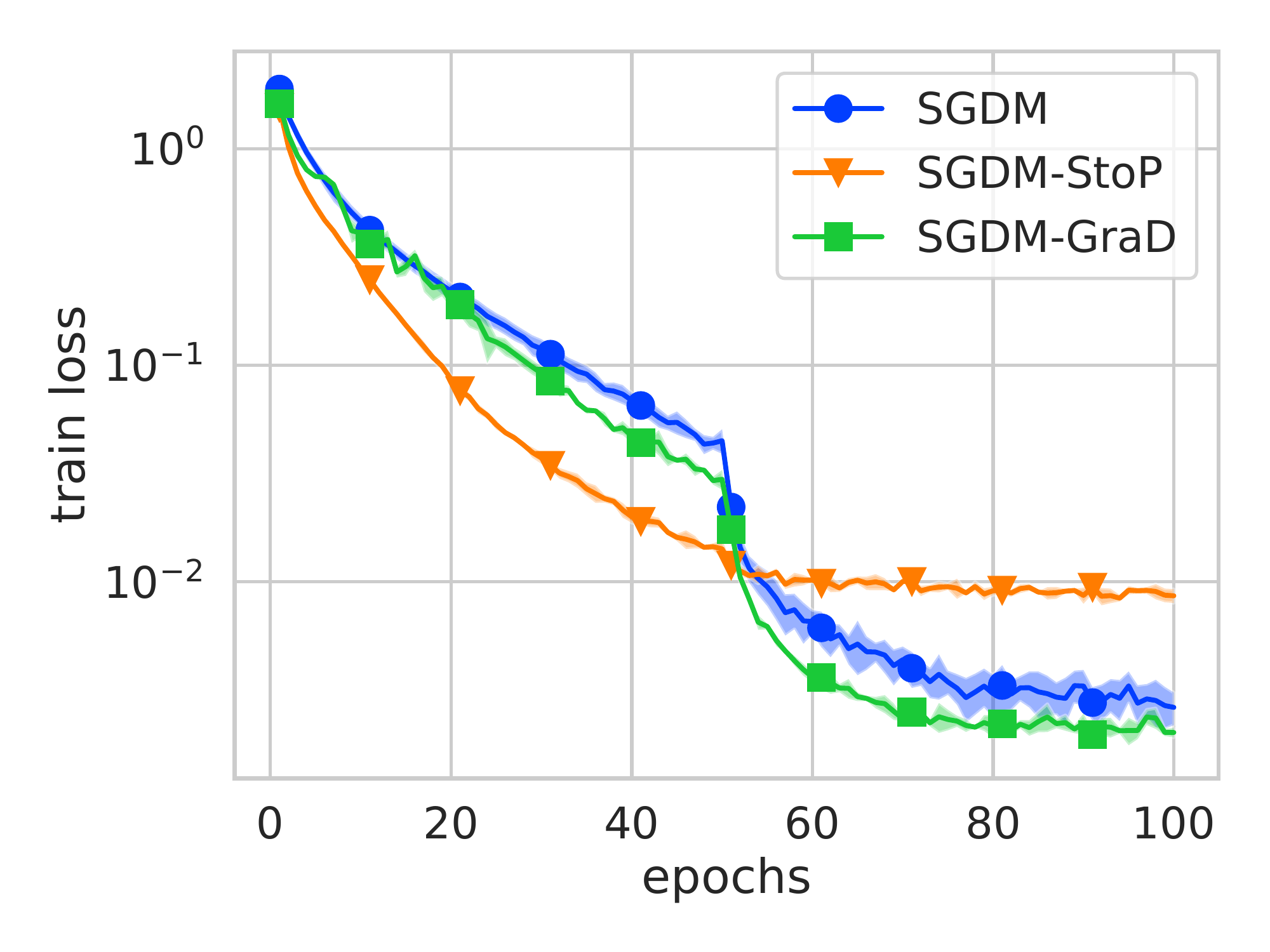}
\includegraphics[width=0.32\textwidth]{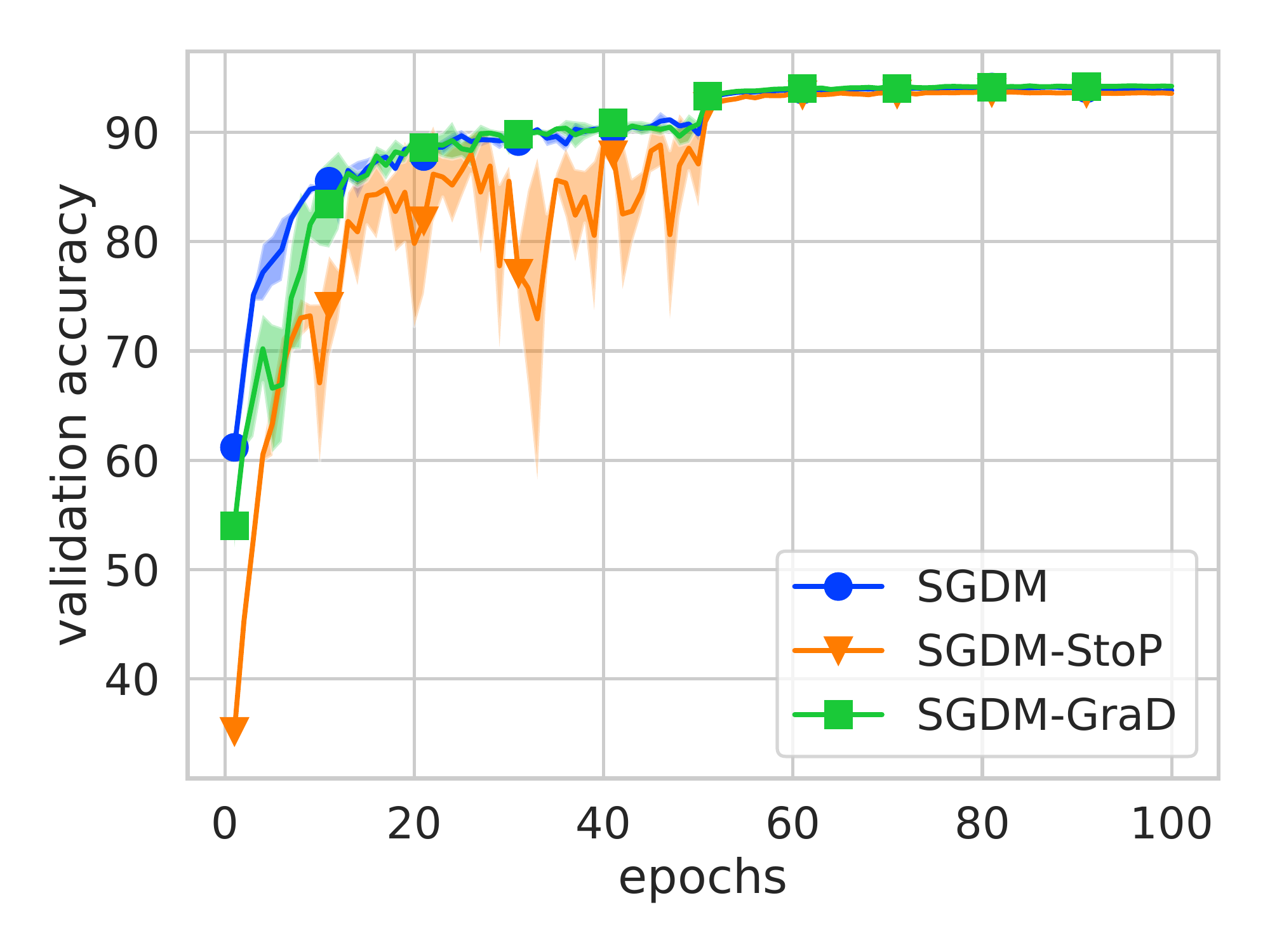}
\includegraphics[width=0.32\textwidth]{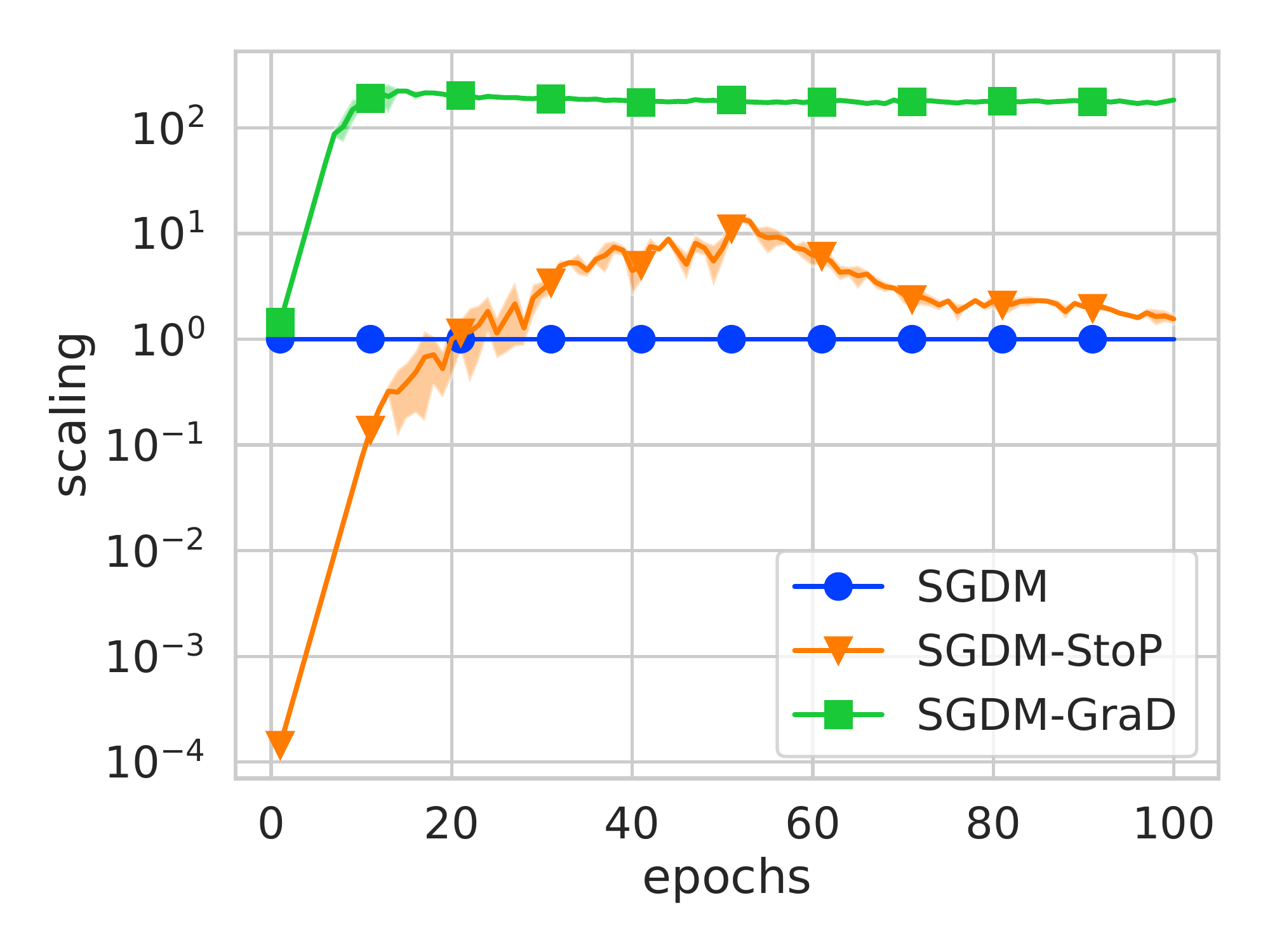}
\includegraphics[width=0.32\textwidth]{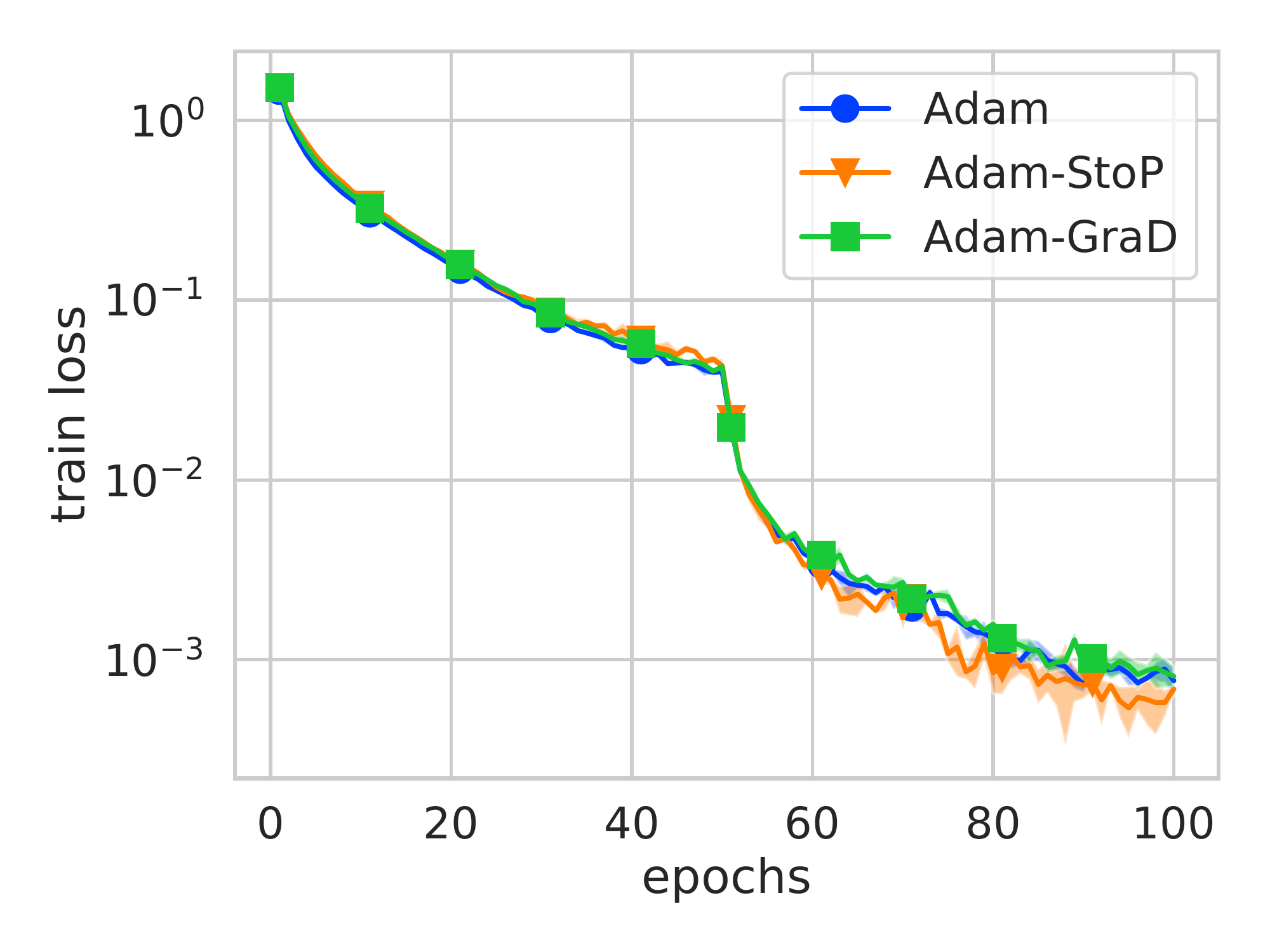}
\includegraphics[width=0.32\textwidth]{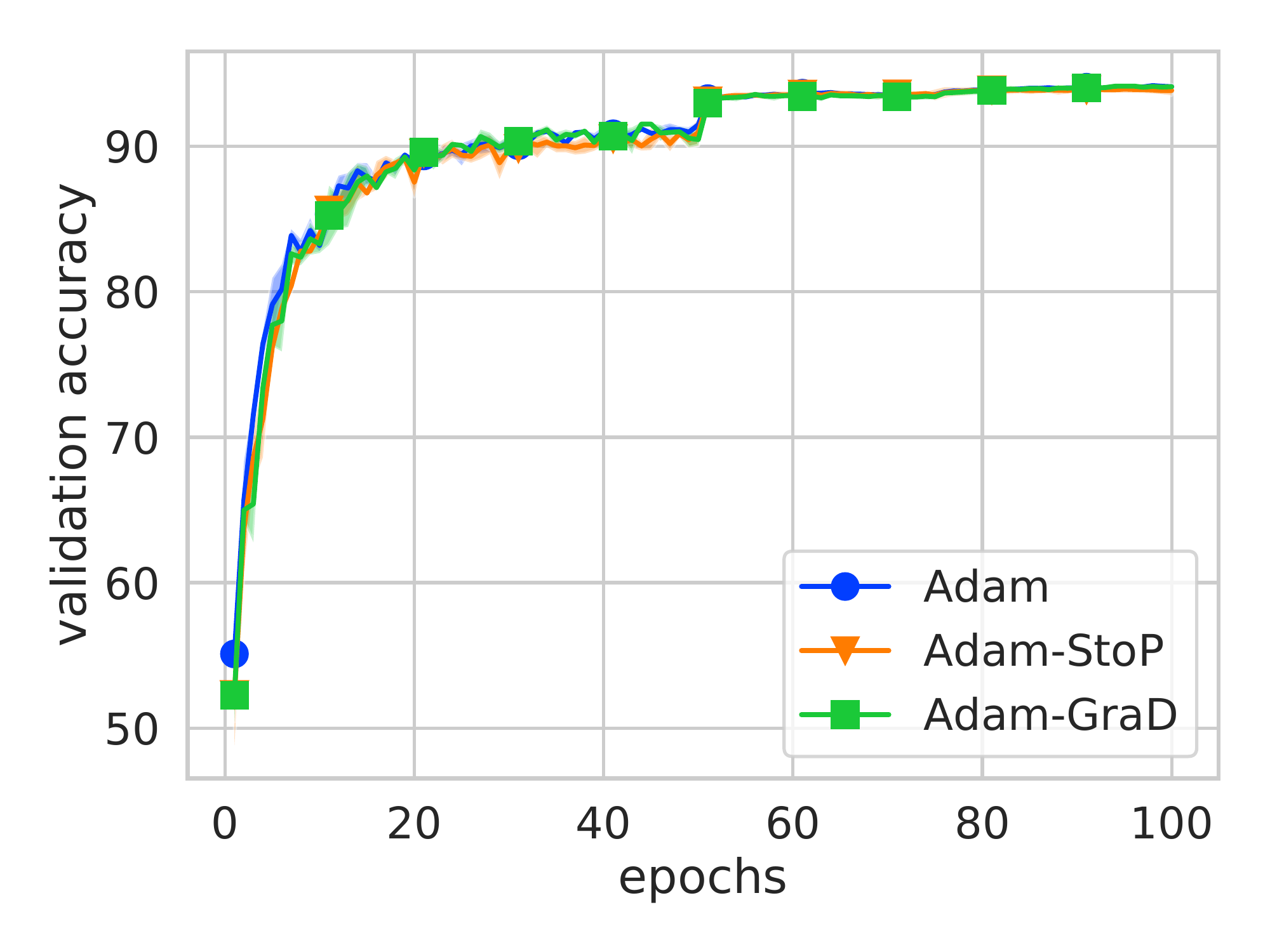}
\includegraphics[width=0.32\textwidth]{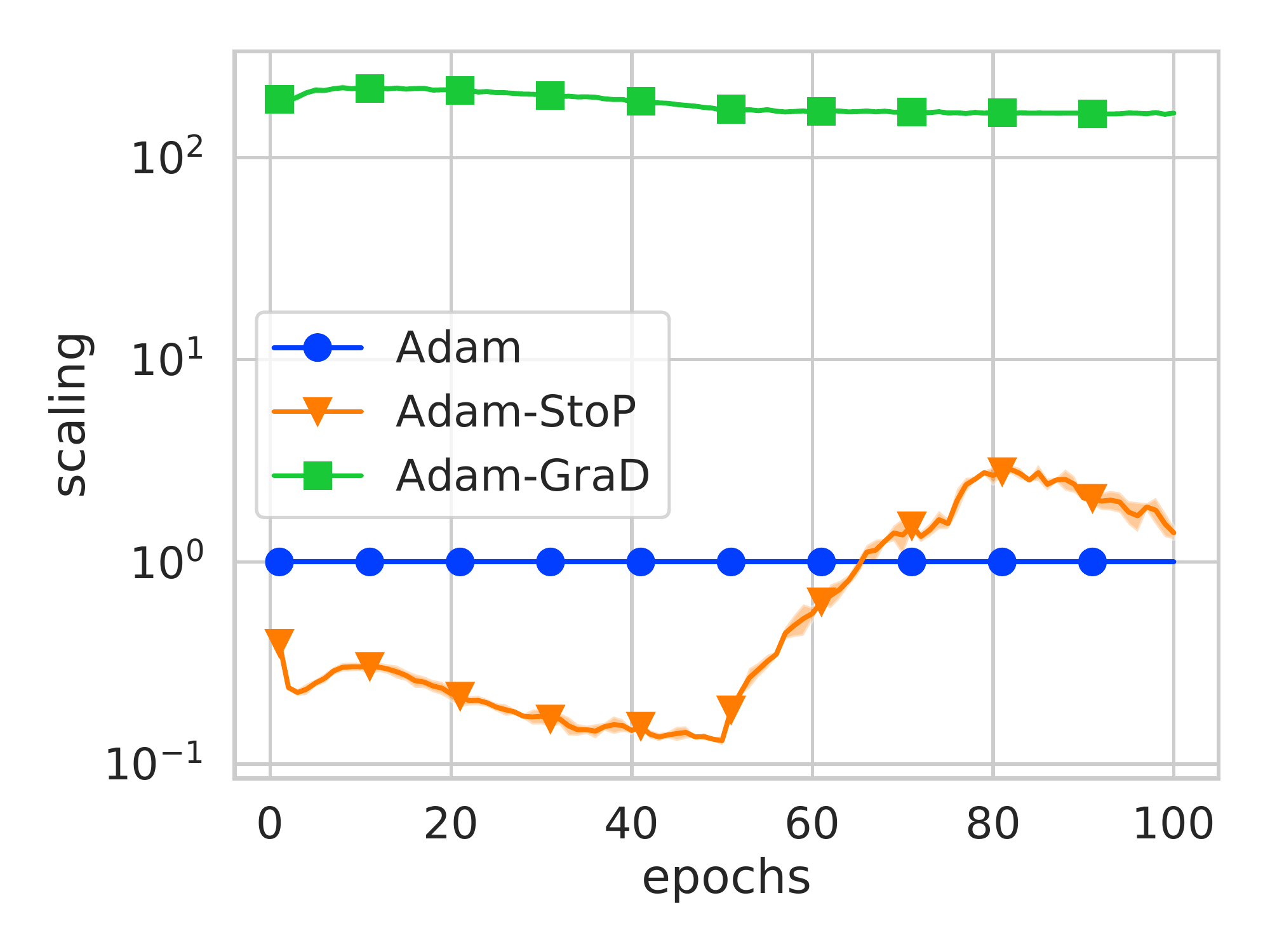}
\caption{ResNet18 on CIFAR10.  For details see caption in Figure~\ref{fig:lenet_mnist}.}
\label{fig:resnet18_cifar10}
\end{figure}

\clearpage

\begin{table}[t]
\centering
\caption{Final validation accuracy of ResNet34 on CIFAR100.}
\label{tab:resnet34_cifar100}
\begin{tabular}{|c|c|c|c|}
\hline
 & \texttt{SGD} & \texttt{SGDM}         & \texttt{Adam}         \\ \hline
\texttt{None}      & $75.82 \pm 0.20$ & $75.48 \pm 0.56$ & $75.13 \pm 0.22$ \\ \hline
\stp      & $72.74 \pm 0.51$ & $70.19 \pm 0.71$ & $75.22 \pm 0.03$ \\ \hline
\grad      & $\mathbf{76.28} \pm 0.23$ & ${\color{red}\mathbf{76.48}} \pm 0.08$ & $\mathbf{75.29} \pm 0.24$ \\ \hline
\end{tabular}
\end{table}

\begin{figure}[t]
\centering
\includegraphics[width=0.32\textwidth]{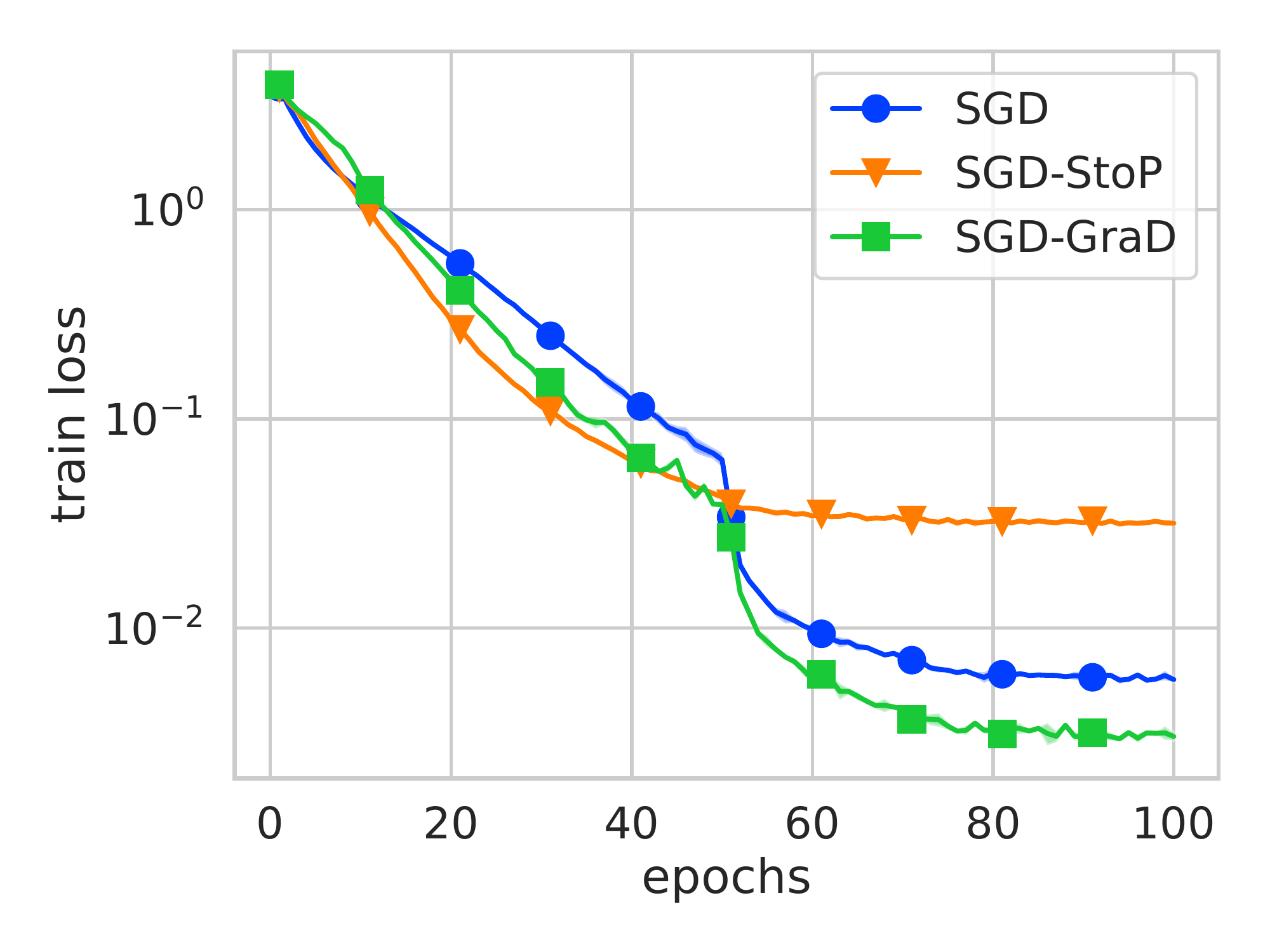}
\includegraphics[width=0.32\textwidth]{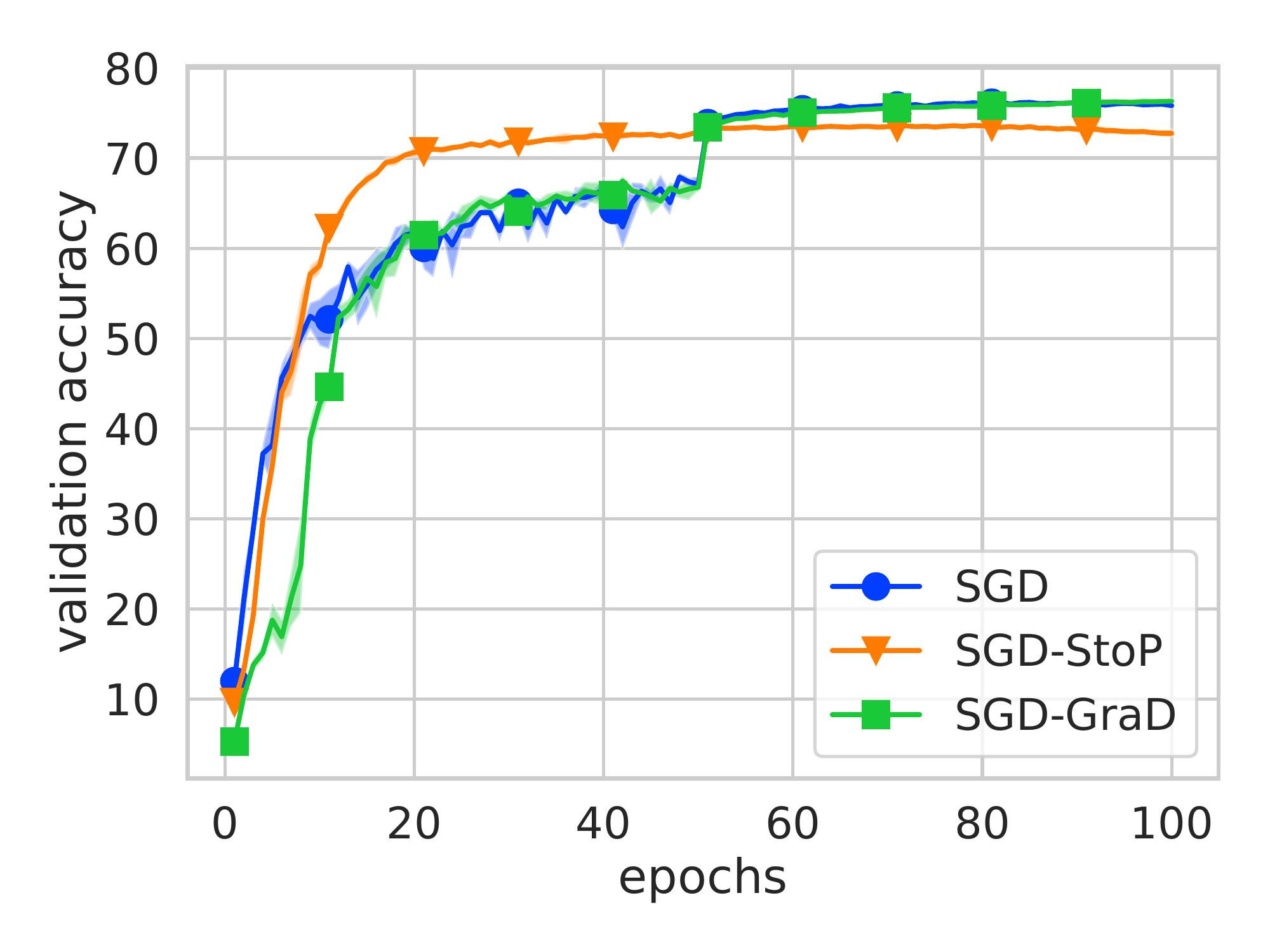}
\includegraphics[width=0.32\textwidth]{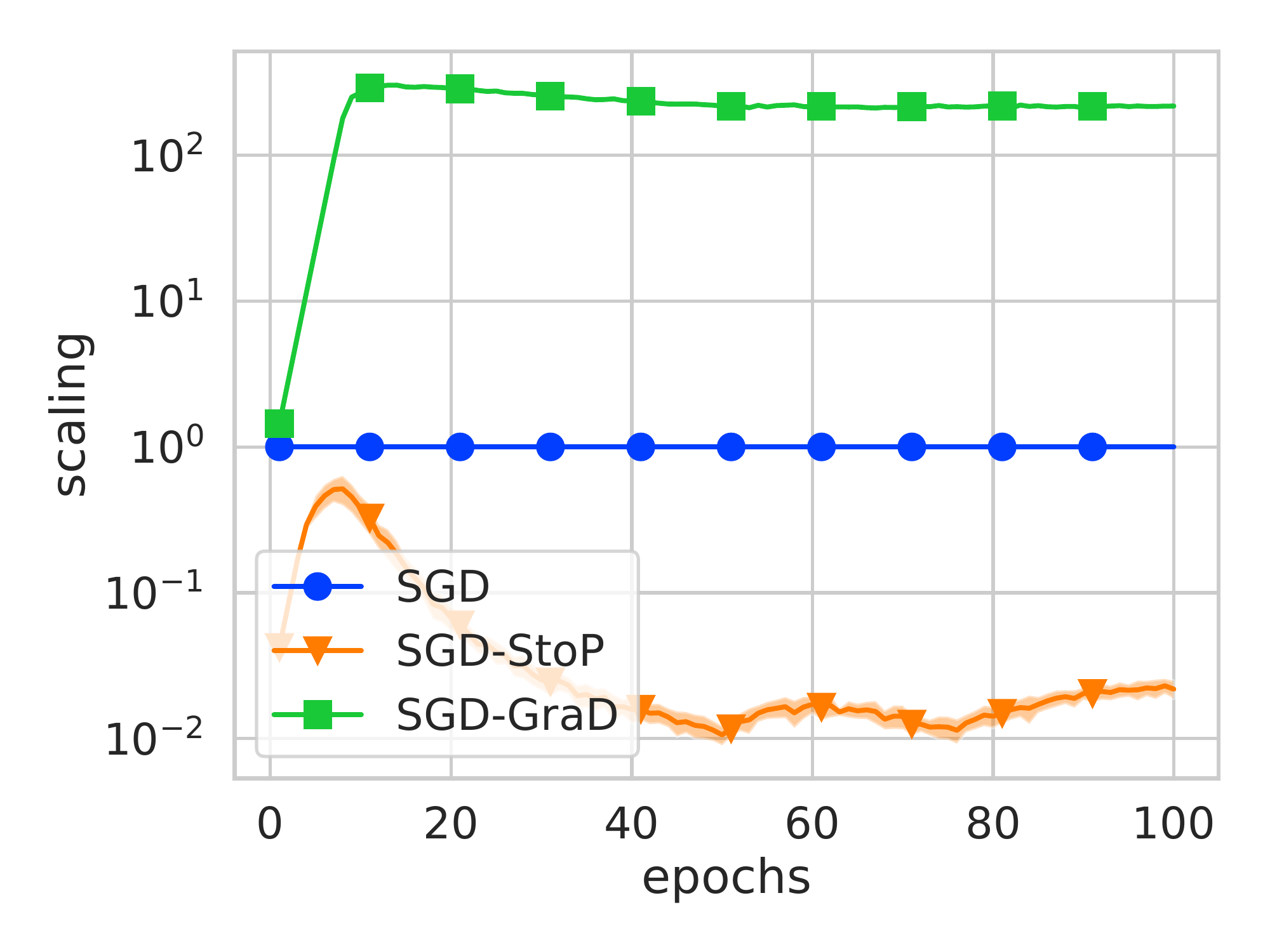}
\includegraphics[width=0.32\textwidth]{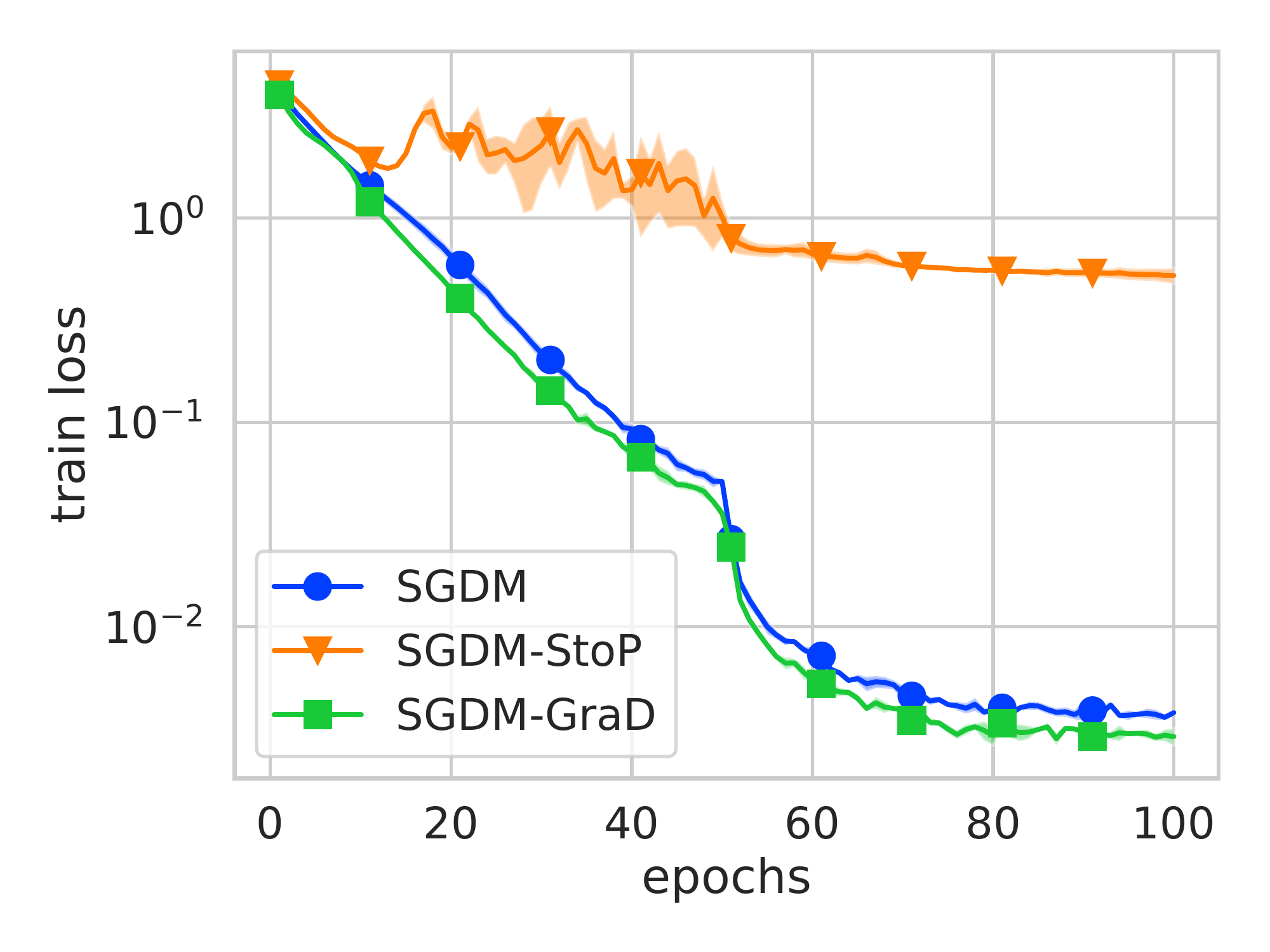}
\includegraphics[width=0.32\textwidth]{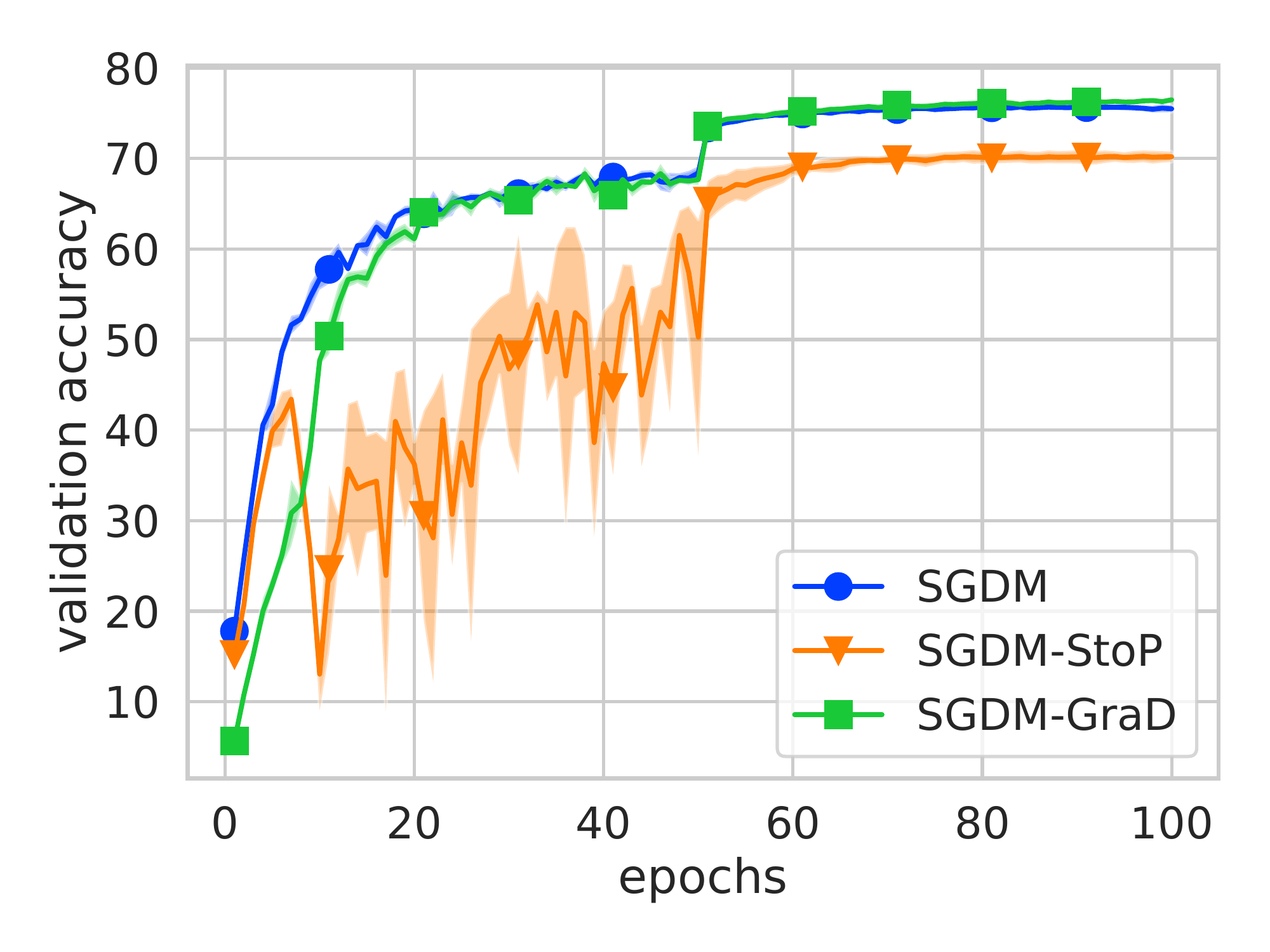}
\includegraphics[width=0.32\textwidth]{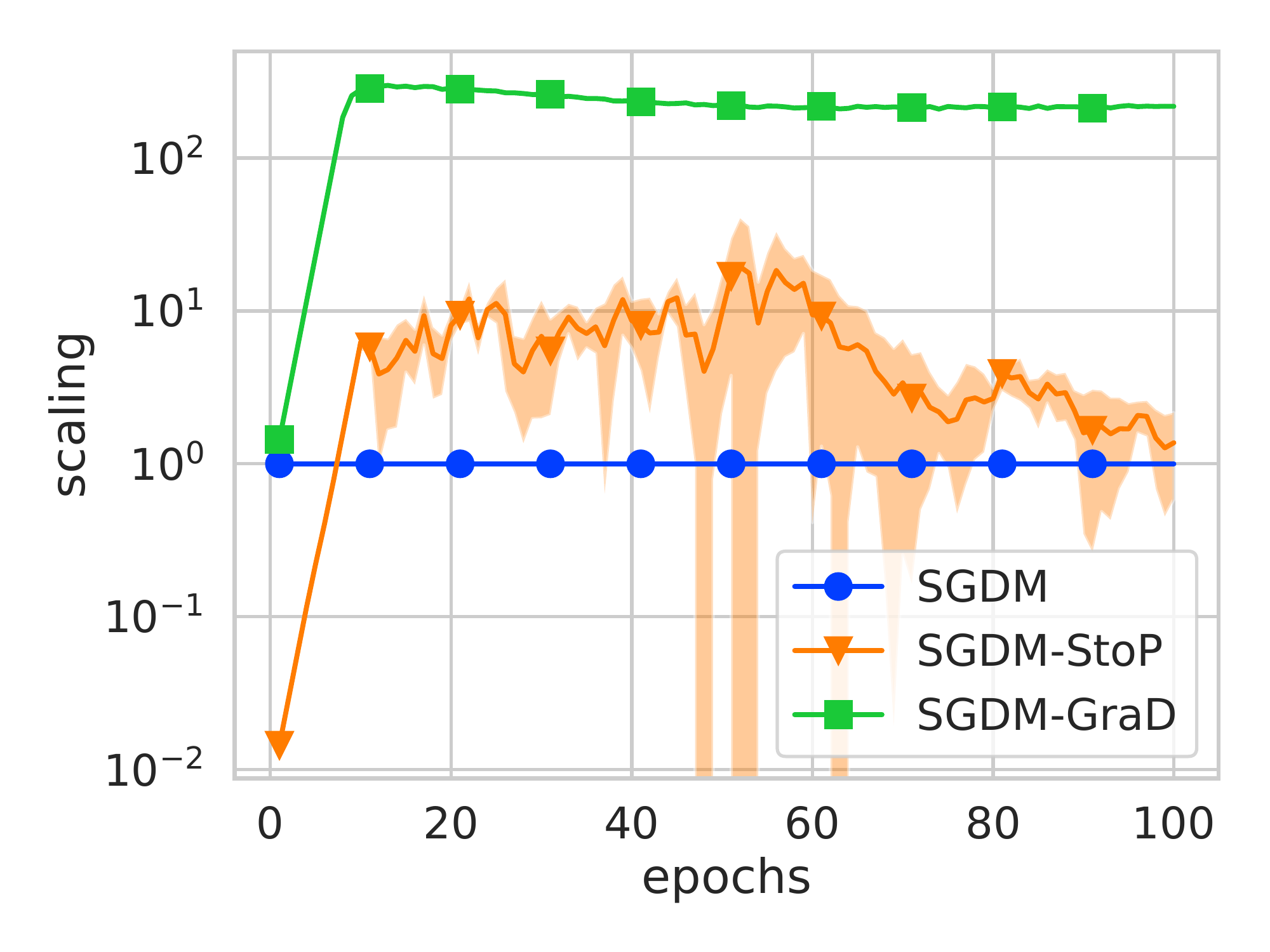}
\includegraphics[width=0.32\textwidth]{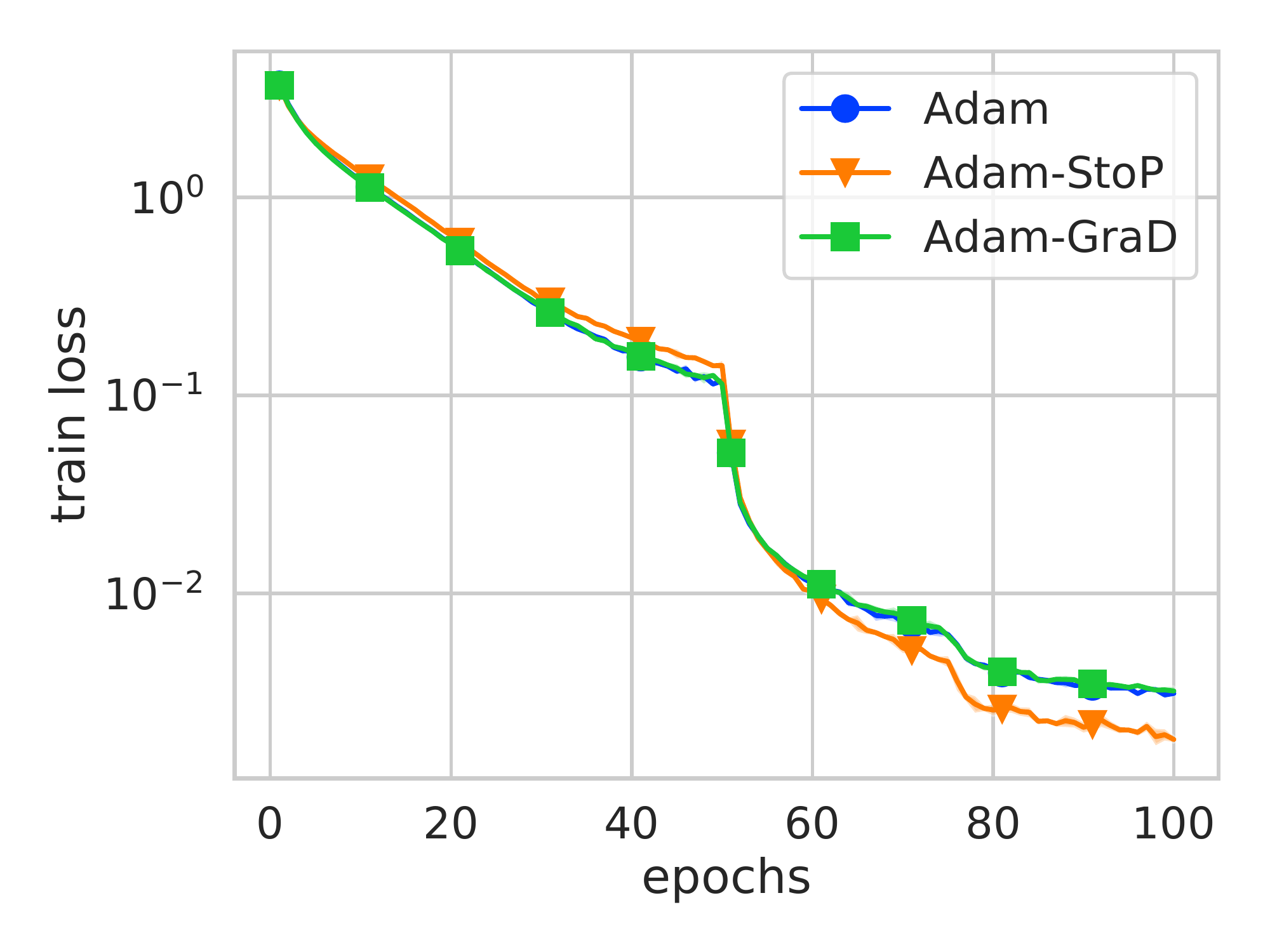}
\includegraphics[width=0.32\textwidth]{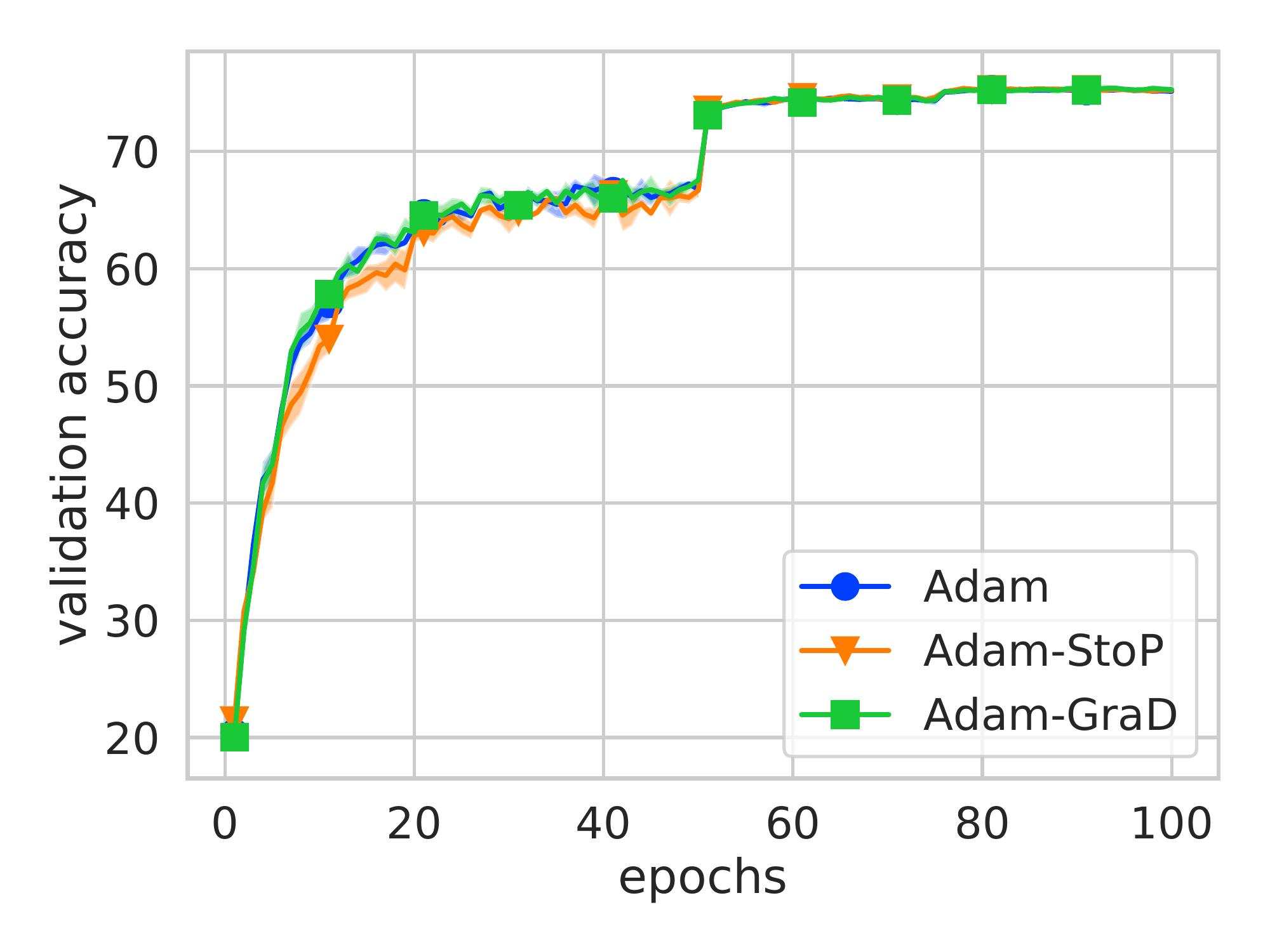}
\includegraphics[width=0.32\textwidth]{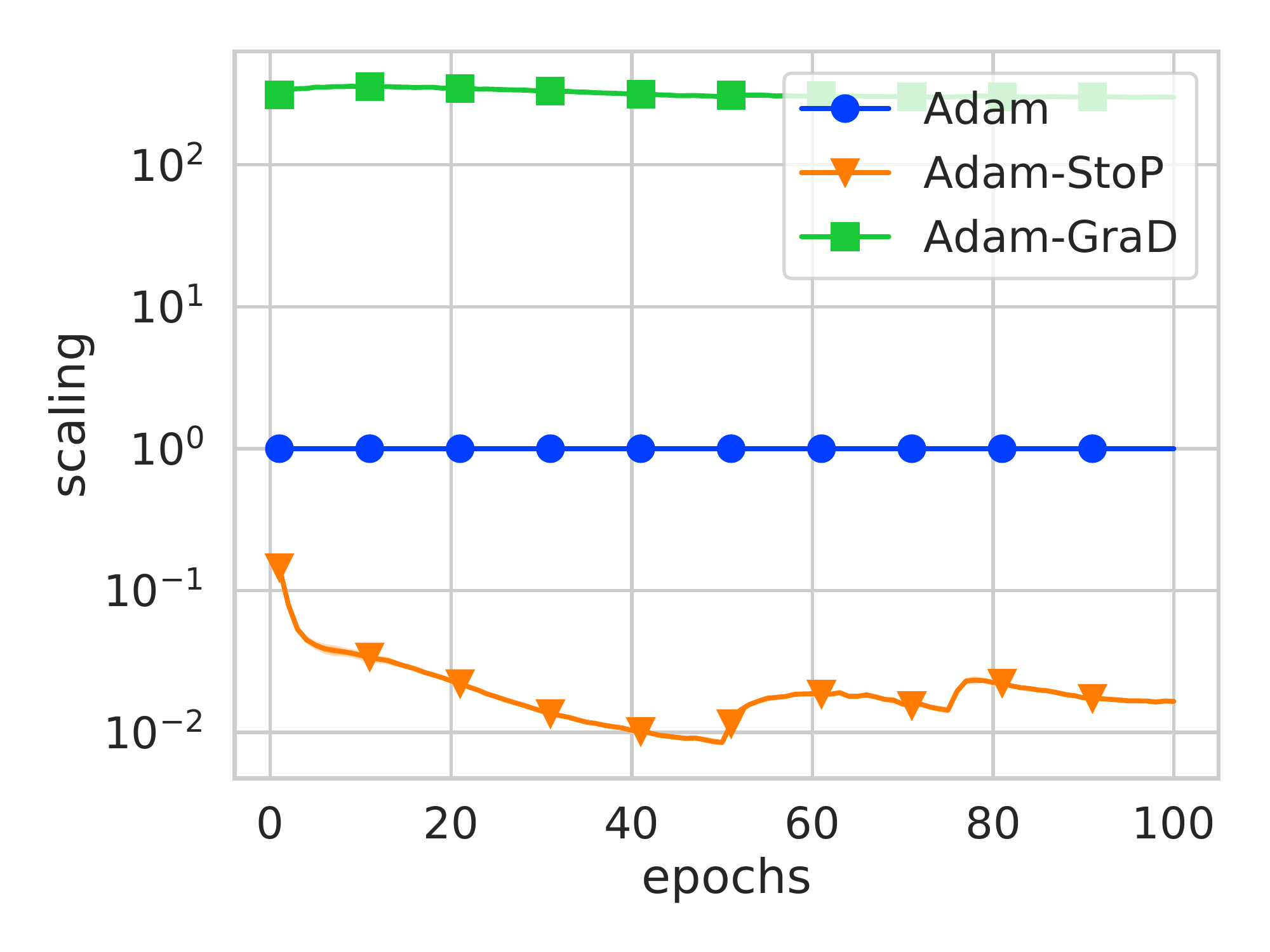}
\caption{ResNet34 on CIFAR100.  For details see caption in Figure~\ref{fig:lenet_mnist}.}
\label{fig:resnet34_cifar100}
\end{figure}

\section{Proofs}

\subsection{Proof of Theorem~\ref{thm:stps_convergence}}

\begin{proof}
We denote $\Ek{\cdot}$ to be expectation conditioned on $\xk$, then
\begin{align*}
\Ek{\ns{\xkp - \xs}} &= \ns{\xk - \xs} - 2 \Ek{\gamma_{\stps}^k \avein \dotprod{\gik -\gis , \xk - \xs}} \\
& \quad + \Ek{(\gamma_{\stps}^k)^2 \ns{\avein (\gik - \gis)}} \\
&\overset{\eqref{eq:stps_stepsize}}{=} \ns{\xk - \xs} - 2 \Ek{\gamma_{\stps}^k\avein \dotprod{\gik -\gis , \xk - \xs}} \\
& \quad + 2\Ek{\gamma_{\stps}^k \avein \lp f(\xk, \xi_i) - f(\xs, \xi_i) - \dotprod{\gis,\xk - \xs} \rp} \\
&= \ns{\xk - \xs} \\
& \quad - 2\Ek{\gamma_{\stps}^k \avein  f(\xk, \xi^k_i) - f(\xs, \xi^k_i) - \dotprod{\gik, \xk-\xs}}.
\end{align*}
Since $ f(x^k, \xi^k_i) - f(\xs, \xi^k_i) - \dotprod{\gik, \xk-\xs} \leq 0$ due to the convexity of $f(\cdot, \xi_i^k)$, we can replace $\gamma_{\stp}^k \rightarrow \gamma_{\stp}^{\min}$ and take expectation. Therefore,
\begin{align*}
\Ek{\ns{\xkp - \xs}} &\leq \ns{\xk - \xs} - 2\gamma_{\stp}^{\min} \lp f(\xk) - f(\xs) - \dotprod{\nabla f(\xk), \xk-\xs} \rp\\
&\overset{\eqref{eq:mu_convexity}}{\leq}\lp 1 - \mu\gamma_{\stp}^{\min} \rp\ns{\xk - \xs}.
\end{align*}
Applying this inequality recursively with the tower property of expectation concludes the proof. 
\end{proof}

\newpage

\subsection{Proof of Theorem~\ref{thm:grads_convergence}}

\begin{proof}
We denote $\Ek{\cdot}$ to be expectation conditioned on $\xk$, then
\begin{align*}
\Ek{\ns{\xkp - \xs}} &= \ns{\xk - \xs} - 2\eta \Ek{\gamma_{\grads}^k \avein \dotprod{\gik - \gis, \xk - \xs}} \\
&\quad + \eta^2\Ek{(\gamma_{\grads}^k)^2 \ns{\avein (\gik - \gis)}} \\
&\overset{\eqref{eq:loc_smoothnes},\; \eqref{eq:convexity}}{\leq} \ns{\xk - \xs} - \eta\Ek{\gamma_{\grads}^k \avein \frac{1}{L}  \ns{\gik - \gis}} \\
& \quad - \eta \gamma_{\grads}^{\min}\Ek{ \avein \dotprod{\gik - \gis, \xk - \xs}} \\
&\quad + \eta^2\Ek{(\gamma_{\grads}^k)^2 \ns{\avein (\gik - \gis)}} \\
&\overset{\eqref{eq:SGD_grads}}{=} \ns{\xk - \xs} - \eta\Ek{\eta\gamma_{\grads}^k \avein \ns{\gik - \gis} \lp\frac{1}{L} - \eta\rp } \\
& \quad - \eta \gamma_{\grads}^{\min}\Ek{ \avein \dotprod{\nabla f(\xk) - \gis, \xk - \xs}} \\
&\overset{\eqref{eq:mu_convexity},\; \eta = \nicefrac{1}{L}}{\leq}  \lp 1 - \gamma_{\grads}^{\min}\frac{\mu}{L} \rp\ns{\xk - \xs}.
\end{align*}
$\gamma_{\grads}^{\min} \geq 1$ follows from directly from the Cauchy-Schwartz inequality in the form $\lp\sum_{i = 1}^n a_i\rp^2 \leq n  \sum_{i = 1}^n a_i^2 $ applied to \eqref{eq:adjusted_grad_div}.
\end{proof}

\newpage

\subsection{Proof of Theorem~\ref{thm:stp_convergence}}

\begin{proof}
We denote $\Ek{\cdot}$ to be expectation conditioned on $\xk$, then
\begin{align*}
\Ek{\ns{\xkp - \xs}} &= \ns{\xk - \xs} - 2 \Ek{\gamma_{\stp}^k \avein \dotprod{\gik , \xk - \xs}} \\
& \quad + \Ek{(\gamma_{\stp}^k)^2 \ns{\avein \gik}} \\
&\overset{\eqref{eq:SGD_stp}}{\leq} \ns{\xk - \xs} - 2 \Ek{\gamma_{\stp}^k\avein \dotprod{\gik , \xk - \xs}} \\
& \quad + 2 \Ek{\gamma_{\stp}^k \avein \lp f(\xs, \xi_i) - \fis \rp} \\
&= \ns{\xk - \xs} + 2 \Ek{\gamma_{\stp}^k \avein \lp f(\xk, \xi_i) - \fis \rp}\\
& \quad - 2\Ek{\gamma_{\stp}^k \avein \lp f(\xk, \xi^k_i) - f(\xs, \xi^k_i) - \dotprod{\gik, \xk-\xs} \rp}.
\end{align*}
Since $ f(x^k, \xi^k_i) - f(\xs, \xi^k_i) - \dotprod{\gik, \xk-\xs} \geq 0$ due to the convexity of $f(\cdot, \xi_i^k)$, we can replace $\gamma_{\stp}^k \rightarrow \gamma_{\stp}^{\min}$ and take expectation. Therefore,
\begin{align*}
\Ek{\ns{\xkp - \xs}} &\leq \ns{\xk - \xs}  + 2 \Ek{\gamma_{\stp}^k \lp f(\xk, \xi_i) - \fis \rp}\\ 
& \quad- 2\gamma_{\stp}^{\min} \lp f(\xk) - f(\xs) - \dotprod{\nabla f(\xk), \xk-\xs} \rp \\
&\overset{\eqref{eq:mu_convexity}}{\leq}\lp 1 - \mu\gamma_{\stp}^{\min} \rp\ns{\xk - \xs}  + \lp 1 + \frac{\mu\gamma_{\stp}^{\min}}{2} \rp^k 2\gamma_{\stp}^{\min}\sigma^2_\stp.
\end{align*}
Applying this inequality recursively with the tower property of expectation, we obtain
\begin{align*}
\E{\ns{\xk - \xs}} &\leq \lp 1 - \mu\gamma_{\stp}^{\min} \rp^k \ns{x^0 - \xs}  + 2\gamma_{\stp}^{\min}\sigma^2_\stp \sum_{t=0}^{k-1} \lp 1 + \frac{\mu\gamma_{\stp}^{\min}}{2} \rp^t \lp 1 - \mu\gamma_{\stp}^{\min} \rp^t \\
&\leq \lp 1 - \mu\gamma_{\stp}^{\min} \rp^k \ns{x^0 - \xs}  + 2\gamma_{\stp}^{\min}\sigma^2_\stp \sum_{t=0}^{k-1} \lp 1 - \frac{\mu\gamma_{\stp}^{\min}}{2} \rp^t  \\
&\leq \lp 1 - \mu\gamma_{\stp}^{\min} \rp^k \ns{x^0 - \xs}  + 4\frac{\sigma^2_\stp}{\mu}.
\end{align*}
\end{proof}

\newpage

\subsection{Proof of Theorem~\ref{thm:grad_convergence}}

\begin{proof}
We denote $\Ek{\cdot}$ to be expectation conditioned on $\xk$, then
\begin{align*}
\Ek{\ns{\xkp - \xs}} &= \ns{\xk - \xs} - 2\eta \Ek{\gamma_{\grad}^k \avein \dotprod{\gik, \xk - \xs}} \\
&\quad + \eta^2\Ek{(\gamma_{\grad}^k)^2 \ns{\avein \gik}} \\
&\overset{\eqref{eq:SGD_grad}}{\leq} \ns{\xk - \xs} - 2\eta \Ek{\gamma_{\grad}^k \avein \lp\dotprod{\gik, \xk - \xis} + \dotprod{\gik, \xis - \xs} \rp} \\
&\quad + \eta^2\Ek{\gamma_{\grad}^k \avein \ns{ \gik}} \\
&\overset{\eqref{eq:loc_smoothnes},\; \eqref{eq:convexity}}{\leq} \ns{\xk - \xs}  - \eta \Ek{\gamma_{\grad}^k \avein \dotprod{\gik, \xk - \xis}}\\ 
& \quad - \eta\Ek{\gamma_{\grad}^k \avein \lp\lp \frac{1}{L} - \eta \rp  \ns{\gik} + 2\dotprod{\gik, \xis - \xs}\rp}.
\end{align*}
Since $ \dotprod{\gik, \xk - \xis} \geq 0$ due to the convexity of $f(\cdot, \xi_i^k)$, we can replace $\gamma_{\grad}^k \rightarrow \gamma_{\grad}^{\min}$. Therefore,
\begin{align*}
\Ek{\ns{\xkp - \xs}} &\leq  \ns{\xk - \xs}  - \eta \gamma_{\grad}^{\min}\Ek{ \avein \dotprod{\gik, \xk - \xs}}\\ 
& \quad - \eta\Ek{ \avein \lp \gamma_{\grad}^k \lp \frac{1}{L} - \eta \rp  \ns{\gik} + \lp 2\gamma_{\grad}^k -  \gamma_{\grad}^{\min}\rp\dotprod{\gik, \xis - \xs} \rp}.
\end{align*}
Taking expectation and upper bounding $$-\lp 2\gamma_{\grad}^k -  \gamma_{\grad}^{\min}\rp\dotprod{\gik, \xis - \xs} \leq  2\gamma_{\grad}^k \lp \frac{1}{4L}\ns{\gik} + 4L\ns{\xis - \xs}\rp $$ using Cauchy-Schwartz inequality, we get
\begin{align*}
\Ek{\ns{\xkp - \xs}} &\leq  \ns{\xk - \xs}  - \eta \gamma_{\grad}^{\min}\dotprod{\nabla f(\xk) - \nabla f(\xs), \xk - \xs}\\ 
& \quad - \eta\Ek{\gamma_{\grad}^k  \avein \lp \lp \frac{1}{2L} - \eta \rp  \ns{\gik} - 8L\ns{\xis - \xs}\rp} \\
&\overset{\eta = \nicefrac{1}{2L},\; \eqref{eq:mu_convexity}}{\leq}  \ns{\xk - \xs}\lp 1 - \frac{\mu \gamma_{\grad}^{\min}}{2L} \rp \\ 
& \quad + \eta 8L\Ek{\gamma_{\grad}^k  \avein \ns{\xis - \xs}} \\
& \leq  \ns{\xk - \xs}\lp 1 - \frac{\mu \gamma_{\grad}^{\min}}{2L} \rp + \frac{8\gamma_{\grad}^{\min}}{2L} \lp 1 + \frac{\mu \gamma_{\grad}^{\min}}{4L} \rp^k\sigma^2_{\grad}.
\end{align*}
Applying this inequality recursively with the tower property of expectation, we obtain
\begin{align*}
\E{\ns{\xk - \xs}} &\leq \lp 1 - \frac{\mu \gamma_{\grad}^{\min}}{2L}  \rp^k \ns{x^0 - \xs} \\
& \quad  + \frac{8\gamma_{\grad}^{\min}}{2L}\sigma^2_\grad \sum_{t=0}^{k-1} \lp 1 + \frac{\mu \gamma_{\grad}^{\min}}{4L}  \rp^t \lp 1 - \frac{\mu \gamma_{\grad}^{\min}}{2L} \rp^t \\
&\leq \lp 1 - \frac{\mu \gamma_{\grad}^{\min}}{2L} \rp^k \ns{x^0 - \xs}  + \frac{8\gamma_{\grad}^{\min}}{2L}\sigma^2_\grad \sum_{t=0}^{k-1} \lp 1 - \frac{\mu \gamma_{\grad}^{\min}}{4L}  \rp^t  \\
&\leq \lp 1 - \frac{\mu \gamma_{\grad}^{\min}}{2L} \rp^k \ns{x^0 - \xs}  + 16\frac{\sigma^2_\grad}{\mu}.
\end{align*}
$\gamma_{\grad}^{\min} \geq 1$ follows directly from the Cauchy-Schwartz inequality in the form $\lp\sum_{i = 1}^n a_i\rp^2 \leq n  \sum_{i = 1}^n a_i^2 $.
\end{proof}

\newpage

\subsection{Proof of Theorem~\ref{thm:momentum_convergence}}

\begin{proof}
Let us denote $G^k = \gamma \avein \gik$. Then \Cref{alg:moscagrad} is equivalent to the following updates
\begin{align*}
	x^{k+1} &= \frac{k}{k+1} x^k + \frac{1}{k+1}z^k, \\
	z^{k+1} &= z^k - 2\eta G^{k +1}.
\end{align*}
It follows directly from above that $z^{k}=(k+1) x^{k+1} - k x^{k}$ and
\begin{align*}
	x^{k+1}
	&= \frac{k}{k+1} x^k + \frac{1}{k+1}(z^{k-1} -2\eta G^k) \\
	&= \frac{k}{k+1} x^k + \frac{1}{k+1}(kx^k - (k-1) x^{k-1} - 2\eta G^k)  \\
	&= x^k - \frac{2}{k+1} \eta G^k + \frac{k-1}{k+1}(x^k - x^{k-1}) \\
	&= x^k - \eta  \gamma^{\min}_{\grad} \lp (1 - \beta^k) G^k+\beta^k m^{k-1} \rp.
\end{align*}
Using this scheme we can continue a similar way to the proof strategy of Theorem~8 from~\citep{taylor19a}. 
	Let us introduce the following Lyapunov function
	\begin{align*}
		\cL^{k}\eqdef \|z^k-\xs\|^2 + 4\eta  \gamma^{\min}_{\grad} k(f(\xk)-f(\xs)).
	\end{align*}
	We will show that $\cL^{k+1}\le \cL^k$, from which the theorem's claim follows immediately. First, note
	\begin{align*}
		\ns{z^{k+1}-\xs}
		&= \ns{z^k - \xs} - 4 \eta \dotprod{G^{k+1}, z^k-\xs} + 4 \eta^2\ns{G^{k+1}} \\
		&= \ns{z^k - \xs} - 4 \eta \dotprod{G^{k+1}, \xkp-\xs} + 4 \eta^2\ns{G^{k+1}}  \\
		& \quad - 4 \eta \dotprod{G^{k+1}, z^k-\xkp}.
	\end{align*}
	In addition, let $\Ekp{\cdot}$ denote expectation conditioned on $\xkp$, then
	\begin{align*}
		&\Ekp{\ns{z^k - \xs} - 4 \gamma^{\min}_{\grad} \avein\dotprod{G^{k+1}, \xkp-\xs} + 4 (\gamma^{\min}_{\grad})^2\ns{G^{k+1}}}\\
		& \quad \overset{\eqref{eq:loc_smoothnes},\;\eqref{eq:convexity}}{\leq} \ns{z^k - \xs} - 4\eta \gamma^{\min}_{\grad} (f(x^{k+1})-f^*) \\
		& \qquad - 2 \gamma^{\min}_{\grad} \eta\Ekp{\frac{1}{L} \avein\ns{g_i^{k+1}} - 2 \gamma^{\min}_{\grad} \eta\ns{\avein g_i^{k+1} }} \\
		& \quad \overset{\eta = \nicefrac{1}{2L}}{\leq} \ns{z^k - \xs} - 4\eta \gamma^{\min}_{\grad} (f(x^{k+1})-f^*).
	\end{align*}
	Now we can take expectation of the outstanding scalar product to obtain
	\begin{align*}
		\Ekp{\dotprod{ G^{k+1}, z^k-\xkp}} = \gamma^{\min}_{\grad}\dotprod{\nabla f(\xkp), z^k-\xkp} = \dotprod{\nabla f(\xkp), \gamma k (\xkp-\xk)}.
	\end{align*}
	Using convexity we deduce
	\begin{align*}
		\dotprod{\nabla f(\xkp), \xkp-\xk} \geq f(\xkp)-f(\xk).
	\end{align*}
	Combining it with the previous bound yields
	\begin{align*}
		\Ekp{\ns{z^{k+1}-\xs} + 2\gamma(k+1)\lp f(\xkp) - f(\xs)\rp}
		\leq \ns{z^k-\xs} + 2\gamma k \lp f(\xk) - f(\xs)\rp,
	\end{align*}
	which is equivalent to $\cL^{k+1}\le \cL^k$.
\end{proof}

\newpage

\section{Efficient implementation of \grad for linear models}
Assume we are interested in using the \grad step size for a least-squares problem,
\begin{align*}
	\min_x \frac{1}{2N}\|Ax-b\|^2.
\end{align*}
Then it is a special case of our framework with $f(x, j) = \frac{1}{2}(a_j^\top x - b_j)^2$. A minibatch gradient can be computed using efficient parallelization of matrix-vector multiplication, namely one can parallize computation of $A_S x$, where $A_S$ is a subset of rows of $A$. Moreover, it is easy to see that
\begin{align*}
	\avein \|g_i^k\|^2
	= \frac{1}{n}\sum_{j \in S^k} \|a_j\|^2(a_j^\top x^k - b_j)^2
	= \<[\|a_j\|^2]_{j \in S^k}, (A_{S^k} x^k - b_{S^k}) ^2>,
\end{align*}
so we can parallelize both computation of $A_{S^k}x$ and $[\|a_j\|^2]_{j \in S^k}$.

Similarly, for logistic regression problem, defining $h(z)\eqdef 1 / (1+e^{-z})$ to be the sigmoid function, the objective is
\begin{align*}
	\min_x \frac{1}{N}\sum_{j=1}^N -(b_j\log(h((a_j^\top x)) +(1-b_j)\log(1- h(a_j^\top x)).
\end{align*}
To compute $\avein g_i^k$ and $\avein \|g_i^k\|^2$ we therefore need to compute in parallel $A_{S^k} x$, the component-wise activation, $h(A_{S^k}x)$ and the vector $[\|a_j\|^2]_{j \in S^k}$. Then
\begin{align*}
	\avein \|g_i^k\|^2
	= \frac{1}{n}\sum_{j \in S^k} \|a_j\|^2(h(a_j^\top x^k) - b_j)^2
	= \<[\|a_j\|^2]_{j \in S^k}, (h(A_{S^k}x^k) - b_{S^k})^2>,
\end{align*}
which is efficiently implementable.

\end{document}